\documentclass[twoside]{article}

\usepackage[accepted]{aistats2026}
\usepackage[authoryear]{natbib}

\setcitestyle{authoryear,aysep={,},yysep={,}}

\usepackage{hyperref}
\usepackage{url}

\usepackage{microtype}
\usepackage{graphicx}
\usepackage{subfigure}
\usepackage{booktabs} 
\usepackage{enumitem}
\usepackage{amsfonts}
\usepackage{amsmath}
\usepackage{amssymb}
\usepackage{mathtools}
\usepackage{amsthm}
\usepackage{multirow}
\PassOptionsToPackage{dvipsnames}{xcolor}
\usepackage{placeins}
\usepackage{bbm}
\usepackage{array}
\usepackage{stackrel}
\usepackage{algpseudocode}
\usepackage{algorithm}
\usepackage{colortbl}
\usepackage{tabu}
\usepackage{enumitem}
\usepackage{caption}
\usepackage{wrapfig}
\usepackage{fancybox}

\usepackage{xspace}
\usepackage{tikz-cd}

\newcommand\ie{i.\,e.\xspace}
\newcommand\eg{e.\,g.\xspace}

\algdef{SE}[SUBALG]{Indent}{EndIndent}{}{\algorithmicend\ }%
\algtext*{Indent}
\algtext*{EndIndent}

\newcommand{\ind}{\perp\!\!\!\perp}

\DeclareMathOperator*{\argmin}{arg\,min}
\newcommand*\diff{\mathop{}\!\mathrm{d}}

\newcommand{\abs}[1]{\left\lvert #1 \right\rvert}

\newcommand{\norm}[1]{\left\lVert#1\right\rVert}

\usepackage{scalerel,stackengine,amsmath}

\usepackage{pifont}
\newcommand{\cmark}{\textcolor{ForestGreen}{\ding{51}}}%
\newcommand{\xmark}{\textcolor{BrickRed}{\ding{55}}}%

\newcolumntype{P}[1]{>{\centering\arraybackslash}p{#1}}

\newtheorem{lemma}{Lemma}
\newtheorem{definition}{Definition}

\newtheorem{prop}{Proposition}

\newtheorem{assumption}{Assumption}

\newtheorem{innercustomlemma}{Lemma}
\newenvironment{numlemma}[1]
  {\renewcommand\theinnercustomlemma{#1}\innercustomlemma}
  {\endinnercustomlemma}

\newtheorem{innercustomprop}{Proposition}
\newenvironment{numprop}[1]
  {\renewcommand\theinnercustomprop{#1}\innercustomprop}
  {\endinnercustomprop}

\usepackage{tcolorbox}

\usepackage{pifont}
\usepackage{tikz}
\usepackage{graphicx}

\newcommand*\circledred[1]{%
\tikz[baseline=(char.base)]{
  \node[shape=circle, draw=black!60, fill=lightgray!10, thick, inner sep=1pt] (char) {\scriptsize{#1}};
}}
\newcommand*\circled[1]{%
\tikz[baseline=(char.base)]{
  \node[shape=circle, draw=black!60, fill=lightgray!10, thick, inner sep=1pt] (char) {\scriptsize{#1}};
}}

\definecolor{tabblue}{HTML}{2077B4}
\definecolor{tabred}{HTML}{FF0000}
\definecolor{yellow}{HTML}{FFFF88}

\newtcbox{\myovalbox}{colback=yellow,boxrule=0.1pt,arc=3pt,
  boxsep=0pt,left=0.5pt,right=0.5pt,top=0.5pt,bottom=0.5pt,}

\newcommand{\ORlearners}{\emph{OR-learners}\xspace}

\begin{document}

%

%

\twocolumn[

\aistatstitle{Orthogonal Representation Learning for Estimating Causal Quantities}

\aistatsauthor{ Valentyn Melnychuk \And Dennis Frauen \And Jonas Schweisthal \And Stefan Feuerriegel }

\aistatsaddress{ LMU Munich \& Munich Center for Machine Learning \\
    Munich, Germany\\
    \texttt{melnychuk@lmu.de}} ]

\begin{abstract}
    End-to-end representation learning has become a powerful tool for estimating causal quantities from high-dimensional observational data, but its efficiency remained unclear.
    Here, we face a central tension: End-to-end representation learning methods often work well in practice but lack asymptotic optimality in the form of the quasi-oracle efficiency. In contrast, two-stage Neyman-orthogonal learners provide such a theoretical optimality property but do not explicitly benefit from the strengths of representation learning. In this work, we step back and ask two research questions: (1) \emph{When do representations strengthen existing Neyman-orthogonal learners?} and (2) \emph{Can a balancing constraint -- a commonly proposed technique in the representation learning literature -- provide improvements to Neyman-orthogonality?} We address these two questions through our theoretical and empirical analysis, where we introduce a unifying framework that connects representation learning with Neyman-orthogonal learners (namely, \emph{OR-learners}). In particular, we show that, under the low-dimensional manifold hypothesis, the OR-learners can strictly \emph{improve} the estimation error of the standard Neyman-orthogonal learners. At the same time, we find that the balancing constraint requires an additional inductive bias and \emph{cannot} generally compensate for the lack of Neyman-orthogonality of the end-to-end approaches. Building on these insights, we offer guidelines for how users can effectively combine representation learning with the classical Neyman-orthogonal learners to achieve both practical performance and theoretical guarantees.
    
\end{abstract}

\vspace{-0.3cm}
\section{INTRODUCTION} 
\vspace{-0.3cm}

Estimating causal quantities has many applications in medicine \citep{feuerriegel2024causal}, policy-making \citep{kuzmanovic2024causal}, marketing \citep{varian2016causal}, and economics \citep{basu2011estimating}. 
In this work, we focus on the \emph{individualized causal quantities}, namely, the conditional average treatment effect (CATE) and the conditional average potential outcomes (CAPOs). 

Recently, \textbf{end-to-end representation learning methods} have gained wide popularity in estimating causal quantities from observational data \citep[e.g.,][]{johansson2016learning,shalit2017estimating,hassanpour2019counterfactual,hassanpour2019learning,zhang2020learning,assaad2021counterfactual,johansson2022generalization}. The key benefit of the representation learning methods is that they utilize a key \emph{low-dimensional manifold hypothesis} \citep{fefferman2016testing}, namely, that the nuisance functions and the ground-truth causal quantity are both defined on some low-dimensional manifold of the covariate space.

A different literature stream seeks to estimate causal quantities through
\textbf{two-stage Neyman-orthogonal learners} \citep{morzywolek2023general,vansteelandt2025orthogonal}. Prominent examples are the DR-learner \citep{kennedy2023towards}, R-learner \citep{nie2021quasi}, and IVW-learner \citep{fisher2024inverse}. Unlike end-to-end representation learning, Neyman-orthogonal learners offer several \emph{favorable asymptotic properties}, namely quasi-oracle efficiency and double robustness \citep{chernozhukov2017double,foster2023orthogonal}.

However, we arrive at a central tension: On the one hand, end-to-end representation learning methods might substantially help with the high-dimensional data; while, on the other, Neyman-orthogonal meta-learners possess asymptotic optimality properties. In this work, we aim to address this tension by answering \emph{two core research questions} (\textbf{RQ}).

\vspace{-0.1cm}
\begin{tcolorbox}[colback=gray!5!white,colframe=black!75!black, boxsep=-1mm,left=10pt,right=10pt]
    \textbf{RQ} \circledred{1}. When do representations
    strengthen the existing Neyman-orthogonal learners?
\end{tcolorbox}
\vspace{-0.1cm}

\textbf{RQ} \circledred{1} has not yet been fully addressed. For example, the prior work \citep{schulte2025adjustment} studied how the representations can facilitate a semi-parametric average-treatment effect (ATE) estimation (analogous to quasi-oracle efficiency when the causal quantity is finitely-dimensional). \citep{curth2021nonparametric} provided an empirical study of CATE meta-learners with representation learning used for the estimation of the nuisance functions, but the learned representations have not been used as the target model inputs. Hence, strategies to reconcile the advantages of representation learning and Neyman-orthogonal learners in a principled manner are unclear.

As an alternative strategy to reduce the variance of the estimation, some end-to-end representation learning methods suggested a \emph{balancing constraint} for the representations  \citep{johansson2016learning,shalit2017estimating}. This motivates our second question.

\vspace{-0.1cm}
\begin{tcolorbox}[colback=gray!5!white,colframe=black!75!black,boxsep=-1mm,left=10pt,right=10pt]
    \textbf{RQ} \circledred{2}. When can the balancing constraint improve the efficiency of learning similarly to Neyman-orthogonality?
\end{tcolorbox}
\vspace{-0.1cm}

To the best of our knowledge, \textbf{RQ} \circledred{2} has not yet been studied. As discovered in \citep{melnychuk2024bounds}, the balancing constraint can lead to a \emph{representation-induced confounding bias} (RICB), and it is only guaranteed to omit the RICB (and perform a consistent estimation) for \emph{invertible} representations \citep{johansson2019support}.

We analyze the \textbf{RQ} \circledred{1} and \textbf{RQ} \circledred{2} by introducing a unifying framework of Neyman-orthogonal representation learners for CAPOs/CATE, namely \emph{OR-learners}. The \emph{OR-learners} employ representation learning at \emph{both stages of training}: (a)~we use representation learning to jointly estimate the nuisance functions, and (b)~we use the learned representations as the inputs for the target models. Based on the OR-learners, we then answer both research questions:  

To answer \textbf{RQ} \circledred{1}, we provide \emph{sufficient theoretical guarantees}, under which the \ORlearners are guaranteed to outperform the existing Neyman-orthogonal learners. Here, unlike \citep{curth2021nonparametric}, we use the learned representations at both stages of Neyman-orthogonal learners.

In response to \textbf{RQ} \circledred{2}, we discover that the balancing constraint heavily relies on an additional inductive bias that the \emph{low-overlap regions of the covariate space also have the low heterogeneity of the ground-truth CAPOs/CATE}. Therefore, in general, the balancing constraint (no matter whether invertibility is enforced) is \emph{not} a substitute for Neyman-orthogonality and is, thus, asymptotically inferior to the \ORlearners. One of the important consequences of \textbf{RQ} \circledred{2} is that the \ORlearners, due to their Neyman-orthogonality, \emph{asymptotically outperform both categories of end-to-end methods: (a)~with unconstrained and (b)~with balanced representations}. 

In sum, \textbf{our main contribution} is the following:
By answering our \textbf{RQ} \circledred{1} and \textbf{RQ} \circledred{2}, we propose \emph{guidelines} on how a causal ML practitioner can effectively combine the representation learning with the Neyman-orthogonal meta-learners. Specifically, we provide the conditions under which the \ORlearners, the combination of both (i)~the representation learning and (ii)~Neyman-orthogonal meta-learners, perform better than each approach (i) and (ii) separately.

\vspace{-0.1cm}
\section{RELATED WORK} \label{sec:related-work}
\vspace{-0.3cm}
Our work aims to unify two streams of work, namely, representation learning methods and Neyman-orthogonal learners. We briefly review both in the following (a detailed overview is in Appendix~\ref{app:extended-rw}).

\textbf{End-to-end representation learning.} Several methods have been previously introduced for \emph{end-to-end} representation learning of CAPOs/CATE  (see, in particular, the seminal works \citep{johansson2016learning,shalit2017estimating,johansson2022generalization}). A large number of works later suggested different extensions to these.
Existing methods fall into three main streams: (1)~One can fit an \emph{unconstrained shared representation} to directly estimate both potential outcome surfaces (e.g., \textbf{TARNet} \citep{shalit2017estimating}). (2)~Some methods additionally enforce a \emph{balancing constraint} based on empirical probability metrics, so that the distributions of the treated and untreated representations become similar (e.g., \textbf{CFR} and \textbf{BNN} \citep{johansson2016learning,shalit2017estimating}). Importantly, balancing based on empirical probability metrics is only guaranteed to perform a consistent estimation for \emph{invertible} representations since, otherwise, balancing leads to a \emph{representation-induced confounding bias} (RICB) \citep{johansson2019support,melnychuk2024bounds}. (3)~One can additionally perform \emph{balancing by re-weighting} the loss and the distributions of the representations with learnable weights (e.g., \textbf{RCFR} \citep{johansson2022generalization}). We later adopt the methods from (1)--(3) as baselines. 

\textbf{Neyman-orthogonal learners}. Causal quantities can be estimated using model-agnostic methods, so-called \emph{meta-learners} \citep{kunzel2019metalearners}. Prominent examples are the DR-learner \citep{kennedy2023towards,curth2020estimating}, R-learner \citep{nie2021quasi}, and IVW-learner \citep{fisher2024inverse}. 
Meta-learners are model-agnostic in the sense that any base model (e.g., neural network) can be used for estimation. Also, meta-learners have several practical advantages \citep{morzywolek2023general}: (i)~they oftentimes offer favorable theoretical guarantees such as Neyman-orthogonality \citep{chernozhukov2017double,foster2023orthogonal}; (ii)~they can address the causal inductive bias that the CATE is ``simpler'' than CAPOs \citep{curth2021inductive}, and (iii)~the target model obtains a clear interpretation as a projection of the ground-truth CAPOs/CATE on the target model class. {\citep{curth2021nonparametric,frauen2025modelagnostic} provided a comparison of meta-learners implemented via neural networks with different representations, yet with the target model based on the original covariates (the representations were only used as an interim tool to estimate nuisance functions). In contrast, here, we study the learned representations as primary inputs to the target model.}

\textbf{Representation learning and efficient estimation}. Perhaps the closest to ours is \citep{schulte2025adjustment}. Therein, the authors provided theoretical and empirical evidence on how representation learning might help with semi-parametric efficient estimation, but only for the ATE. Yet, a comprehensive study for CAPOs/CATE about the combination of representation learning and Neyman-orthogonal learners 
(= an analogue of semi-parametric estimators for the infinitely-dimensional estimands) 
is \emph{missing}.

\textbf{Research gap.} From a \emph{representation learning perspective}, multiple works simply suggested specific representation learning models, yet they did not properly explain where the gain of their performance comes from. 
On the other hand, from a perspective of \emph{Neyman-orthogonal learners}, rigorous studies of structural assumptions and representation learning are very recent and have not been done for CAPOs/CATE. Our work is the first to unify representation learning methods and Neyman-orthogonal learners and give definitive answers to \textbf{RQ} \circledred{1} and \textbf{RQ} \circledred{2}. 
As a result, we develop guidelines for how practitioners can effectively combine representation learning with the classical Neyman-orthogonal learners.

\vspace{-0.1cm}
\section{PRELIMINARIES} \label{sec:prelim}
\vspace{-0.3cm}

\textbf{Notation.} We denote random variables with capital letters $Z$, their realizations with small letters $z$, and their domains with  calligraphic letters $\mathcal{Z}$. Let $\mathbb{P}(Z)$, $\mathbb{P}(Z = z)$, $\mathbb{E}(Z)$ be the distribution, probability mass function/density, and expectation of $Z$, respectively. Let $\mathbb{P}_n\{f(Z)\} = \frac{1}{n} \sum_{i=1}^n f(z_i)$ be the sample average of $f(Z)$. We also denote the $L_p$ norm as $\norm{f}_{L_p} = (\mathbb{E}(\abs{f(Z)}^p))^{1/p}$.
We define the following nuisance functions: $\pi_a^x(x) = \mathbb{P}(A = a \mid X = x)$ is the \emph{covariate propensity score} for the treatment $A$, and $\mu_a^x(x) = \mathbb{E}(Y \mid X = x, A = a)$ is the \emph{expected covariate-level outcome} for the outcome $Y$. We also denote $\mu^x(x) = \pi_0^x(x) \mu_0^x(x) + \pi_1^x(x) \mu_1^x(x)$. Similarly, we define $\pi_a^\phi(\phi) = \mathbb{P}(A = a \mid \Phi(X) = \phi)$ and  $\mu_a^\phi(\phi) = \mathbb{E}(Y \mid \Phi(X) = \phi, A = a)$ as the \emph{representation propensity score} and the \emph{expected representation-level outcome} for a representation $\Phi(x) = \phi$, respectively. Importantly, the superscripts in $\pi_a^x,\mu_a^x, \pi_a^\phi,\mu_a^\phi$ indicate whether the corresponding nuisance functions depend on the covariates $x$ or on the representation $\phi$. 

\textbf{Problem setup.} To estimate the causal quantities, we make use of an observational dataset $\mathcal{D}$ that contains high-dimensional covariates $X \in \mathcal{X} \subseteq \mathbb{R}^{d_x}$, a binary treatment $A \in \{0, 1\}$, and a continuous outcome $Y \in \mathcal{Y} \subseteq \mathbb{R}$. For example, a common setting is an anti-cancer therapy, where the outcome is the tumor growth, the treatment is whether chemotherapy is administered, and covariates are patient information such as age and sex. The dataset $\mathcal{D} = \{z_i = (x_i, a_i, y_i)\}_{i=1}^{n}$ is assumed to be sampled i.i.d. from a joint distribution $\mathbb{P}(Z) =\mathbb{P}(X, A, Y)$ with dataset size $n$.

\textbf{Causal quantities.} We are interested in the estimation of two important causal quantities at the covariate level of heterogeneity: $\bullet$\,\emph{conditional average potential outcomes (CAPOs)} given by $\xi_a^x(x)$, and $\bullet$\,the \emph{conditional average treatment effect (CATE)} given by $\tau^x(x)$, with  $\xi_a^x(x) = \mathbb{E}(Y[a] \mid X = x) \quad \text{and} \quad \tau^x(x) = \mathbb{E}(Y[1] - Y[0] \mid X = x) = \xi_1^x(x) - \xi_0^x(x)$. If we could directly observe the true outcomes $Y[a]$ under both treatments (or the corresponding treatment effect $Y[1] - Y[0]$), then the consistent estimation of CAPOs and CATE, respectively, would reduce to a standard regression problem, but this is impossible due to the fundamental problem of causal inference.

\textbf{Assumptions.} To consistently estimate the causal quantities given only the observational data $\mathcal{D}$, we need to make standard assumptions \citep{rubin1974estimating,curth2021nonparametric,kennedy2023towards}. \textbf{(1)}~For \textbf{identifiability}, we assume (i)~\emph{consistency}: if $A = a$, then $Y[a] = Y$; (ii)~\emph{overlap}: $\mathbb{P}(0 < \pi^x_a(X) < 1) = 1$;  and (iii)~\emph{unconfoundedness}: $(Y[0], Y[1]) \ind A \mid X$. Given the assumptions (i)--(iii), both CAPOs and CATE are identifiable from $\mathbb{P}(Z)$ as expected covariate-level outcomes, $\xi_a^x(x) = \mu_a^x(x)$, or as the difference of expected covariate-level outcomes, $\tau^x(x) = \mu_1^x(x) - \mu_0^x(x)$, respectively. \textbf{(2)}~For \textbf{estimability} of the CAPOs/CATE from the finite-sized dataset $\mathcal{D}$ (\eg, with neural networks), we assume that (i)~$\mathcal{X}$ is compact; and (ii)~the H\"older smoothness of the ground-truth causal quantities and the nuisance functions (see Appendix~\ref{app:background} for definitions). Specifically, we assume the CAPOs, $\xi_a^x(\cdot) = \mu_a^x(\cdot)$, to be $s_{\mu_a}^x$-smooth; the CATE $\tau^x(\cdot)$ to be $s_{\tau}^x$-smooth; and the ground-truth propensity score $\pi^x_a(\cdot)$ to be $s_{\pi}^x$-smooth ($s_{\mu_a}^x, s_{\tau}^x, s_{\pi}^x > 0$). We also denote the corresponding H\"older norms $\norm{\cdot}_{C^{s^x}(\mathcal{X})}$ as $L_{\mu_a}^x, L_{\tau}^x, L_{\pi}^x > 0$; and Lipschitz constants as $[\cdot]_{C^1(\mathcal{X})}=\operatorname{Lip}(\cdot)$, where $[\cdot]_{C^s(\mathcal{X})}$ is a H\"older semi-norm. 

\textbf{End-to-end representation learning.} Representation learning methods make use of the identifiability formulas for CAPOs/CATE to fit a \emph{representation network}. Hence, the majority of the methods \citep[e.g.,][]{johansson2016learning,shalit2017estimating,hassanpour2019counterfactual,hassanpour2019learning,zhang2020learning,assaad2021counterfactual,johansson2022generalization} aim to (jointly) learn both expected covariate-level outcomes $\mu_a^x$ as a composition of a (a)~representation subnetwork $\Phi(x): \mathcal{X} \to \mathit{\Phi}$ and an (b)~outcome subnetwork $h_a(\phi): \mathit{\Phi} \times \mathcal{A} \to \mathcal{Y}$. Both (a) and (b) then aim to minimize a factual mean squared error (MSE) risk:
\begingroup\makeatletter\def\f@size{10}\check@mathfonts
\vspace{-0.1cm}
\begin{align} \label{eq:repr-plug-in}
    \hat{\mathcal{L}}_{\mathit{\Phi}}(h_a\circ{\Phi}) = \mathbb{P}_n \big\{(Y - h_A(\Phi(X)))^2 \big\}.
\end{align}
\endgroup
Some approaches further apply the covariate or representation propensity weights in Eq.~\eqref{eq:repr-plug-in}. As a result, all end-to-end representation learning methods are either plug-in or inverse probability of treatment weighted (IPTW) learners \citep{curth2021nonparametric}, which both generally suffer from the \emph{first-order error} of the model misspecification \citep{kennedy2023towards,morzywolek2023general}. 

\textbf{Neyman-orthogonal learners.} Here, we focus on \emph{two-stage Neyman-orthogonal learners} due to their theoretical advantages \citep{chernozhukov2017double,foster2023orthogonal}. Formally, two-stage learners aim to find the best projection of CAPOs/CATE onto a \emph{target model class} $\mathcal{G} = \{g(\cdot): \mathcal{V} \subseteq \mathcal{X} \to \mathcal{Y}\}$ by minimizing different \emph{target risks} wrt. $g(V)$ ($V \subseteq X$ is a conditioning set and the input for the target model). Then, the target risk is chosen as a (weighted) MSE:
\vspace{-0.1cm}
\begin{align} \label{eq:target-risk}
    \mathcal{L}_{\mathcal{G}}(g, \eta) = \mathbb{E} \big[w(\pi^x_a(X)) \, (\chi^x(X, \eta) - g(V))^2  \big],
\end{align}
where $\eta = (\mu_0^x, \mu_1^x, \pi^x_1)$ are the nuisance functions; $w(\cdot) >0 $ is a weighting function; and $\chi^x(\cdot)$ is the target causal quantity (\ie, $\chi^x(x, \eta) = \mu_a^x(x)$ for CAPOs and $\chi^x(x, \eta) = \mu_1^x(x) - \mu_0^x(x)$ for CATE). The Neyman-orthogonal learners proceed in two stages: (i)~the nuisance functions are learned, $\hat{\eta}$, and (ii)~\emph{debiased estimators} of the target risks ${\mathcal{L}}_{\mathcal{G}}(g, {\eta})$ are fitted wrt. $g$:
\begin{align} \label{eq:NO-loss}
    \hat{\mathcal{L}}_{\mathcal{G}}(g, \hat{\eta}) = \mathbb{P}_n \big\{\rho(A, \hat{\pi}^x_a(X)) \, (\phi(Z, \hat{\eta}) - g(V))^2 \big\},
\end{align}
where $\rho(\cdot)$ and $\phi(\cdot)$ are a learner-specific function and pseudo-outcome, respectively. For example, the DR-learner in the style\footnote{We introduce an alternative DR-learner in the style of \citet{foster2023orthogonal} in Appendix~\ref{app:background}.} of \citet{kennedy2023towards} is given by $w = \rho = 1$, $\phi_{\xi_a}(Z, {\eta}) = \mathbbm{1}\{A = a\} (Y - {\mu}_a^x(X)) / {\pi}^x_a(X) + {\mu}_a^x(X)$ for CAPOs estimation, and $\phi_{\tau}(Z, {\eta}) = (A - {\pi}^x_1(X)) (Y - {\mu}_A^x(X)) / ({\pi}^x_0(X) \, {\pi}^x_1(X)) + {\mu}_1^x(X)- {\mu}_0^x(X)$ for CATE estimation. The R-learner \citep{nie2021quasi} is given by  $w = {\pi}^x_0(X) {\pi}^x_1(X)$, $\rho = (A - {\pi}^x_1(X))^2$, and $\phi_{\tau}(Z, {\eta}) = (Y - \mu^x(X)) / (A - \pi^x_1(X))$. Similarly, the IVW-learner \citep{fisher2024inverse} is defined by the combination of the weighting functions of the R-learner, $w$ and $\rho$, and the pseudo-outcome of the DR-learner, $\phi_{\tau}$. Because the Neyman-orthogonal learners use the debiased target risks, they possess a \emph{quasi-oracle efficiency} and \emph{double robustness} (= asymptotic optimality properties). We provide a more detailed overview of meta-learners in Appendix~\ref{app:background}.

\textbf{End-to-end Neyman-orthogonality.} In a special case, the end-to-end IPTW-learner for CAPOs, \citep[e.g.,][]{assaad2021counterfactual} can possess Neyman-orthogonality. Yet, this IPTW-learner is a \emph{special case} of a general DR-learner \citep{foster2023orthogonal}, where the target model and the nuisance models coincide (see Appendix~\ref{app:background} for details).

\vspace{-0.1cm}
\section{ORTHOGONAL REPRESENTATION LEARNING} \label{sec:OR-learners}
\vspace{-0.3cm}

To answer our two core research questions \textbf{RQ} \circledred{1} and \textbf{RQ} \circledred{2}, we first introduce our unified representation of the existing Neyman-orthogonal learners, called \ORlearners. We provide proofs of theoretical statements in Appendix~\ref{app:proofs}.


\textbf{OR-learners}. The \ORlearners proceed in three stages.\footnote{Code is available at \url{https://github.com/Valentyn1997/OR-learners}.}  
In stage~\circled{0}, we fit a representation function $\hat{\Phi}$ that minimizes the plug-in/IPTW MSE from Eq.~\eqref{eq:repr-plug-in}. Then, in stage~\circled{1}, we fit additional nuisance functions $\hat{\eta}$ ($\hat{\pi}_a^x$ and, optionally, $\hat{\mu}_a^x$). Finally, in stage~\circled{2}, we use the learned representations from stage~\circled{0} as inputs to the target model $\hat{g}(\phi)$ that minimizes a Neyman-orthogonal loss from Eq.~\eqref{eq:NO-loss}. For \textbf{implementation details}, we refer to Appendix~\ref{app:implementation}.  

\begin{figure}
    \centering
    \vspace{-0.4cm}
    \begin{tikzcd}
        \large\mathcal{X}   \arrow[rrrrrrrr, "\xi^x_a/\tau^x \in C^{s^x}(\mathcal{X})" description, bend left=15, shift left] \arrow[rrrr, "\Phi^* \in C^{s^{x}}(\mathcal{X})" description] &  &  &  & \large \mathit{\Phi^*} \arrow[rrrr, "\xi^\phi_a/\tau^\phi \in C^{s^{\phi^*}}(\mathit{\Phi}^*)" description] \arrow[llll, "J^*\in C^{s^x+1}(\mathit{\Phi}^*)" description, bend left=20, shift left] &  &  &  & \large\mathcal{Y}
    \end{tikzcd}
    \vspace{-0.3cm}
    \caption{Visual summary of the relationships between the representation map $\Phi^*$, its pullback $J^*$, and target CAPOs/CATE from Assumption~\ref{ass:manifold}.}
    \label{fig:ass1-summary}
    \vspace{-0.5cm}
\end{figure}

\textbf{``Best of both worlds''}. The OR-learners combine the benefits of two streams of literature. On the one hand, (i)~the OR-learners are Neyman-orthogonal learners and thus inherit the same favorable asymptotic properties such as quasi-oracle efficiency and double robustness. On the other hand, they use (ii)~representation learning twice (unlike \citep{curth2021nonparametric}), namely, once for nuisance-function estimation and once for target-model fitting. As we will show later, under reasonable assumptions, the \ORlearners outperform both (i) standard Neyman-orthogonal learners with the original covariates as the inputs to the target models and (ii) end-to-end representation learning. 

\vspace{-0.1cm}
\subsection[RQ 1: OR-learners vs. standard Neyman-orthogonal learners]{RQ \circledred{1}: When Can Representations Improve Neyman-orthogonality? }
\vspace{-0.2cm}

To compare the standard Neyman-orthogonal learners with $V = X$ and the \ORlearners with $V = \Phi(X)$, we make use of the following lemma:
\begin{lemma}[Quasi-oracle efficiency of a non-parametric model] \label{lemma:quasi-oracle-eff}
    Assume a non-parametric\footnote{Similar convergence rate can be shown for neural networks, see Sec. 5.2 in \citep{schulte2025adjustment}.} target model $g(v), v \in \mathcal{V} \subseteq \mathbb{R}^{d_v}$. Then, the error between $g^{*v} = \argmin_{g \in \mathcal{G}}\mathcal{L}_{\mathcal{G}}(g, \eta)$ and $\hat{g}^v = \argmin_{g \in \mathcal{G}}\hat{\mathcal{L}}_{\mathcal{G}}(g, \hat{\eta})$ can be upper-bounded as:
    \vspace{-0.1cm}
    \begin{align} \label{eq:quasi-oracle-v}
        \norm{g^{*v} - \hat{g}^v}_{L_2}^2 \lesssim (L^v)^{\frac{2d_v}{(2s^v+d_v)}} n^{-\frac{{2s^v}}{{(2s^v + d_v)}}} + R_2,
    \end{align}
    where $g^{*v}$ is $s^v$-H\"older smooth with H\"older norm $L^v$, and $R_2 = R_2(\eta, \hat{\eta})$ is a second-order remainder that depends on $n$, $s^x_{\mu_a}, s^x_{\pi}$, and $d_x$.
\end{lemma}

We immediately see that we can reduce the error between $g^{*v}(v)$ and $\hat{g}^v(v)$ by  (1)~decreasing the dimensionality of the conditioning set $d_v$, (2)~decreasing the H\"older norm $L^v$ of the $g^{*v}(v)$, and (3)~increasing the H\"older smoothness $s^v$. 

\textbf{Heterogeneity trade-off.} In principle, all (1)-(3) can be optimized by choosing $V = \emptyset$, and, in this case, we recover a well-known \emph{semi-parametric efficient} estimator of  APOs/ATE. However, we then lose all the \emph{heterogeneity} of the potential outcomes/treatment effect: Although the error $\norm{g^{*v} - \hat{g}^v}_{L_2}^2$ gets smaller, the error between the ground-truth causal quantity and $\hat{g}$, $\norm{\xi^x_a - \hat{g}^v}_{L_2}^2$ or $\norm{\tau^x - \hat{g}^v}_{L_2}^2$, gets larger. Thus, we get a trade-off between \emph{(i)~reducing the error of the second-stage estimation} and \emph{(ii)~reducing the error with the ground-truth}.

\textbf{Structure-agnostic learner.} If we only care about the \emph{(ii)~reducing the error with the ground-truth}, the learner with $V = X$ is generally the best choice. That is, without any assumptions on the structure of covariates, $\hat{g}^v(v)$ provides a min-max optimal estimator of CAPOs/CATE \citep{balakrishnan2023fundamental,jin2025structure}. Specifically, in this case, we implicitly rely on the ``worst-case'' scenario that a ground-truth causal quantity \emph{densely depends on the full covariate set} $X$. Yet, at the same time, we struggle with the curse of dimensionality.  

\begin{figure}[t]
    \centering
    \vspace{-0.3cm}
    \includegraphics[width=\linewidth]{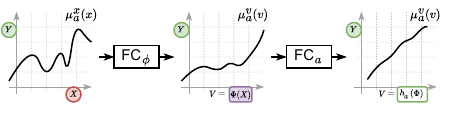}
    \vspace{-1.2cm}
    \caption{Hidden layers of the representation network induce spaces where the regression task is simpler.}
    \vspace{-0.5cm}
    \label{fig:holder-smooth}
\end{figure}

\textbf{Low-dimensional manifold hypothesis.} The CAPOs/CATE estimation can be greatly improved if we depart from the ``worst-case'' scenario and assume a low-dimensional manifold hypothesis.

\begin{assumption}[Manifold hypothesis] \label{ass:manifold}
    We assume (i)~the ground-truth causal quantities are supported on the low-dimensional compact smooth manifold (representation space) $\mathit{\Phi}^* \subseteq \mathbb{R}^{d_{\phi^*}}, d_{\phi^*} \ll d_x$ embedded into the covariate space:
    \vspace{-0.1cm}
    \begin{equation} \label{eq:valid-repr}
        \xi_a^x(x) =\xi_a^{\phi^*}(\Phi^*(x)) \text{ and } \tau^x(x) =\tau^{\phi^*}(\Phi^*(x)),
    \end{equation}
    where $\Phi^*(\cdot): \mathcal{X} \to \mathit{\Phi}^*$ is an $s^x$-H\"older smooth surjective embedding.\footnote{Note that we \emph{do not assume} that the propensity score is supported on the same manifold.} Furthermore, (ii)~there exists a (not necessarily unique) pullback map $J^*: \mathit{\Phi}^* \to \mathcal{X}$, such that $\Phi^* \circ J^* = \operatorname{id}_{\mathit{\Phi}^*}$, and $J^*$ is  $(s^x+1)$-H\"older smooth with a H\"older norm $L^{J^*}$.
\end{assumption}

Assumption~\ref{ass:manifold} (see a summary in Fig.~\ref{fig:ass1-summary}) helps to improve the error bound in Eq.~\eqref{eq:quasi-oracle-v}. Specifically, if we assume that the CAPOs/CATE $\tau^{\phi^*}/\xi_a^{\phi^*} \in \mathcal{G}$ (w.l.o.g.), then $g^{*\phi^*} \circ \Phi^*=g^{*x} =\xi_a^{x} /\tau^{x}$. However, the error bound gets lower when we learn $g^{*\phi^*}$ compared to $g^{*x}$. 
\begin{prop} \label{prop:manifold-quasi-oracle-eff}
    Under Assumption~\ref{ass:manifold}, the following holds:
    (1)~$d_{\phi^*} \ll d_x$, (2)~$g^{*\phi^*}$ is an $s^{\phi^*}$-H\"older smooth function with H\"older norm $L^{\phi^*}$ such that $s^{\phi^*}\ge s^x$ and $L^{\phi^*} \le c(L^{J^*}) \cdot L^x$ with non-decreasing $c(\cdot)$. Also, when $L^{J^*}$ is sufficiently small (\ie, a contractive map),
    \vspace{-0.1cm}
    \begin{align}
        \vert\vert{g^{*\phi^*} - \hat{g}^{\phi^*}}\vert\vert_{L_2}^2 &\lesssim \norm{g^{*x} - \hat{g}^x}_{L_2}^2.
    \end{align}
\end{prop}
\vspace{-0.1cm}

Note that in both cases ($V = X$ and $V = \Phi^*(X)$) the second-order remainder $R_2(\eta, \hat{\eta})$ is the same and depends on the dimensionality and the smoothness of the nuisance functions in the original space $\mathcal{X}$.
Proposition~\ref{prop:manifold-quasi-oracle-eff} then answers our main \textbf{RQ} \circledred{1}: Under Assumption~\ref{ass:manifold}, the \ORlearners with the known representation function ${\Phi}^*$ outperform the standard Neyman-orthogonal learners that are based on $V = X$. 
Now, we ask two follow-up questions: (i)~\emph{Is Assumption~\ref{ass:manifold} reasonable?} and (ii)~\emph{Can we learn $\Phi^*$?}

\textbf{(i)~Is Assumption~\ref{ass:manifold} reasonable?} This assumption has two main parts: (i)~Eq.~\eqref{eq:valid-repr} and (ii)~H\"older smoothness of the pullback $J^*$. (i)~Eq.~\eqref{eq:valid-repr} describes so-called \emph{valid representations} \citep{melnychuk2024bounds} or, alternatively, \emph{outcome mean sufficient representations} \citep{christgau2024efficient}. It means that the low-dimensional representation $\Phi^*$ has to contain all the sufficient information to model the ground-truth CAPOs/CATE. A trivial example of a valid representation is $\Phi^* = (\mu_0^x(x), \mu_1^x(x)) \in \mathbb{R}^2$.
The condition (ii) additionally requires that the representation pullback $J^*$ is H\"older smooth and, when $L^{J^*}$ is sufficiently small, it contracts the representation space. For example, by properties of H\"older smooth functions (see Appendix~\ref{app:background}), when $L^{J^*} \le 1/\sqrt{d_{\phi^*}}$, $\operatorname{Lip}(J^*) \le 1$. 
Alternatively, Assumption~\ref{ass:manifold} states that, (i)~when $ d_{\phi^*} \ll d_x$, the representation projects or averages irrelevant dimensions of $\mathcal{X}$; and, (ii)~when $L^{J^*}$ is sufficiently small,  the representation smoothens/expands the original covariate space $\mathcal{X}$ (\ie, when $s^x \ge 1$ and $L^{J^*} \le 1/\sqrt{d_{\phi^*}}$, then $\operatorname{Lip}(\Phi^*) \ge \operatorname{Lip}(J^*)^{-1} \ge  1$).

We argue that the manifold hypothesis (Assumption~\ref{ass:manifold}) is a very flexible and, often, the most realistic assumption for CATE/CAPOs estimation among \emph{different forms of structural knowledge} (such as additivity, sparsity, or linearity \citep{schulte2025adjustment}). For example, in the context of anti-cancer therapy, we might consider X-ray scans as high-dimensional covariates $X$. For them, it can be reasonable to assume that the ground-truth CATE/CAPOs lie in some low-dimensional manifold of the whole image space (\eg, a tumor might be fully characterized by several high-level covariates like shape, size, density, etc.).

\textbf{(ii)~Can we learn $\Phi^*$?} A natural question arises is whether we can learn the representation $\Phi^*(\cdot)$ from observational data $\mathcal{D}$. Interestingly, a result similar to Proposition~\ref{prop:manifold-quasi-oracle-eff} can be obtained by using a \emph{neural representation} $\hat{\Phi}$ that minimizes a plug-in MSE in Eq.~\eqref{eq:repr-plug-in}. 
 
\begin{prop}[Smoothness of the hidden layers] \label{prop:smoothness}
    We denote the trained representation network as $\hat{\mu}_a^x = \hat{h}_a \circ \hat{\Phi} = \argmin \hat{\mathcal{L}}_{\mathit{\Phi}}(h_a\circ{\Phi})$. Then, under mild conditions on the representation network, there exists a hidden layer $V = \hat{f}(X)$ where the regression target becomes smoother:
        $s^{v}_{\mu_a} \ge s^x_{\mu_a} \text{ and } L^{v}_{\mu_a} \le L^x_{\mu_a}$.
\end{prop}
\vspace{-0.1cm}

We illustrate Proposition~\ref{prop:smoothness} in Fig.~\ref{fig:holder-smooth}. Importantly, Proposition~\ref{prop:smoothness} ensures that, with the plug-in loss, we can obtain a representation $\hat{\Phi}$ that simplifies learning $\mu_a^x$ (analogously to how Proposition~\ref{prop:manifold-quasi-oracle-eff} follows from  Assumption~\ref{ass:manifold}) and, thus, \emph{can serve as a substitute for the ideal} $\Phi^*$. Yet, as mentioned previously, the plug-in loss is \emph{not} Neyman-orthogonal and is thus sub-optimal for learning the causal quantities $\xi_a^x/\tau^x$. Hence, by trying to debias the prediction based on $\hat{\Phi}$, we yield the \ORlearners. In this way, we provide efficient learners of the representation-level causal quantities, namely $\tau^{\hat{\phi}} / \xi^{\hat{\phi}}_a$. 

\textbf{Alternative debiasing strategies.} We might wonder whether we can use the Neyman-orthogonal losses differently. For example, we can \textbf{(a)}~learn $\hat{\Phi}$ better (\eg, by using a Neyman-orthogonal loss in Eq.~\eqref{eq:NO-loss} with $g = h_a\circ \Phi$ and $V = X$), or \textbf{(b)}~use the output of the representation network (the smoothest hidden layer according to Proposition~\ref{prop:smoothness}) as the input for the target model ($V = (\hat{h}_0 (\hat{\Phi}(X)), \hat{h}_1(\hat{\Phi}(X)))$).

\textbf{(a) Re-learning the representation.} As we will see later in our experiments, using the learned representation $\hat{\Phi}$ (as suggested by the \ORlearners) \emph{is more effective than learning the representation from scratch by using the Neyman-orthogonal losses} from Eq.~\eqref{eq:NO-loss}. A possible reason for this is that Neyman-orthogonal losses (\eg, DR-learners) have larger variance and, thus, may fail to learn a low-dimensional representation well. 

\textbf{(b) Usage of the outputs.} As another extreme, we can use the outputs of the representation network $V = (\hat{h}_0 (\hat{\Phi}(X)), \hat{h}_1(\hat{\Phi}(X)) ) \in \mathbb{R}^2$ to fit the target model. Yet, in this case, the debiasing with the Neyman-orthogonal losses only \emph{calibrates} the outputs of the representation network $\hat{h}_a \circ \hat{\Phi}$ and cannot compensate for larger errors in learning $\mu_a^x$.

\vspace{-0.1cm}
\begin{tcolorbox}[colback=gray!5!white,colframe=black!75!black,boxsep=-1.5mm,left=10pt,right=10pt]
    \textbf{Guidelines from \textbf{RQ} \circledred{1}.} We suggest using the \ORlearners as they instrumentalize the core Assumption~\ref{ass:manifold}: Under it, the representation-based Neyman-orthogonal learners outperform the standard Neyman-orthogonal learners. Furthermore, \ORlearners offer a middle-ground solution between \textbf{(a)}~the full re-training of the representation network at the second-stage and \textbf{(b)}~debiasing only the representation network outputs.  
\end{tcolorbox}
\vspace{-0.1cm}

\vspace{-0.1cm}
\subsection[RQ 2: OR-learners vs. balancing constraint]{RQ \circledred{2}: Can a Balancing Constraint Substitute Neyman-orthogonality?} \label{sec:rq2}
\vspace{-0.2cm}

\textbf{Balancing constraint.} A balancing constraint was introduced in \citep{johansson2016learning,shalit2017estimating,johansson2022generalization} to reduce finite-sample estimation variance for the end-to-end representation learning methods. It then modifies the plug-in loss of Eq.~\eqref{eq:repr-plug-in}:
\vspace{-0.1cm}
\begin{equation} \label{eq:balancing}
    \hat{\mathcal{L}}_{\operatorname{Bal}(\mathit{\Phi})}(h_a\circ{\Phi}) = \hat{\mathcal{L}}_{\mathit{\Phi}}(h_a\circ{\Phi}) + \alpha \, \hat{\mathcal{L}}_{\text{Bal}}({\Phi}),
\end{equation}
where $\alpha \ge 0$ is a balancing strength, and $\hat{\mathcal{L}}_{\text{Bal}}({\Phi}) = \widehat{\operatorname{dist}}(\mathbb{P}(\Phi(X) \mid A=0), \mathbb{P}(\Phi(X) \mid A=1))$ is an empirical probability metric (\eg, Wasserstein metric (WM) or maximum mean discrepancy (MMD)). The main intuition behind balancing is that it tries to construct a representation space $\mathit{\hat{\Phi}}$ in which both treatments are equally probable, namely, $\pi^{\hat{\phi}}_0 (\phi) = \pi^{\hat{\phi}}_1(\phi)$. Yet, as we will show in the following, this strategy generally harms the CAPOs/CATE estimation.

\textbf{Representation-induced confounding bias (RICB).} As discovered in \citep{johansson2019support,melnychuk2024bounds}, setting $\alpha$ too high might lead to the RICB. In this case, the learned representation \emph{stops being asymptotically valid} (= it does not contain the sufficient information to adjust for the covariates $X$):
\vspace{-0.1cm}
\begin{align}
    \xi^{\hat{\phi}}_a(\phi) \neq \mu^{\hat{\phi}}_a(\phi) \text{ and } \tau^{\hat{\phi}}(\phi) \neq \mu^{\hat{\phi}}_1(\phi) - \mu^{\hat{\phi}}_0(\phi),
\end{align}
where $\hat{\Phi}$ is learned with the population version of Eq.~\ref{eq:balancing}.
As a simple demonstration, we consider $\hat{\Phi}(x) = \text{const}$ that minimizes the loss in Eq.~\eqref{eq:balancing} when $\alpha \to \infty$. In this case, $\xi^{\hat{\phi}}_a(\phi)$ is the average potential outcome (APO), while $\mu^{\hat{\phi}}_a(\phi)$ is the mean outcome $\mathbb{E}(Y \mid A=a)$ (analogously $\tau^{\hat{\phi}}$ is the ATE, and $\mu^{\hat{\phi}}_1(\phi) - \mu^{\hat{\phi}}_0(\phi)$ is the difference in means).

\textbf{Addressing RICB with OR-learners.} Even if the learned representation $\hat{\Phi}$ contains the RICB, the \ORlearners still yield the quasi-oracle efficient estimator of the representation-level causal quantities (see Lemma~\ref{lemma:quasi-oracle-eff}), as they have access to the \emph{unconstrained} nuisance function estimators. Specifically, if we use DR-learners, we yield the augmented IPTW (A-IPTW) estimators of the APOs/ATE; and if we employ R-/IVW-learners, we get the A-IPTW estimators of the overlap-weighted ATE.  

\textbf{Omitting RICB with invertibility.} As a remedy for the RICB, originally suggested by \citep{johansson2022generalization}, one can use \emph{invertible representations} (\eg, $\Phi(X)$ can be implemented as a normalizing flow \citep{rezende2015variational}). In this case, we face a trade-off: we do not allow for the RICB, but also cannot benefit from Assumption~\ref{ass:manifold}.

\textbf{\textbf{RQ} \circledred{2}.} A central question arises: Did it make sense to use the balancing constraint in the first place? In the following, we demonstrate that the balancing constraint relies on the \emph{additional inductive bias}: low-overlap regions of the covariate space exhibit low CAPOs/CATE heterogeneity. To see that, we state two propositions. 

First, as a consequence of Proposition~\ref{prop:smoothness}, the representation subnetwork acts as an expanding mapping. 

\begin{prop}[Smoothing via expanding mapping] \label{prop:expand}
    Assume that the trained representation network $\hat{\Phi}$ minimizes $\hat{\mathcal{L}}_{\mathit{\Phi}}(h_a \circ \Phi)$ and is $s^{\hat{\Phi}}$-H\"older smooth ($s^{\hat{\Phi}} \ge 1$). Then, under mild conditions on $\hat{\mu}_a^x =\hat{h}_a \circ \hat{\Phi}$ and $\mu_a^x$, (1) $\hat{\Phi}$ is an expanding mapping, namely, $\operatorname{Lip}({\hat{\Phi}}) \ge 1$. 
\end{prop}

On the other hand, by trying to enforce the balancing constraint, we actually fit a \emph{contracting mapping}.

\begin{prop}[Balancing via contracting mapping] \label{prop:contract}
    Assume that the trained representation network $\hat{\Phi}$ minimizes $\hat{\mathcal{L}}_\text{\emph{Bal}}(\Phi)$ with WM / MMD and is $s^{\hat{\Phi}}$-H\"older smooth ($s^{\hat{\Phi}} \ge 1$). Then, under mild conditions on $\hat{\Phi}$, (1)~$\hat{\Phi}$ is a contracting mapping, namely $\operatorname{Lip}(\hat{\Phi}) \le 1$. Furthermore, if an analogue of Assumption~\ref{ass:manifold} holds for $\hat{\Phi}$ with a pullback $\hat{J}$ (\eg, $\hat{\Phi}$ is smoothly invertible), (2)~the pullback map is expanding, namely, $\operatorname{Lip}(\hat{J}) \ge 1$.
\end{prop}

\textbf{Interpretation.} Hence, by minimizing the joint loss from Eq.~\eqref{eq:balancing}, two things happen simultaneously. On the one hand, the plug-in loss $\hat{\mathcal{L}}_{\mathit{\Phi}}(h_a \circ \Phi)$ aims to expand the regions of the covariate space where $\mu_a^x$ (and thus CAPOs/CATE) are heterogeneous (to make the regression surface smoother). On the other hand, the balancing loss $\hat{\mathcal{L}}_\text{{Bal}}(\Phi)$ contracts the low-overlap regions of the covariate space (to minimize an empirical probability metric). Those considerations bring us to the following inductive bias.

\textbf{``Low overlap -- low heterogeneity'' inductive bias.} For the joint loss in Eq.~\eqref{eq:balancing} to perform well, we implicitly require that the regions of the covariate space with low CAPOs/CATE heterogeneity to \emph{coincide} with the low-overlap regions. For example, instrumental variables $X_I\subseteq X$ induce no heterogeneity in CAPOs/CATE and, at the same time, create the low-overlap regions.

\textbf{Inductive bias and OR-learners.} The \ORlearners, on the other hand, \emph{do not require such an inductive bias to perform well}: they consider the low-overlap regions as \emph{inherently uncertain}. In contrast, the DR-learners scale up the MSE risk in the low-overlap regions (as they rely on the IPTW weights); and R-/IVW-learners de-emphasize those regions and only fit CATE well in the overlapping parts of $\mathcal{X}$. Therefore, only \emph{in special finite-sample cases} when the inductive bias is true, the end-to-end methods with the balancing constraint might yield a better estimator of CAPOs/CATE than the \ORlearners. Yet, \emph{in general}, the \ORlearners are \emph{asymptotically optimal} due to the Neyman-orthogonality.    
We sum up our findings in Fig.~\ref{fig:or-learner-summary} of Appendix~\ref{app:visual-summary}.

\vspace{-0.1cm}
\begin{tcolorbox}[colback=gray!5!white,colframe=black!75!black,boxsep=-1.5mm,left=10pt,right=10pt]
    \textbf{Guidelines from \textbf{RQ} \circledred{2}}. The balancing constraint relies on the strong inductive bias that the low-overlap regions of the covariate space coincide with the low CAPOs/CATE heterogeneity. 
    The \ORlearners, on the other hand, do not make such an assumption and provide general asymptotic optimality guarantees.
\end{tcolorbox}
\vspace{-0.1cm}

\vspace{-0.1cm}
\section{EXPERIMENTS}
\vspace{-0.3cm}

The primary aim of our numerical experiments is \emph{\textbf{not}} standard benchmarking but \emph{to validate the insights from} \textbf{RQ} \circledred{1} and \textbf{RQ} \circledred{2}, that is, when to use the OR-learners instead of the standard Neyman-orthogonal learners or instead of representations with balancing constraint. 

\textbf{Setup.} We follow prior literature \citep{curth2021nonparametric,melnychuk2024bounds} and use several \mbox{(semi-)synthetic} datasets where both counterfactual outcomes $Y[0]$ and $Y[1]$ and ground-truth covariate-level CAPOs / CATE are available. We perform experiments in two settings that correspond to each research question. $\bullet$\,In \textbf{Setting} \circledred{1}, we compare different \ORlearners based on different target model inputs (\ie, original covariates, pre-trained representations, or the outputs of the pre-trained representation network). $\bullet$\,In \textbf{Setting} \circledred{2}, we show when the \ORlearners improve the representation networks trained with the balancing constraint.

\begin{table}[tb]
    \centering
    \centering
    \vspace{-0.3cm}
    \caption{\textbf{Results for 77 ACIC 2016 datasets in Setting \protect\circledred{1}.} Reported: the  $\%$  of runs, where the \ORlearners improve over plug-in representation networks wrt. out-of-sample rMSE / rPEHE. Here, $d_{\hat{\phi}} = 8$.} \label{tab:acic2016-setting-a}
      \vspace{-0.5cm}
      \begin{center}
        \scriptsize
        \scalebox{0.63}{\begin{tabu}{lr|cc|cc|ccc}
\toprule
 &  & $\text{DR}_0^{\text{K}}$ & $\text{DR}_0^{\text{FS}}$ & $\text{DR}_1^{\text{K}}$ & $\text{DR}_1^{\text{FS}}$ & $\text{DR}^{\text{K}}$ & $\text{R}$ & $\text{IVW}$ \\
\midrule
\multirow{4}{*}{TARNet} & $V = (\hat{\mu}^x_0,\hat{\mu}^x_1)$ & 22.3$\%$ & 20.9$\%$ & 27.6$\%$ & 25.5$\%$ & 37.4$\%$ & 37.1$\%$ & 37.4$\%$ \\
 & $V = X$ & 25.0$\%$ & 20.4$\%$ & 23.5$\%$ & 13.2$\%$ & 19.3$\%$ & 6.8$\%$ & 15.3$\%$ \\
 & $V = X^*$ & 27.0$\%$ & 28.7$\%$ & 26.0$\%$ & 23.4$\%$ & 13.2$\%$ & 6.2$\%$ & 10.8$\%$ \\
 & $V = \hat{\Phi}(X)$ & \textcolor{ForestGreen}{64.7$\%$} & \textcolor{ForestGreen}{60.3$\%$} & \textcolor{ForestGreen}{69.0$\%$} & \textcolor{ForestGreen}{57.9$\%$} & \textcolor{ForestGreen}{68.6$\%$} & \textcolor{ForestGreen}{69.1$\%$} & \textcolor{ForestGreen}{67.4$\%$} \\
\cmidrule(lr){1-9}
\multirow{4}{*}{BNN ($\alpha$ = 0.0)} & $V = (\hat{\mu}^x_0,\hat{\mu}^x_1)$ & 40.9$\%$ & 41.1$\%$ & 40.7$\%$ & 42.1$\%$ & 45.4$\%$ & 45.8$\%$ & 44.6$\%$ \\
 & $V = X$ & 38.2$\%$ & 37.6$\%$ & 33.5$\%$ & 29.6$\%$ & 24.4$\%$ & 8.7$\%$ & 19.6$\%$ \\
 & $V = X^*$ & 40.5$\%$ & 50.0$\%$ & 34.6$\%$ & 39.6$\%$ & 13.8$\%$ & 7.7$\%$ & 10.9$\%$ \\
 & $V = \hat{\Phi}(X)$ & \textcolor{ForestGreen}{70.6$\%$} & \textcolor{ForestGreen}{70.6$\%$} & \textcolor{ForestGreen}{68.6$\%$} & \textcolor{ForestGreen}{73.4$\%$} & \textcolor{ForestGreen}{84.2$\%$} & \textcolor{ForestGreen}{79.4$\%$} & \textcolor{ForestGreen}{82.5$\%$} \\
\bottomrule
\multicolumn{8}{l}{Higher $=$ better. Improvement over the baseline in more than 50$\%$ of runs marked in \textcolor{ForestGreen}{green}}
\end{tabu}
}
    \end{center}
    \vspace{-0.8cm}
\end{table}

\begin{table*}[tb]
    \centering
    \centering
    \vspace{-0.3cm}
    \caption{\textbf{Results for HC-MNIST experiments in Setting \protect\circledred{1}.} Reported: improvements of the \ORlearners over plug-in representation networks wrt. out-of-sample rMSE / rPEHE; mean $\pm$ std over 10 runs. Here, $d_{\hat{\phi}} = 78$.} \label{tab:hcmnist-setting-a}
      \vspace{-0.5cm}
      \begin{center}
        \scriptsize
        \scalebox{0.8}{\begin{tabu}{lr|cc|cc|ccc}
\toprule
 &  & $\text{DR}_0^{\text{K}}$ & $\text{DR}_0^{\text{FS}}$ & $\text{DR}_1^{\text{K}}$ & $\text{DR}_1^{\text{FS}}$ & $\text{DR}^{\text{K}}$ & $\text{R}$ & $\text{IVW}$ \\
\midrule
\multirow{4}{*}{TARNet} & $V = (\hat{\mu}^x_0,\hat{\mu}^x_1)$ & \textcolor{BrickRed}{$+$0.549 $\pm$ 0.006} & \textcolor{BrickRed}{$+$0.564 $\pm$ 0.006} & \textcolor{BrickRed}{$+$0.589 $\pm$ 0.003} & \textcolor{BrickRed}{$+$0.589 $\pm$ 0.003} & \textcolor{BrickRed}{$+$0.509 $\pm$ 0.004} & \textcolor{BrickRed}{$+$0.510 $\pm$ 0.004} & \textcolor{BrickRed}{$+$0.509 $\pm$ 0.004} \\
 & $V = X$ & \textcolor{BrickRed}{$+$0.011 $\pm$ 0.006} & \textcolor{BrickRed}{$+$0.082 $\pm$ 0.065} & \textcolor{BrickRed}{$+$0.017 $\pm$ 0.005} & \textcolor{BrickRed}{$+$0.011 $\pm$ 0.005} & $+$0.002 $\pm$ 0.007 & \textcolor{BrickRed}{$+$0.215 $\pm$ 0.247} & $+$0.004 $\pm$ 0.008 \\
 & $V = X^*$ & \textcolor{BrickRed}{$+$0.033 $\pm$ 0.009} & $-$0.001 $\pm$ 0.007 & \textcolor{BrickRed}{$+$0.052 $\pm$ 0.014} & \textcolor{ForestGreen}{$-$0.017 $\pm$ 0.003} & \textcolor{BrickRed}{$+$0.063 $\pm$ 0.012} & \textcolor{BrickRed}{$+$0.129 $\pm$ 0.179} & \textcolor{BrickRed}{$+$0.052 $\pm$ 0.005} \\
 & $V = \hat{\Phi}(X)$ & \textcolor{ForestGreen}{$-$0.011 $\pm$ 0.004} & $+$0.007 $\pm$ 0.053 & \textcolor{ForestGreen}{$-$0.014 $\pm$ 0.002} & \textcolor{ForestGreen}{$-$0.014 $\pm$ 0.006} & \textcolor{ForestGreen}{$-$0.017 $\pm$ 0.005} & \textcolor{ForestGreen}{$-$0.014 $\pm$ 0.020} & \textcolor{ForestGreen}{$-$0.016 $\pm$ 0.005} \\
\cmidrule(lr){1-9}
\multirow{4}{*}{BNN ($\alpha$ = 0.0)} & $V = (\hat{\mu}^x_0,\hat{\mu}^x_1)$ & $-$0.004 $\pm$ 0.015 & $\pm$0.000 $\pm$ 0.017 & \textcolor{ForestGreen}{$-$0.013 $\pm$ 0.014} & \textcolor{ForestGreen}{$-$0.014 $\pm$ 0.011} & $+$0.001 $\pm$ 0.010 & $-$0.002 $\pm$ 0.008 & $-$0.002 $\pm$ 0.009 \\
 & $V = X$ & $+$0.013 $\pm$ 0.028 & \textcolor{BrickRed}{$+$0.054 $\pm$ 0.043} & $+$0.005 $\pm$ 0.021 & $-$0.012 $\pm$ 0.025 & \textcolor{BrickRed}{$+$0.021 $\pm$ 0.025} & \textcolor{BrickRed}{$+$0.121 $\pm$ 0.102} & \textcolor{BrickRed}{$+$0.025 $\pm$ 0.031} \\
 & $V = X^*$ & \textcolor{BrickRed}{$+$0.040 $\pm$ 0.056} & $-$0.006 $\pm$ 0.037 & \textcolor{BrickRed}{$+$0.048 $\pm$ 0.043} & \textcolor{ForestGreen}{$-$0.039 $\pm$ 0.022} & \textcolor{BrickRed}{$+$0.087 $\pm$ 0.032} & \textcolor{BrickRed}{$+$0.075 $\pm$ 0.056} & \textcolor{BrickRed}{$+$0.096 $\pm$ 0.040} \\
 & $V = \hat{\Phi}(X)$ & \textcolor{ForestGreen}{$-$0.019 $\pm$ 0.019} & \textcolor{ForestGreen}{$-$0.029 $\pm$ 0.022} & \textcolor{ForestGreen}{$-$0.034 $\pm$ 0.019} & \textcolor{ForestGreen}{$-$0.040 $\pm$ 0.023} & \textcolor{ForestGreen}{$-$0.020 $\pm$ 0.020} & \textcolor{ForestGreen}{$-$0.027 $\pm$ 0.020} & \textcolor{ForestGreen}{$-$0.022 $\pm$ 0.021} \\
\bottomrule
\multicolumn{8}{l}{Lower $=$ better. Significant improvement over the baseline in \textcolor{ForestGreen}{green}, significant worsening of the baseline in \textcolor{BrickRed}{red}}
\end{tabu}
}
    \end{center}
    \vspace{-0.5cm}
\end{table*}

\begin{table}[t]
    \vspace{-0.2cm}
    \begin{minipage}{\linewidth}
      \caption{\textbf{Results for IHDP experiments in Setting \protect\circledred{1}.} Reported: out-of-sample rMSE / rPEHE for different causal quantities ($\xi^x_a / \tau^x$, respectively), median $\pm$ std over 100 train/test splits. Here, $d_{\hat{\phi}} = 12$ for neural baselines.} \label{tab:ihdp-setting-a-full}
      \vspace{-0.5cm}
      \begin{center}
            \scriptsize
            \scalebox{0.74}{\begin{tabu}{lr|c|c|c}
\toprule
 &  & $\xi_0^x$ & $\xi_1^x$ & $\tau^x$ \\
\midrule
\multirow{3}{*}{XGBoost S} & $\text{DR}^{\text{K}}$ & 0.496 $\pm$ 0.118 & 0.723 $\pm$ 0.241 & 0.826 $\pm$ 0.239 \\
 & $\text{R}$ & --- & --- & 1.631 $\pm$ 0.260 \\
 & $\text{IVW}$ & --- & --- & 0.820 $\pm$ 0.247 \\
\cmidrule(lr){1-5}
\multirow{3}{*}{XGBoost T} & $\text{DR}^{\text{K}}$ & 0.501 $\pm$ 0.117 & 0.531 $\pm$ 0.283 & 0.754 $\pm$ 0.240 \\
 & $\text{R}$ & --- & --- & 1.762 $\pm$ 0.258 \\
 & $\text{IVW}$ & --- & --- & 0.749 $\pm$ 0.242 \\
\cmidrule(lr){1-5}
\multirow{5}{*}{TARNet} & Plug-in & 0.367 $\pm$ 0.160 & \textbf{0.379 $\pm$ 0.226} & \textbf{0.518 $\pm$ 0.270} \\
 & $\text{DR}^{\text{K}}$ & \underline{0.366 $\pm$ 0.168} & 0.390 $\pm$ 0.228 & \underline{0.523 $\pm$ 0.280} \\
 & $\text{DR}^{\text{FS}}$ & \textbf{0.364 $\pm$ 0.169} & 0.421 $\pm$ 0.247 & --- \\
 & $\text{R}$ & --- & --- & 0.563 $\pm$ 0.295 \\
 & $\text{IVW}$ & --- & --- & 0.530 $\pm$ 0.285 \\
\cmidrule(lr){1-5}
\multirow{5}{*}{BNN ($\alpha$ = 0.0)} & Plug-in & 0.386 $\pm$ 0.157 & 0.478 $\pm$ 0.140 & 0.595 $\pm$ 0.168 \\
 & $\text{DR}^{\text{K}}$ & 0.379 $\pm$ 0.168 & 0.465 $\pm$ 0.160 & 0.568 $\pm$ 0.180 \\
 & $\text{DR}^{\text{FS}}$ & 0.376 $\pm$ 0.169 & \underline{0.382 $\pm$ 0.213} & --- \\
 & $\text{R}$ & --- & --- & 0.543 $\pm$ 0.198 \\
 & $\text{IVW}$ & --- & --- & 0.568 $\pm$ 0.186 \\
\cmidrule(lr){1-5}
Oracle &  & 0.303 & 0.315 & 0.434 \\
\bottomrule
\multicolumn{4}{l}{Lower $=$ better. Best in \textbf{bold}, second best \underline{underlined} }
\end{tabu}
}
        \end{center}
    \end{minipage}%
    \vspace{-0.6cm}
\end{table}

\textbf{Datasets.} We used three standard datasets for benchmarking in causal inference: (1)~a fully-synthetic dataset ($d_x = 2$) \citep{kallus2019interval,melnychuk2024bounds}; (2)~the IHDP dataset ($n = 747; d_x=25$) \citep{hill2011bayesian,shalit2017estimating}; (3)~a collection of 77 ACIC 2016 datasets ($n = 4802, d_x=82$) \citep{dorie2019automated}; and (4)~a high-dimensional HC-MNIST dataset \citep{jesson2021quantifying} ($n = 70000, d_x=785$). 
Details are in Appendix~\ref{app:datasets}. All the datasets (1)-(4) then help us to empirically study \textbf{RQ} \circledred{1} and \textbf{RQ} \circledred{2}. Specifically, for \textbf{RQ} \circledred{1}, we know a priori that the manifold hypothesis \emph{does not hold} for (1) the fully-synthetic dataset; \emph{is believed to hold} for (3) ACIC 2016 datasets;  and \emph{definitely holds} for (2) the IHDP dataset (\ie, CAPOs/CATE are defined on the linear combinations of the covariates) and for (4) the HC-MNIST dataset (\ie, CAPOs/CATE depend on a two-dimensional latent manifold that encodes an image’s mean pixel intensity and digit label). Furthermore, for \textbf{RQ} \circledred{2}, we know that the “low overlap – low heterogeneity” inductive bias can only be assumed for (2) the IHDP dataset.

\textbf{Performance metrics.} We report (i)~the out-of-sample root mean squared error (rMSE) and (ii)~the root precision in estimating heterogeneous effect (rPEHE) for CAPOs and CATE, respectively. Recall that, in \textbf{RQ} \circledred{1} and \textbf{RQ} \circledred{2}, we are primarily interested in how the \ORlearners improve the existing methods, and, therefore, we report the difference in the performance between the baseline representation learning method and the \ORlearners. Formally, we compute $\Delta(\text{rMSE})$ for CAPOs and $\Delta(\text{rPEHE})$ for CATE for different variants of the \ORlearners: DR-learner in the style of \citet{kennedy2023towards} (\textbf{DR$_a^\text{K}$}) and DR-learner in the style of \citet{foster2023orthogonal} (\textbf{DR$_a^\text{FS}$}) for CAPOs; and DR-learner \citep{kennedy2023towards} (\textbf{DR$^\text{K}$}) / R-learner \citep{nie2021quasi} (\textbf{R}) / IVW-learner \citep{fisher2024inverse} (\textbf{IVW}) for CATE. Furthermore, we followed best benchmarking practices for CAPOs/CATE estimation \citep{curth2021really}. Namely, we often report robust performance metrics (\ie, medians/percentage of best runs) and always compare baselines with a similar structure of the nuisance functions (\eg, S-learners vs. S-learner-based DR/R-learners).

\textbf{Baselines.} We implemented various state-of-the-art representation learning methods and combine each baseline with the \ORlearners (see Appendix~\ref{app:implementation}):
\textbf{TARNet} \citep{shalit2017estimating}; several variants of \textbf{BNN} \citep{johansson2016learning} (w/ or w/o balancing); several variants of \textbf{CFR} \citep{shalit2017estimating,johansson2022generalization} (w/ balancing, non-/ invertible); several variants of \textbf{RCFR} \citep{johansson2018learning,johansson2022generalization} (different types of balancing); several variants of \textbf{CFR-ISW} \citep{hassanpour2019counterfactual} (w/ or w/o balancing, non-/ invertible); and \textbf{BWCFR} \citep{assaad2021counterfactual} (w/ or w/o balancing, non-/invertible).

\begin{figure}[t]
    \centering
    \vspace{-0.4cm}
    \includegraphics[width=\linewidth]{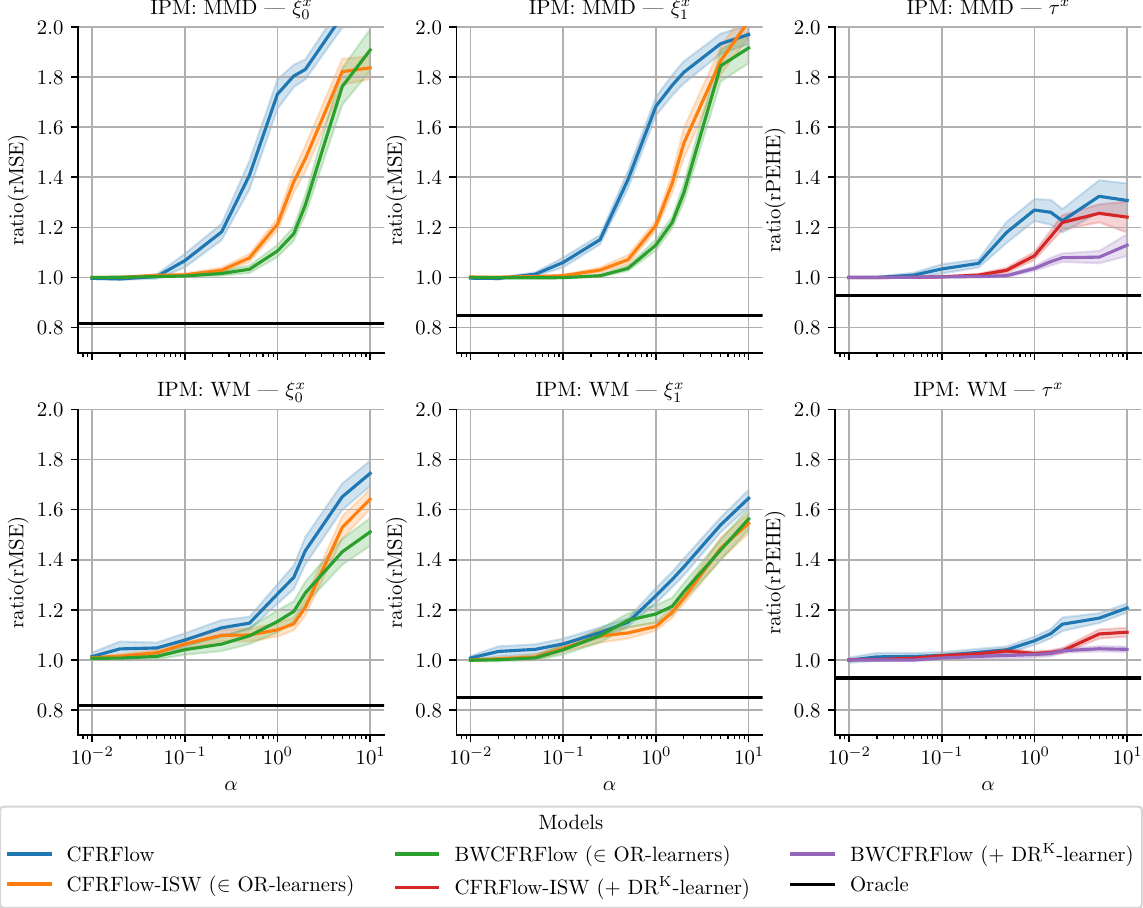}
    \vspace{-0.7cm}
    \caption{\textbf{Results for synthetic data in Setting \protect\circledred{2}.} Reported: ratio between the performance of TARFlow (CFRFlow with $\alpha = 0$) and invertible representation networks with varying $\alpha$; mean $\pm$ SE over 15 runs. Lower is better. Here: $n_{\text{train}} = 500$, $d_{\hat{\phi}} = 2$.}
    \label{fig:synthetic-setting-b}
    \vspace{-0.7cm}
\end{figure}

{\tiny$\blacksquare$}~\textbf{Setting} \circledred{1}.  In Setting \circledred{1}, we want to confirm our theoretical insights for the manifold hypothesis (Assumption~\ref{ass:manifold}) by comparing the performance of vanilla representation networks (\ie, TARNet and BNN ($\alpha = 0.0$)) versus the \ORlearners applied on top of the learned unconstrained representations, where the latter is denoted $V = \hat{\Phi}(X)$. We compare two further variants of the \ORlearners, where the target network has different inputs: (a)~$V = (\hat{h}_0(\hat{\Phi}(X)), \hat{h}_1(\hat{\Phi}(X))) = (\hat{\mu}^x_0,\hat{\mu}^x_1)$, and (b)~$V = X$, yet the same depth of one hidden layer. We also compare the \ORlearners with (c)~the target network, which matches the depth of the original representation network $V = X^*$. Therefore, (b) and (c) both provide a fair comparison of the \ORlearners and the standard Neyman-orthogonal learners with $V = X$.

\underline{\textbf{Results.}} Tables \ref{tab:acic2016-setting-a} and \ref{tab:hcmnist-setting-a} show the results for the ACIC 2016 datasets and the HC-MNIST dataset, where, due to high-dimensionality, it is reasonable to assume the low-dimensional manifold hypothesis. We find that the \ORlearners outperform the baseline representation learning networks due to their Neyman-orthogonality. Furthermore, the \ORlearners with $V = \hat{\Phi}(X)$ outperform the standard Neyman-orthogonal learners (namely, with $V = X/X^*$) for CAPOs/CATE estimation. This confirms that \emph{using the pre-trained representation for the target model input $V = \hat{\Phi}(X)$ is more effective than training the representation network from scratch at the second stage}. Furthermore, Table~\ref{tab:ihdp-setting-a-full} shows the results for the IHDP dataset. Here, we report the absolute performance of different methods: non-neural Neyman-orthogonal learners instantiated with XGBoost \citep{chen2015xgboost} (with S-/T-learners for the first-stage models); plug-in representation learning methods; and the OR-learners used with the pre-trained representations $V = \hat{\Phi}(X)$. We see that the OR-learners improve over other non-neural Neyman-orthogonal learners: This was expected as \emph{the potential outcomes for the IHDP dataset are defined via the low-dimensional manifold of the covariate space}. 
In Appendix~\ref{app:experiments}, we also provide \textbf{additional results} for (i)~the synthetic dataset (where the low-dimensional manifold hypothesis cannot be assumed) and (ii)~the HC-MNIST dataset (where we compare the \ORlearners with the non-neural Neyman-orthogonal learners).  This confirms our theory: (i)~as expected, both $V = \hat{\Phi}(X)$ and $V = X/X^*$ perform similarly, and (ii)~\ORlearners outperform non-neural learners.

\begin{table}[t]
     \vspace{-0.2cm}
      \caption{\textbf{Results for 77 semi-synthetic ACIC 2016 experiments in Setting \protect\circledred{2}.} Reported: the $\%$ of runs, where the \ORlearners improve over non-invertible plug-in/IPTW representation networks wrt. out-of-sample rMSE / rPEHE. Here, $d_{\hat{\phi}} = 8$.} \label{tab:acic2016-setting-c}
      \vspace{-0.5cm}
      \begin{center}
            \scriptsize
            \scalebox{0.68}{\begin{tabu}{l|cc|cc|ccc}
\toprule
 & $\text{DR}_0^{\text{K}}$ & $\text{DR}_0^{\text{FS}}$ & $\text{DR}_1^{\text{K}}$ & $\text{DR}_1^{\text{FS}}$ & $\text{DR}^{\text{K}}$ & $\text{R}$ & $\text{IVW}$ \\
\midrule
CFR (MMD; $\alpha$ = 0.1) & \textcolor{ForestGreen}{67.6$\%$} & \textcolor{ForestGreen}{60.3$\%$} & \textcolor{ForestGreen}{67.1$\%$} & \textcolor{ForestGreen}{62.9$\%$} & \textcolor{ForestGreen}{72.5$\%$} & \textcolor{ForestGreen}{66.8$\%$} & \textcolor{ForestGreen}{69.0$\%$} \\
CFR (WM; $\alpha$ = 0.1) & \textcolor{ForestGreen}{73.2$\%$} & \textcolor{ForestGreen}{68.3$\%$} & \textcolor{ForestGreen}{72.8$\%$} & \textcolor{ForestGreen}{69.0$\%$} & \textcolor{ForestGreen}{74.6$\%$} & \textcolor{ForestGreen}{74.5$\%$} & \textcolor{ForestGreen}{75.2$\%$} \\
BNN (MMD; $\alpha$ = 0.1) & \textcolor{ForestGreen}{74.0$\%$} & \textcolor{ForestGreen}{79.0$\%$} & \textcolor{ForestGreen}{66.3$\%$} & \textcolor{ForestGreen}{66.4$\%$} & \textcolor{ForestGreen}{70.6$\%$} & \textcolor{ForestGreen}{70.3$\%$} & \textcolor{ForestGreen}{68.7$\%$} \\
BNN (WM; $\alpha$ = 0.1) & \textcolor{ForestGreen}{74.6$\%$} & \textcolor{ForestGreen}{80.5$\%$} & \textcolor{ForestGreen}{70.4$\%$} & \textcolor{ForestGreen}{73.2$\%$} & \textcolor{ForestGreen}{75.8$\%$} & \textcolor{ForestGreen}{77.5$\%$} & \textcolor{ForestGreen}{76.6$\%$} \\
RCFR (MMD; $\alpha$ = 0.1) & \textcolor{ForestGreen}{78.8$\%$} & \textcolor{ForestGreen}{71.6$\%$} & \textcolor{ForestGreen}{75.1$\%$} & \textcolor{ForestGreen}{71.7$\%$} & \textcolor{ForestGreen}{72.4$\%$} & \textcolor{ForestGreen}{70.7$\%$} & \textcolor{ForestGreen}{74.5$\%$} \\
RCFR (WM; $\alpha$ = 0.1) & \textcolor{ForestGreen}{80.2$\%$} & \textcolor{ForestGreen}{75.7$\%$} & \textcolor{ForestGreen}{77.5$\%$} & \textcolor{ForestGreen}{76.3$\%$} & \textcolor{ForestGreen}{71.0$\%$} & \textcolor{ForestGreen}{73.5$\%$} & \textcolor{ForestGreen}{75.3$\%$} \\
CFR-ISW (MMD; $\alpha$ = 0.1) & \textcolor{ForestGreen}{70.8$\%$} & \textcolor{ForestGreen}{63.4$\%$} & \textcolor{ForestGreen}{69.2$\%$} & \textcolor{ForestGreen}{65.7$\%$} & \textcolor{ForestGreen}{67.2$\%$} & \textcolor{ForestGreen}{64.2$\%$} & \textcolor{ForestGreen}{69.8$\%$} \\
CFR-ISW (WM; $\alpha$ = 0.1) & \textcolor{ForestGreen}{76.5$\%$} & \textcolor{ForestGreen}{69.6$\%$} & \textcolor{ForestGreen}{71.6$\%$} & \textcolor{ForestGreen}{71.3$\%$} & \textcolor{ForestGreen}{70.7$\%$} & \textcolor{ForestGreen}{73.7$\%$} & \textcolor{ForestGreen}{77.0$\%$} \\
BWCFR (MMD; $\alpha$ = 0.1) & \textcolor{ForestGreen}{69.5$\%$} & \textcolor{ForestGreen}{66.7$\%$} & \textcolor{ForestGreen}{66.3$\%$} & \textcolor{ForestGreen}{63.9$\%$} & \textcolor{ForestGreen}{70.5$\%$} & \textcolor{ForestGreen}{68.4$\%$} & \textcolor{ForestGreen}{68.7$\%$} \\
BWCFR (WM; $\alpha$ = 0.1) & \textcolor{ForestGreen}{73.4$\%$} & \textcolor{ForestGreen}{73.9$\%$} & \textcolor{ForestGreen}{73.2$\%$} & \textcolor{ForestGreen}{72.7$\%$} & \textcolor{ForestGreen}{71.5$\%$} & \textcolor{ForestGreen}{72.1$\%$} & \textcolor{ForestGreen}{71.7$\%$} \\
\bottomrule
\multicolumn{8}{l}{Higher $=$ better. Improvement over the baseline in more than 50$\%$ of runs marked in \textcolor{ForestGreen}{green}}
\end{tabu}
}
        \end{center}
    \vspace{-0.8cm}
\end{table}

{\tiny$\blacksquare$}~\textbf{Setting} \circledred{2}.  Here, we want to verify our finding that the balancing constraint only helps when the “low overlap – low heterogeneity” inductive bias can be assumed. For that, we study how the \ORlearners compare with the representation networks trained with the balancing constraint and varying amounts of balancing strength $\alpha$. We consider both invertible (TARFlow, CFRFlow, CFRFlow-ISW, and BWCFRFlow) and non-invertible (CFR, BNN, RCFR, CFR-ISW, and BWCFR) representation networks. 

\underline{\textbf{Results.}} Fig.~\ref{fig:synthetic-setting-b} and Table~\ref{tab:acic2016-setting-c} show the results for the synthetic data and the ACIC 2016 datasets. In both cases, the \ORlearners manage to improve the representation networks that use balancing constraints (as, in general, the ``low overlap -- low heterogeneity'' inductive bias cannot be assumed for these datasets). We also refer to Appendix~\ref{app:experiments} for \textbf{additional results} for (i)~the synthetic and (ii)~IHDP datasets. For (i), by varying the size of data $n_{\text{train}} \in \{250, 1000\}$, we show a similar pattern to Fig.~\ref{fig:or-learner-summary} of Appendix~\ref{app:visual-summary}. Also, we visualize the expanding/contracting mappings, suggested by Propositions~\ref{prop:expand}-\ref{prop:contract}. For (ii), the balancing constraint helps the CAPOs/CATE estimation, as the underlying inductive bias is present in the IHDP dataset.

\textbf{Limitations.} Applications of our \ORlearners should follow a cautious approach: We rely on a crucial manifold hypothesis (Assumption~\ref{ass:manifold}) that has to be supported by some \emph{background knowledge} about the ground-truth causal quantity.

\textbf{Takeaways.} Our experiments confirm our theoretical findings for \textbf{RQ} \circledred{1} and \textbf{RQ} \circledred{2}. In general, there is no nuisance-free way to do CATE/CAPOs model selection based solely on the observational data \citep{curth2023search}. However, \circledred{1} given Assumption~\ref{ass:manifold}, one can \emph{simplify} the task of CAPOs/CATE estimation and use the suggested framework of \ORlearners. Similarly, \circledred{2} we advise \emph{against} the balancing constraint, unless one can assume the underlying inductive bias.

\newpage
\textbf{Acknowledgments.} This paper is supported by the DAAD program “Konrad Zuse Schools of Excellence in Artificial Intelligence”, sponsored by the Federal Ministry of Education and Research. S.F. acknowledges funding via Swiss National Science Foundation Grant 186932. This work has been supported by the German Federal Ministry of Education and Research (Grant: 01IS24082).

\bibliography{bibliography}

\section*{Checklist}

\begin{enumerate}

  \item For all models and algorithms presented, check if you include:
  \begin{enumerate}
    \item A clear description of the mathematical setting, assumptions, algorithm, and/or model. [Yes] (see Sec.~\ref{sec:prelim} and Appendix~\ref{app:implementation})
    \item An analysis of the properties and complexity (time, space, sample size) of any algorithm. [Yes] (see Appendix~\ref{app:experiments})
    \item (Optional) Source code, with specification of all dependencies, including external libraries. [Yes] (Code is available at \url{https://github.com/Valentyn1997/OR-learners}.)
  \end{enumerate}

  \item For any theoretical claim, check if you include:
  \begin{enumerate}
    \item Statements of the full set of assumptions of all theoretical results. [Yes] (See Sec.~\ref{sec:prelim} and \ref{sec:OR-learners})
    \item Complete proofs of all theoretical results. [Yes] (See Appendix~\ref{app:proofs})
    \item Clear explanations of any assumptions. [Yes]  (See Sec.~\ref{sec:OR-learners})
  \end{enumerate}

  \item For all figures and tables that present empirical results, check if you include:
  \begin{enumerate}
    \item The code, data, and instructions needed to reproduce the main experimental results (either in the supplemental material or as a URL). [Yes] (Code is available at \url{https://github.com/Valentyn1997/OR-learners}.)
    \item All the training details (e.g., data splits, hyperparameters, how they were chosen). [Yes] (See Appendix~\ref{app:implementation})
    \item A clear definition of the specific measure or statistics and error bars (e.g., with respect to the random seed after running experiments multiple times). [Yes] (In the caption of figures and tables)
    \item A description of the computing infrastructure used. (e.g., type of GPUs, internal cluster, or cloud provider). [Yes] (see Appendix~\ref{app:experiments})
  \end{enumerate}

  \item If you are using existing assets (e.g., code, data, models) or curating/releasing new assets, check if you include:
  \begin{enumerate}
    \item Citations of the creator, if your work uses existing assets. [Yes] (We used publicly available synthetic and semi-synthetic datasets).
    \item The license information of the assets, if applicable. [Not Applicable]
    \item New assets either in the supplemental material or as a URL, if applicable. [Not Applicable]
    \item Information about consent from data providers/curators. [Not Applicable]
    \item Discussion of sensible content if applicable, e.g., personally identifiable information or offensive content. [Not Applicable]
  \end{enumerate}

  \item If you used crowdsourcing or conducted research with human subjects, check if you include:
  \begin{enumerate}
    \item The full text of instructions given to participants and screenshots. [Not Applicable]
    \item Descriptions of potential participant risks, with links to Institutional Review Board (IRB) approvals if applicable. [Not Applicable]
    \item The estimated hourly wage paid to participants and the total amount spent on participant compensation. [Not Applicable]
  \end{enumerate}

\end{enumerate}

\clearpage
\appendix
\thispagestyle{empty}
\onecolumn 
\aistatstitle{Orthogonal Representation Learning for Estimating Causal Quantities: Appendix}

\section{EXTENDED RELATED WORK} \label{app:extended-rw}

Our work aims to unify two streams of work, namely, end-to-end representation learning methods (Sec.~\ref{app:extended-rw-representation-learning}) and two-stage meta-learners (Sec.~\ref{app:extended-rw-orthogonal-learners}). We review both in the following and then discuss the implications for our work.

\subsection{End-to-end Representation Learning Methods} \label{app:extended-rw-representation-learning}

Several methods have been previously introduced for \emph{end-to-end} representation learning of CAPOs/CATE  \citep[see, in particular, the seminal works by][]{johansson2016learning,shalit2017estimating,johansson2022generalization}. Existing methods fall into three main streams: (1)~One can fit an \emph{unconstrained shared representation} to directly estimate both potential outcome surfaces \citep[e.g., \textbf{TARNet};][]{shalit2017estimating}. (2)~Some methods additionally enforce a \emph{balancing constraint based on empirical probability metrics}, so that the distributions of the treated and untreated representations become similar \citep[e.g., \textbf{CFR} and \textbf{BNN};][]{johansson2016learning,shalit2017estimating}. Importantly, the balancing constraint is only guaranteed to perform a consistent estimation for \emph{invertible} representations since, otherwise, balancing leads to a \emph{representation-induced confounding bias} (RICB) \citep{johansson2019support,melnychuk2024bounds}. Finally, (3)~one can additionally perform \emph{balancing by re-weighting} the loss and the distributions of the representations with learnable weights \citep[e.g., \textbf{RCFR};][]{johansson2022generalization}.

\begin{table}[t]
    \vspace{-0.2cm}
    \caption{Overview of representation learning methods for CAPOs/CATE estimation. Here, parentheses imply the possibility of an extension.}
    \label{tab:methods-comparison}
    \vspace{-0.2cm}
    \begin{center}
        \vspace{-0.1cm}
        \scalebox{0.95}{
            \scriptsize
            \begin{tabular}{p{3.7cm}|p{1.8cm}|p{1.6cm}|p{1.4cm}|>{\centering\arraybackslash}p{2cm}|>{\centering\arraybackslash}p{2cm}|>{\centering\arraybackslash}p{2.2cm}}
                \toprule
                \multirow{2}{*}{Method} & \multirow{2}{*}{\hspace{-0.3cm} \begin{tabular}{l}   Learner \\  type \end{tabular}} &  \multirow{2}{*}{\hspace{-0.3cm} \begin{tabular}{l}   Balancing \\  constraint \end{tabular}} & \multirow{2}{*}{Invertibility} & \multirow{2}{*}{\begin{tabular}{c} Consistency \\ of estimation \end{tabular}} & \multicolumn{2}{c}{Neyman-orthogonality}  \\
                 \cmidrule(lr){6-7}& 
                       &  &  &  & CAPOs & CATE  \\
                \midrule
                TARNet \citep{shalit2017estimating, johansson2022generalization} & PI & {--} & {--} & {\cmark} &  {\xmark} & {\xmark}\\
                \midrule
                BNN \citep{johansson2016learning}; CFR \citep{shalit2017estimating, johansson2022generalization}; ESCFR \citep{wang2024optimal}; ORIC \citep{yan2025reducing}  & \multirow{5}{*}{PI} &  \multirow{5}{*}{IPM} & \multirow{5}{*}{(any) / --} & \multirow{5}{*}{\xmark $\,$[\cmark: invertible]} & \multirow{5}{*}{\xmark} & \multirow{5}{*}{\xmark}  \\
                \midrule
                RCFR \citep{johansson2018learning,johansson2022generalization} & \multirow{2}{*}{WPI} & \multirow{2}{*}{IPM + LW} & \multirow{2}{*}{(any) / --} & \multirow{2}{*}{\xmark $\,$[\cmark: invertible]} & \multirow{2}{*}{\xmark} & \multirow{2}{*}{\xmark}  \\
                \midrule
                DACPOL \citep{atan2018counterfactual}; CRN \citep{bica2020estimating}; ABCEI \citep{du2021adversarial}; CT \citep{melnychuk2022causal}; MitNet \citep{guo2023estimating};  BNCDE \citep{hess2024bayesian} & \multirow{6}{*}{PI} &  \multirow{6}{*}{JSD} & \multirow{6}{*}{--} & \multirow{6}{*}{\xmark} & \multirow{6}{*}{\xmark} & \multirow{6}{*}{\xmark}\\
                \midrule
                SITE \citep{yao2018representation}& PI & LS & MPD & \xmark $\,$[\cmark: invertible] & \xmark & \xmark \\
                \midrule
                DragonNet \citep{shi2019adapting} & PI / (DR) & -- & -- & \cmark & (\cmark$^{\text{DR}^\text{K}}$) & (\cmark$^{\text{DR}^{\text{K}}}$)\\
                \midrule
                PM \citep{schwab2018perfect}; StableCFR \citep{wu2023stable} & \multirow{2}{*}{WPI} & \multirow{2}{*}{IPM + UVM} & \multirow{2}{*}{--} & \multirow{2}{*}{\cmark} & \multirow{2}{*}{\xmark} & \multirow{2}{*}{\xmark} \\
                \midrule
                CFR-ISW \citep{hassanpour2019counterfactual}; & \multirow{2}{*}{IPTW} & \multirow{2}{*}{IPM + RP} & \multirow{2}{*}{--} & \multirow{2}{*}{\xmark} & \multirow{2}{*}{\xmark} & \multirow{2}{*}{\xmark} \\
                \midrule
                DR-CFR \citep{hassanpour2019learning}; DeR-CFR \citep{wu2022learning} & \multirow{3}{*}{IPTW} & \multirow{3}{*}{IPM + CP} & \multirow{3}{*}{--}  & \multirow{3}{*}{\cmark} & \multirow{3}{*}{\xmark} & \multirow{3}{*}{\xmark} \\ 
                \midrule
                DKLITE \citep{zhang2020learning} & PI & CV & RL  & \xmark $\,$[\cmark: invertible] & \xmark & \xmark \\
                \midrule
                BWCFR \citep{assaad2021counterfactual} & IPTW & IPM + CP & -- & \cmark & \mbox{\xmark} & \xmark \\
                \midrule
                SNet \citep{curth2021nonparametric,chauhan2023adversarial} & \multirow{3}{*}{DR} & \multirow{3}{*}{--} & \multirow{3}{*}{--} & \multirow{3}{*}{\cmark} & \multirow{3}{*}{(\cmark$^{\text{DR}^\text{K}}$)} & \multirow{3}{*}{\cmark$^{\text{DR}^{\text{K}}}$}\\
                \midrule GWIB \citep{yang2024revisiting} & PI & MI & -- & \xmark & \xmark & \xmark \\
                \midrule CausalEGM \citep{liu2024encoding} & PI & -- & GAN & \cmark & \xmark & \xmark \\
                \midrule
                \textbf{\ORlearners} (our paper) & DR / R / IVW & (any) & NFs / -- & \cmark &\cmark$^{\text{DR}^\text{FS}}$, \cmark$^{\text{DR}^\text{K}}$ & \cmark$^{\text{DR}^{\text{K}}}$, \cmark$^{\text{R}}$, \cmark$^{\text{IVW}}$\\
                \bottomrule
                \multicolumn{7}{p{17.3cm}}{\emph{Legend}:} \\ 
                \multicolumn{7}{p{17.3cm}}{\quad $\bullet$ Learner type: plug-in (PI); weighted plug-in (WPI); inverse probability of treatment weighted (IPTW); doubly-robust (DR); \mbox{\null \quad\quad Robinson's / residualized (R)}} \\
                \multicolumn{7}{p{17.3cm}}{\quad $\bullet$ Balancing: integral probability metric (IPM); learnable weights (LW); Jensen-Shannon divergence (JSD); local similarity (LS); \mbox{\null \quad\quad upsampling via matching (UVM); representation propensity (RP)}; covariate propensity (CP); counterfactual variance (CV);} \\ 
                \multicolumn{7}{p{17.3cm}}{\mbox{\null \quad\quad  mutual information (MI)}} \\
                \multicolumn{7}{p{17.3cm}}{
                \quad $\bullet$ Invertibility: middle point distance (MPD); reconstruction loss (RL); normalizing flows (NFs); GAN-based loss (GAN)}\\
                \multicolumn{7}{p{17.3cm}}{\quad $\bullet$ Neyman-orthogonality: DR-learner in the style of \citet{kennedy2023towards} (${\text{DR}^{\text{K}}}$); DR-learner in the style of 
                \citet{foster2023orthogonal} (${\text{DR}^{\text{FS}}}$)}
            \end{tabular}
        }
    \end{center}
    \vspace{-0.3cm}
\end{table}

Table~\ref{tab:methods-comparison} provides a summary of the main representation learning methods for the estimation of causal quantities. Therein, we showed (1)~how different constraints imposed on the representations relate to the consistency of estimation and (2)~Neyman-orthogonality of the underlying methods. We highlight several important constrained representations below and discuss the implications for estimating causal quantities.

\textbf{Balancing constraint and invertibility.} Following CFR and BNN, several works proposed alternative strategies for implementing the balancing constraints, \eg, based on adversarial learning \citep{atan2018counterfactual,curth2021inductive,du2021adversarial,melnychuk2022causal,guo2023estimating}; metric learning \citep{yao2018representation}; counterfactual variance minimization \citep{zhang2020learning}; and empirical mutual information \citep{yang2024revisiting}. To enforce \emph{invertibility} (and, thus, consistency of estimation), several works suggested metric learning heuristics \citep{yao2018representation} or reconstruction loss \citep{zhang2020learning}.  

\textbf{Balancing by re-weighting.}  Other methods extended \emph{balancing by re-weighting}, as in RCFR but, for example, with weights based on matching \citep{schwab2018perfect,wu2023stable}; or with inverse probability of treatment weights (IPTW) \citep{hassanpour2019counterfactual,hassanpour2019learning,assaad2021counterfactual,wu2022learning}.  Importantly, balancing by re-weighting on itself does not harm the consistency of the estimation and only changes the type of the underlying meta-learner (\ie, weighted plug-in or IPTW).

\begin{figure}[ht]
    \vspace{-0.5cm}
    \centering
    \includegraphics[width=0.9\textwidth]{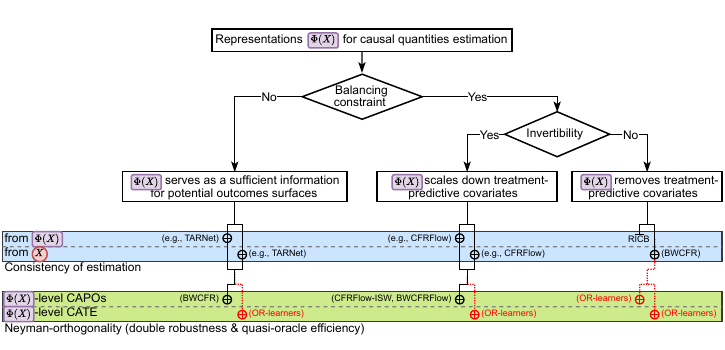}
    \vspace{-0.4cm}
    \caption{Flow chart of consistency and Neyman-orthogonality for representation learning methods. The \ORlearners fill the gaps shown by \textcolor{red}{red dotted lines}.}
    \label{fig:rw-consistentcy-orthogonality}
    \vspace{-0.1cm}
\end{figure}

\textbf{Validity of representations for consistent estimation.} As mentioned previously, balancing representations with empirical probability metrics without strictly enforcing invertibility generally leads to \emph{inconsistent estimation based on representations}. This issue was termed as a \emph{representation-induced adaptation error} \citep{johansson2019support} in the context of unsupervised domain adaptation and as a \emph{representation-induced confounding bias (RICB)} \citep{melnychuk2024bounds} in the context of estimation of causal quantities. More generally, the RICB can be recognized as a type of runtime confounding \citep{coston2020counterfactual}, \ie, when only a subset of covariates is available for the estimation of the causal quantities. Several works offered solutions to circumvent the RICB and achieve consistency. For example, \citet{assaad2021counterfactual} employed IPTW based on original covariates, and \citet{melnychuk2024bounds} used a sensitivity model to perform a partial identification. However, to the best of our knowledge, \underline{no} Neyman-orthogonal method was proposed to resolve the RICB (see Fig.~\ref{fig:rw-consistentcy-orthogonality}).  

\textbf{Balancing and finite-sample generalization error.} In the original works on balancing representations \citep{shalit2017estimating,johansson2022generalization}, the authors provided finite-sample generalization error bounds for any estimator of CAPOs/CATE based on a factual estimation error and a distributional distance between treated and untreated populations. Therein, the authors employed integral probability metrics as the distributional distance. These bounds were further improved with other distributional distances, \eg, counterfactual variance \citep{zhang2020learning}, $\chi^2$-divergence \citep{csillag2024generalization}, and KL-divergence \citep{huang2024unveiling}. However, the work by \citep{shalit2017estimating,johansson2022generalization} suggests that the large distributional distance only \emph{acknowledges the lack of overlap between treated and untreated covariates} (and, hence, the hardness of the estimation) but it \emph{does not instruct how much balancing needs to be applied}. Moreover, the finite-sample generalization error bounds \emph{do not instruct} how to design Neyman-orthogonal learners, and, thus, they are not relevant to our work.

{\textbf{Note on non-neural representations.} Multiple works also explored the use of non-neural representations for the estimation of causal quantities, also known under the umbrella term of \emph{scores}. Examples include propensity/balancing scores \citep{rosenbaum1983central,antonelli2018doubly}, prognostic scores \citep{hansen2008prognostic,huang2017joint,luo2020matching,antonelli2018doubly,d2021deconfounding}, and deconfounding scores \citep{d2021deconfounding}. However, we want to highlight that these works focus on \underline{different}, rather simpler than our settings:
\begin{itemize}[leftmargin=0.5cm,itemsep=-0.55mm]
    \item \textit{Propensity, balancing, and deconfounding scores} \citep{rosenbaum1983central} were employed to estimate \emph{average} causal quantities \citep{antonelli2018doubly,d2021deconfounding}. Examples are average potential outcomes (APOs) and average treatment effect (ATE). This is because they lose information about the heterogeneity of the potential outcomes/treatment effect. In our work, on the other hand, we study a general class of \emph{heterogeneous} causal quantities, namely, representation-conditional CAPOs/CATE.
    \item \textit{Prognostic scores}  \citep{hansen2008prognostic} can be used for both averaged \citep{antonelli2018doubly,luo2020matching,d2021deconfounding} and heterogeneous causal quantities \citep{huang2017joint}. In \citep{huang2017joint,luo2020matching}, they are used in the context of a sufficient covariate dimensionality reduction. Yet, these works either (i)~make simplifying strong assumptions \citep{antonelli2018doubly,luo2020matching,d2021deconfounding}, so that the prognostic scores coincide with the expected covariate-conditional outcome; or (ii)~consider only linear prognostic scores \citep{huang2017joint,luo2020matching}. To the best of our knowledge, the first practical method for non-linear, learnable representations was proposed in \citep{johansson2016learning,shalit2017estimating,johansson2022generalization}.
\end{itemize}

Hence, the above-mentioned works operate in much simpler settings and, therefore, are \underline{not} relevant baselines for our work.
}

\subsection{Two-stage Meta-learners} \label{app:extended-rw-orthogonal-learners}

\textbf{Meta-learners.} Causal quantities can be estimated using model-agnostic methods, so-called \emph{meta-learners} \citep{kunzel2019metalearners}. Meta-learners typically combine multiple models to perform two-stage learning, namely, (1)~nuisance functions estimation and (2)~target model fitting. As such, meta-learners must be instantiated with some machine learning model (e.g., a neural network) to perform (1) and (2). Notable examples include X- and U-learners \citep{kunzel2019metalearners}, R-learner \citep{nie2021quasi}, DR-learner \citep{kennedy2023towards,curth2020estimating}, and IVW-learner \citep{fisher2024inverse}. \citet{curth2021nonparametric} provided a comparison of meta-learners implemented via neural networks, where unconstrained representations are used solely to estimate (1)~nuisance functions but not as inputs to the (2)~target model (as we analyze in our work).

\textbf{Neyman-orthogonal learners.} Neyman-orthogonality \citep{foster2023orthogonal}, or double/debiased machine learning \citep{chernozhukov2017double}, directly extend the idea of semi-parametric efficiency to infinite-dimensional target estimands such as CAPOs and the CATE. Informally, Neyman-orthogonality means that the population loss of the target model is first-order insensitive to the misspecification of the nuisance functions. Examples of model-agnostic\footnote{Several works extended the theory of targeted maximum likelihood estimation \citep{van2011targeted} and proposed sieves-based Neyman-orthogonal learners (\eg, EP-learner for CATE \citep{van2024combining} and i-learner for CAPOs \citep{vansteelandt2025orthogonal}). Yet, those methods are not fully model-agnostic (namely, they cannot be instantiated with neural networks) and are thus not considered in our work.} Neyman-orthogonal learners are DR-learners for CAPOs \citep{vansteelandt2025orthogonal}; and DR-, R-, IVW-learners for CATE \citep{morzywolek2023general}.

\subsection{Estimation of Causal Quantities for General-purpose Learned Representations} 

Other constraints may be applied to the representations, for example, to achieve algorithmic fairness \citep{zemel2013learning,madras2018learning}. Some works combined Neyman-orthogonal learners and fairness constraints, but different from our setting. For example, \citep{kim2023fair} provided a DR-learner for fair CATE estimation based on the linear combination of the basis functions; and \citep{frauen2024fair} built fair representations for policy learning with DR-estimators of policy value. The latter work, nevertheless, can be seen as a special case of the general \ORlearners.

\newpage
\section{BACKGROUND MATERIALS} \label{app:background}

In this section, we provide the formal definitions of Neyman-orthogonality, H{\"o}lder smoothness, and integral probability metrics; we state the identifiability and smoothness assumptions; and we offer an overview of meta-learners for CAPOs/CATE estimation. 

\subsection{Neyman-orthogonality and Double Robustness} \label{app:background-NO-DR}

We use the following additional notation: $\norm{\cdot}_{L_p}$ denotes the $L_p$-norm with  $\norm{f}_{L_p} = {\mathbb{E}(\abs{f(Z)}^p)}^{1/p}$, $a \lesssim b$ means there exists $C \ge 0$ such that $a \le C \cdot b$, and $X_n = o_{\mathbb{P}}(r_n)$ means $X_n/r_n \stackrel{p}{\to} 0$.

\begin{definition}[Neyman-orthogonality \citep{foster2023orthogonal,morzywolek2023general}]
    A risk $\mathcal{L}$ is called \emph{Neyman-orthogonal} if its pathwise cross-derivative equals zero, namely,
    \begin{equation} \label{eq:neym-orth-def}
         D_\eta D_g {\mathcal{L}}(g^*, \eta)[g- g^*, \hat{\eta} - \eta] = 0 \quad \text{{for all} } g \in \mathcal{G} \text{ and } \hat{\eta} \in \mathcal{H},
    \end{equation}
    where $D_f F(f)[h] = \frac{\diff}{\diff{t}} F (f + th) \vert_{t=0}$ and $D_f^k F(f)[h_1, \dots, h_k] = \frac{\partial^k}{\partial{t_1} \dots \partial{t_k}} F (f + t_1 h_1 + \dots + t_k h_k)  \vert_{t_1=\dots=t_k = 0}$ are pathwise derivatives \citep{foster2023orthogonal}; $g^* = \argmin_{g \in \mathcal{G}} \mathcal{L}(g, \eta)$; and $\eta$ is the ground-truth nuisance function. 
\end{definition}

Informally, this definition means that the risk is first-order insensitive wrt. the misspecification of the nuisance functions.

{
\begin{definition}[Double robustness]\label{def:rate-dr}
An estimator \(\hat{g}^* = \argmin_{g \in \mathcal{G}} \mathcal{L}(g, \hat{\eta})\) of \(g^* = \argmin_{g \in \mathcal{G}}\mathcal{L}({g}, {\eta})\) is said to be \emph{double robust} if, for any estimators \(\hat{\mu}_a^x\) and $\hat{\pi}_1^x$ of the nuisance functions $\mu_a^x$ and $\pi_1^x$, it holds that
\begin{equation} \label{eq:rate-dr-def}
    \norm{\hat{g}^* - g^*}_{L_2}^2 \lesssim \mathcal{L}(\hat{g}^*, \hat{\eta}) - \mathcal{L}({g}^*, \hat{\eta}) + \underbrace{\norm{\hat{\pi}_1^x - \pi_1^x}^2_{L_4} \norm{\hat{\mu}_a^x - \mu_a^x}^2_{L_4}}_{R_2(\eta, \hat{\eta})} ,
\end{equation}
where \(\mathcal{L}(\hat{g}^*, \hat{\eta}) - \mathcal{L}({g}^*, \hat{\eta})\) is the difference between the risks of the estimated target model and the optimal target model where the estimated nuisance functions are used, and ${R_2(\eta, \hat{\eta})}$ is a second-order remainder. 
\end{definition}

\begin{definition}[Quasi-oracle efficiency]\label{def:quasi-oracle}
An estimator \(\hat{g}^* = \argmin_{g \in \mathcal{G}} \mathcal{L}(g, \hat{\eta})\) of \(g^* = \argmin_{g \in \mathcal{G}}\mathcal{L}({g}, {\eta})\) is said to be \emph{quasi-oracle efficient} if the estimators \(\hat{\mu}_a^x\) and $\hat{\pi}_1^x$ of the nuisance functions $\mu_a^x$ and $\pi_1^x$ are allowed to have slow rates of convergence, $o_{\mathbb{P}}(n^{-1/4})$, and the following still holds asymptotically:
\begin{equation} \label{eq:oracle-eff}
    \norm{\hat{g}^* - g^*}_{L_2}^2 \lesssim \mathcal{L}(\hat{g}^*, \hat{\eta}) - \mathcal{L}({g}^*, \hat{\eta}) + \underbrace{o_{\mathbb{P}}(n^{-1/2})}_{R_2(\eta, \hat{\eta})},
\end{equation}
where \(\mathcal{L}(\hat{g}^*, \hat{\eta}) - \mathcal{L}({g}^*, \hat{\eta})\) is the difference between the risks of the estimated target model and the optimal target model where the estimated nuisance functions are used, and ${R_2(\eta, \hat{\eta})}$ is a second-order remainder. 

Furthermore, if the finite-sample estimator is used, namely $\hat{g} = \argmin_{g \in \mathcal{G}} \hat{\mathcal{L}}(g, \hat{\eta})$, the error between $\hat{g}$ and $g^*$ can be upper-bounded as 
\begin{equation} \label{eq:oracle-eff-fin}
    \norm{\hat{g} - g^*}_{L_2}^2 \lesssim \operatorname{Rate}_{\mathcal{D}}(\mathcal{G}; \hat{g}, \hat{\eta}) + \underbrace{o_{\mathbb{P}}(n^{-1/2})}_{_{R_2(\eta, \hat{\eta})}},
\end{equation}
where $\operatorname{Rate}_{\mathcal{D}}(\mathcal{G}; \hat{g}, \hat{\eta})$ is the convergence rate of the target model that satisfies $\mathcal{L}(\hat{g}, \hat{\eta}) - \mathcal{L}({g}^*, \hat{\eta}) \le \operatorname{Rate}_{\mathcal{D}}(\mathcal{G}; \hat{g}, \hat{\eta})$ for any $\hat{\eta} \in \mathcal{H}$ \citep{foster2023orthogonal}.
\end{definition}}

\subsection{H{\"o}lder Smoothness} \label{app:background-holder}

\begin{definition}[H{\"o}lder smoothness] \label{def:holder}
Let $s > 0, s -\lfloor s \rfloor \in (0, 1]$, and $\mathcal{X} \subseteq \mathbb{R}^{d_x}$. A function $f: \mathcal{X} \rightarrow \mathbb{R}$ is said to be \emph{$s$-H{\"o}lder smooth} (i.e., belongs to the H{\"o}lder class $C^s(\mathcal{X})$) if it is $\lfloor s \rfloor$-times continuously differentiable, and, for any $x, x' \in \mathcal{X}$, there exists a constant $L > 0$ such that for every $m$:
\begin{equation}
    \abs{D^mf(x) - D^mf(x')} \le L \norm{x-x'}_2^{s -\lfloor s \rfloor},
\end{equation}
where $m = (m_1, \dots, m_{d_x})$ such that $\abs{m} = \sum_j m_j = \lfloor s \rfloor$, $D^m = \frac{\partial^{\lfloor s \rfloor}}{\partial^{m_1}_{x_1} \dots \partial^{m_{d_x}}_{x_{d_x}}}$, and $\norm{\cdot}_2$ is the Euclidean norm. In our work, we set the constant $L$ to be a \emph{H{\"o}lder norm} $\norm{\cdot}_{C^s(\mathcal{X})}$:
\begin{equation}
    L = \norm{f}_{C^s(\mathcal{X})} := \sum_{\abs{m} \le \lfloor s \rfloor} \sup_{x\in \mathcal{X}}\abs{D^m f(x)} + \sum_{\abs{m} = \lfloor s \rfloor} \sup_{x, x'\in \mathcal{X}, x \neq x'} \frac{\abs{D^mf(x) - D^mf(x')}}{\norm{x-x'}_2^{s -\lfloor s \rfloor}},
\end{equation}
where the second term is also called a H\"older semi-norm $[D^mf]_{C^{0, s-\lfloor s\rfloor}(\mathcal{X})}$.
\end{definition}

H\"older smooth functions have a following useful property \citep{de1999regularity}: they can be well approximated by $\lfloor s \rfloor$-order Taylor approximations, namely,
\begin{equation}
    f(x) = f(x') + \sum_{j=1}^{\lfloor s \rfloor} \frac{1}{j!} D^jf(x')[x-x', \dots, x-x'] + R_{\lfloor s \rfloor}(x, x'),
\end{equation}
where $D^jf(x')[x-x', \dots, x-x']$ is a Taylor polynomial of degree $j$, and  $R_{\lfloor s \rfloor}(x, x')$ is a remainder term for which $\abs{R_{\lfloor s \rfloor}(x, x')} \le (1/\lfloor s \rfloor !) \norm{f}_{C^s(\mathcal{X})} \norm{x-x'}_2^s $.

Another useful property connects the H\"older norm $L$ with Lipschitz constant $\operatorname{Lip}(f)$ when $s \ge 1$:
\begin{equation}
    \operatorname{Lip}(f) \le \sup_{x \in \mathcal{X}}\norm{\nabla_x f(x)} = \sup_{x \in \mathcal{X}}\left(\sum_{j=1}^{d_x} \abs{D^jf(x)}^2 \right)^{1/2} \le \left(\sum_{j=1}^{d_x} L^2\right)^{1/2} = \sqrt{d_x} \, L,
\end{equation}
where $\nabla_x f(x)$ is a Jacobian of $f$.

\subsection{Integral Probability Metrics} \label{app:background-ipm}

Integral probability metrics (IPMs) are a broad class of distances between probability distributions, defined in terms of a family of functions $\mathcal{F}$. Given two probability distributions \( \mathbb{P}(Z_1) \) and \( \mathbb{P}(Z_2) \) over a domain \( \mathcal{Z} \), an IPM measures the maximum difference in expectation over a class of functions \( \mathcal{F} \):
\begin{equation}
    \operatorname{IPM}(\mathbb{P}(Z_1), \mathbb{P}(Z_2)) = \sup_{f \in \mathcal{F}} \left| \mathbb{E}(f(Z_1)) - \mathbb{E}(f(Z_2)) \right|.
\end{equation}
In this framework, \( \mathcal{F} \) specifies the allowable ways in which the difference between the distributions can be measured. Depending on the choice of \( \mathcal{F} \), different IPMs arise. 

\textbf{Wasserstein metric (WM).} The Wasserstein metric   is a specific IPM where the function class \( \mathcal{F} \) is the set of 1-Lipschitz functions, which are functions where the absolute difference between outputs is bounded by the absolute difference between inputs:
\begin{equation}
    W(\mathbb{P}(Z_1), \mathbb{P}(Z_2)) = \sup_{f \in \mathcal{F}_{1}} \left| \mathbb{E}(f(Z_1)) - \mathbb{E}(f(Z_2)) \right|.
\end{equation}
This metric can be interpreted as the minimum cost required to transport probability mass from one distribution to another, where the cost is proportional to the distance moved.

\textbf{Maximum mean discrepancy (MMD).} Another popular example is the maximum mean discrepancy, where the function class \( \mathcal{F} \) corresponds to functions in the unit ball of a reproducing kernel Hilbert space (RKHS), $\mathcal{F}_{\text{RKHS, 1}} = \{f \in \mathcal{H}: \norm{f}_{\mathcal{H}} \le 1\}$:
\begin{equation}
    \text{MMD}(\mathbb{P}(Z_1), \mathbb{P}(Z_2)) = \sup_{f \in \mathcal{F}_{\text{RKHS},1}} \left| \mathbb{E}(f(Z_1)) - \mathbb{E}(f(Z_2)) \right|.
\end{equation}
The MMD is often used in hypothesis testing and in training generative models, particularly when the distributions are defined over high-dimensional data.

\subsection{Meta-learners for CAPOs and CATE Estimation}

\textbf{Plug-in learners.} A na{\"i}ve way to estimate CAPOs and CATE is to simply estimate $\hat{\mu}_0^x(x)$ and $\hat{\mu}_1^x(x)$ and `plug them into' the identification formulas for CAPOs and CATE. For example, an S-learner (S-Net) fits a single model with the treatment as an input, while a T-learner (T-Net) builds two models for each treatment \citep{kunzel2019metalearners}. Many end-to-end representation learning methods, such as TARNet \citep{shalit2017estimating} and BNN without the balancing constraint \citep{johansson2016learning}, can be seen as variants of the plug-in learner: In the end-to-end fashion, they build a representation of the covariates $\hat{\Phi}(x) \in \mathit{\Phi} \subseteq \mathbb{R}^{d_\phi}$ and then use $\hat{\Phi}$ to estimate $\hat{\mu}_a^x(x) = h_a(\Phi(x))$ with the S-Net (BNN w/o balancing) or the T-Net (TARNet).

Yet, plug-in learners have several major drawbacks \citep{morzywolek2023general,vansteelandt2025orthogonal}. (a)~They do not account for the selection bias, namely, that $\hat{\mu}_0^x$ is estimated better for the treated population and $\hat{\mu}_1^x$ for the untreated (this is also known as a plug-in bias \citep{kennedy2023towards}). (b)~In the case of CATE estimation, the plug-in learners might additionally fail to address the causal inductive bias that the CATE is a ``simpler'' function than both CAPOs \citep{kunzel2019metalearners,curth2021inductive}, as it is impossible to add additional smoothing for the CATE model separately from CAPOs models. (c)~It is also unclear how to consistently estimate the CAPOs/CATE depending on the subset of covariates $V \subseteq X$ with the aim of reducing the variance of estimation. For example, it is unclear how to estimate representation-level CAPOs, $\xi_a^\phi(\phi) = \mathbb{E}(Y[a] \mid \Phi(X) = \phi)$, and CATE, $\tau^\phi(\phi) = \mathbb{E}(Y[1] - Y[0] \mid \Phi(X) = \phi)$, especially when the representations are constrained.  

\textbf{Two-stage Neyman-orthogonal learners.} To address the shortcomings of plug-in learners, two-stage meta-learners were proposed (see Appendix~\ref{app:extended-rw-orthogonal-learners}). These proceed as follows: 
\begin{itemize}
    \item \textbf{(i)}~First, one chooses a \emph{target model class} $\mathcal{G} = \{g(\cdot): \mathcal{V} \subseteq \mathcal{X} \to \mathcal{Y}\}$ such as, for example, neural networks. {A target model takes a (possibly confounded) subset $V$ of the original covariates $X$ as an input and outputs the prediction of causal quantities conditioned on $V$, namely, CAPOs $\xi_a^v(v) = \mathbb{E}(Y[a] \mid V = v)$ or CATE $\tau^v(v) = \mathbb{E}(Y[1] - Y[0] \mid V = v)$.}
    \item \textbf{(ii)}~Then, two-stage meta-learners formulate one of the \emph{target risks} for $g(v)$, where $v \in \mathcal{V}$:
    \begin{align} \label{eq:target-risk-app}
        \mathcal{L}_{\mathcal{G}}(g, \eta) = \mathbb{E} \big[w(\pi^x_a(X)) \, (\chi^x(X, \eta) - g(V))^2  \big],
    \end{align}
    where $\eta = (\mu_0^x, \mu_1^x, \pi^x_1)$ are the nuisance functions, $w(\cdot) > 0$ is the weighting function, $\chi^x(x, \eta) = \mu_a^x(x)$ for CAPOs, and $\chi^x(x, \eta) = \mu_1^x(x) - \mu_0^x(x)$ for CATE. Based on $w(\cdot)$, there are multiple choices for choosing a target risk, each with different interpretations and implications for finite-sample two-stage estimation. For example, the DR-learners for CAPOs/CATE use $w(\pi^x_a(x))=1$, and the R-/IVW-learners for CATE use overlap weights $w(\pi^x_a(x))=\pi^x_0(x) \, \pi^x_1(x)$. Furthermore, it is easy to see that the minimization of the target risks in Eq.~\eqref{eq:target-risk-app} yields the best projection of the ground-truth $V$-level causal quantities: DR-learners yield the projection of CAPOs/CATE, $\xi_a^v(v)/\tau^v(v)$; and the R-/IVW-learners yield the projection of the overlap-weighted CATE, $\tau^v_{\pi_0 \, \pi_1}(v) = \mathbb{E}[\pi^x_0(X) \, \pi^x_1(X) (Y[1] - Y[0]) \mid V = v] \big/ \mathbb{E}[\pi^x_0(X) \, \pi^x_1(X) \mid V = v]$ \citep{morzywolek2023general,vansteelandt2025orthogonal}.
    \item \textbf{(iii)}~In the last step, two-stage meta-learners minimize an empirical version of the target risk $\hat{\mathcal{L}}_{\mathcal{G}}(g, \hat{\eta})$, which is estimated using observational data and the nuisance functions $\hat{\eta}$ estimated at the first stage. The latest step then yields so-called \emph{Neyman-orthogonal learners} when the target risk is estimated with semi-parametric efficient estimators \citep{robins1995semiparametric,foster2023orthogonal}. We provide specific definitions of different Neyman-orthogonal learners in Table~\ref{tab:meta-learners}. Therein, we also specify the second-order remainder terms associated with Neyman-orthogonality $R_2(\eta, \hat{\eta})$ that relate to the \emph{quasi-oracle efficiency} and \emph{double robustness} (see Appendix~\ref{app:background-NO-DR} for definitions). 
\end{itemize}

\begin{table}[t]
    \vspace{-0.2cm}
    \caption{Overview of two-stage Neyman-orthogonal learners. Here, $\eta = (\mu^x_0, \mu^x_1, \pi^x_1)$ are the nuisance functions.}
    \label{tab:meta-learners}
    \begin{center}
        \vspace{-0.25cm}
        \hspace{-0.25cm}
        \scalebox{0.88}{
            \scriptsize
            \begin{tabular}{p{1.1cm}|p{0.9cm}|p{9.3cm}|p{6.2cm}}
                \toprule
                 Causal quantity & Meta-learner & \multirow{2}{*}{Neyman-orthogonal loss, $\hat{\mathcal{L}}_{\mathcal{G}}(g, \hat{\eta})$} & \multirow{2}{*}{Second-order remainder, $R_2(\eta, \hat{\eta})$}  \\
                \midrule \multirow{5}{*}{CAPOs} & \multirow{2}{*}{$\text{DR}^{\text{K}}_a$} & \multirow{2}{*}{$\mathbb{P}_n \bigg\{ \bigg(\frac{\mathbbm{1}\{A = a\}}{\hat{\pi}^x_a(X)} \big( Y - \hat{\mu}_a^x(X) \big) + \hat{\mu}_a^x(X) - g(V)\bigg)^2 \bigg\}$} 
                &  \multirow{2}{*}{$\norm{\hat{\pi}_1^x - \pi_1^x}^2_{L_4} \norm{\hat{\mu}_a^x - \mu_a^x}^2_{L_4}$}\\
                & & & \\
                 \cmidrule(lr){2-4}   & \multirow{2}{*}{$\text{DR}^{\text{FS}}_a$} & \multirow{2}{*}{$\mathbb{P}_n \bigg\{ \frac{\mathbbm{1}\{A = a\}}{\hat{\pi}^x_a(X)} \big( Y - g(V)\big)^2  +  \bigg(1 - \frac{\mathbbm{1}\{A = a\}}{\hat{\pi}^x_a(X)}\bigg) \, \big(\hat{\mu}_a^x(X) - g(V)\big)^2  \bigg\}$} & \multirow{2}{*}{$\norm{\hat{\pi}_1^x - \pi_1^x}^2_{L_4} \norm{\hat{\mu}_a^x - \mu_a^x}^2_{L_4}$} \\ 
                 & & & \\
                \midrule \multirow{7}{*}{CATE} & \multirow{2}{*}{$\text{DR}^{\text{K}}$} & \multirow{2}{*}{$\mathbb{P}_n \bigg\{ \bigg(\frac{A - \hat{\pi}^x_1(X)}{\hat{\pi}^x_0(X) \, \hat{\pi}^x_1(X)} \big( Y - \hat{\mu}_A^x(X) \big) + \hat{\mu}_1^x(X) - \hat{\mu}_0^x(X) - g(V)\bigg)^2 \bigg\}$} & \multirow{2}{*}{$\sum_{a\in \{0, 1\}}\norm{\hat{\pi}_1^x - \pi_1^x}^2_{L_4} \norm{\hat{\mu}_a^x - \mu_a^x}^2_{L_4}$} \\
                & & & \\
                \cmidrule(lr){2-4} & \multirow{2}{*}{$\text{R}$} & \multirow{2}{*}{$\mathbb{P}_n \bigg\{\big(A - \hat{\pi}_1^x(X)\big)^2\Big(\frac{Y - \hat{\mu}^x(X)}{A - \hat{\pi}_1^x(X)} -  g(V)\Big)^2 \bigg\}$} & \multirow{2}{*}{$\sum_{a\in \{0, 1\}}\norm{\hat{\pi}_1^x - \pi_1^x}^2_{L_4} \norm{\hat{\mu}_a^x - \mu_a^x}^2_{L_4} + \norm{\hat{\pi}_1^x - \pi_1^x}^4_{L_4}$} \\
                & & & \\
                \cmidrule(lr){2-4} & \multirow{2}{*}{$\text{IVW}$} & \multirow{2}{*}{$\mathbb{P}_n \bigg\{\big(A - \hat{\pi}_1^x(X)\big)^2\Big(\frac{A - \hat{\pi}^x_1(X)}{\hat{\pi}^x_0(X) \, \hat{\pi}^x_1(X)} \big( Y - \hat{\mu}_A^x(X) \big) + \hat{\mu}_1^x(X) - \hat{\mu}_0^x(X) -  g(V)\Big)^2 \bigg\}$} & \multirow{2}{*}{$\sum_{a\in \{0, 1\}}\norm{\hat{\pi}_1^x - \pi_1^x}^2_{L_4} \norm{\hat{\mu}_a^x - \mu_a^x}^2_{L_4} + \norm{\hat{\pi}_1^x - \pi_1^x}^4_{L_4}$} \\
                & & & \\
                \bottomrule
                \multicolumn{4}{p{16cm}}{\emph{References}:} \\ 
                \multicolumn{4}{p{16cm}}{\quad $\text{DR}^{\text{K}}_a$ \citep{kennedy2023towards}; $\text{DR}^{\text{FS}}_a$ \citep{foster2023orthogonal}; $\text{DR}^{\text{K}}$ \citep{kennedy2023towards}; $\text{R}$ \citep{nie2021quasi}; $\text{IVW}$ \citep{fisher2024inverse}}
            \end{tabular}
        }
    \end{center}
    \vspace{-0.2cm}
\end{table}

\textbf{End-to-end Neyman-orthogonality.} As noted by \citet{vansteelandt2025orthogonal}, under \emph{certain conditions}, the end-to-end IPTW-learners for CAPOs might possess Neyman-orthogonality. Specifically, if we assume that the ground-truth CAPOs are contained in the target model class (\ie, $V = X$ and $\xi^a_x \in \mathcal{G}$ so that $g^* = \xi^a_x$), the target causal quantity $\xi^a_x$ coincides with one of the nuisance functions $\mu^a_x$. Therefore, by setting $g = \mu^a_x$, we can simplify the original DR-loss in the style of \citet{foster2023orthogonal}:
\begin{align} \label{eq:ete-iptw-FS}
    \hat{\mathcal{L}}^{\text{DR}^{\text{FS}}_a}_{\mathcal{G}}(g, \hat{\eta} = (\mu^x_a = g, \pi^x_a)) = \mathbb{P}_n \bigg\{ \frac{\mathbbm{1}\{A = a\}}{\hat{\pi}^x_a(X)} \big( Y - g(X)\big)^2 \bigg\},
\end{align}
and original DR-loss in the style of \citet{kennedy2023towards}:
\begin{align} \label{eq:ete-iptw-K}
    \hat{\mathcal{L}}^{\text{DR}^{\text{K}}_a}_{\mathcal{G}}(g, \hat{\eta} = (\mu^x_a = g, \pi^x_a)) = \mathbb{P}_n \bigg\{ \frac{\mathbbm{1}\{A = a\}}{(\hat{\pi}^x_a(X))^2} \big( Y - g(X)\big)^2 \bigg\}.
\end{align}
Both losses in Eq.~\eqref{eq:ete-iptw-FS}-\eqref{eq:ete-iptw-K} are examples of weighted plug-in (WPI) / IPTW learners, and it can be easily shown that they both possess Neyman-orthogonality and quasi-oracle efficiency. Thus, some of the end-to-end representation learning methods that use IPTW weighting (namely, CFR-ISW \citep{hassanpour2019counterfactual} and BWCFR \citep{assaad2021counterfactual}) are Neyman-orthogonal when no balancing constraint is enforced or the invertible representations are used (so that the above-mentioned conditions are met).

\newpage
\section{{THEORETICAL RESULTS}} \label{app:proofs}

\begin{numlemma}{1}[Quasi-oracle efficiency of a non-parametric model] \label{lemma:quasi-oracle-eff-app}
    Assume (i)~a non-parametric target model $g(v), v \in \mathcal{V} \subseteq \mathbb{R}^{d_v}$ and (ii)~that a convergence rate for the target model does not depend on whether estimated or the ground-truth nuisance functions are used. Then, the error between $g^{*v} = \argmin_{g \in \mathcal{G}}\mathcal{L}_{\mathcal{G}}(g, \eta)$ and $\hat{g}^v = \argmin_{g \in \mathcal{G}}\hat{\mathcal{L}}_{\mathcal{G}}(g, \hat{\eta})$ can be upper-bounded as:
    \vspace{-0.1cm}
    \begin{align} \label{eq:quasi-oracle-v-app}
        \norm{g^{*v} - \hat{g}^v}_{L_2}^2 \lesssim (L^v)^{2d_v/(2s^v+d_v)} \cdot n^{-{2s^v}/{(2s^v + d_v)}} + R_2(\eta, \hat{\eta}),
    \end{align}
    where $g^{*v}$ is an $s^v$-H\"older smooth function with the H\"older norm $L^v = \norm{g^{*v}}_{C^{s^v}(\mathcal{V})}$, and $R_2(\eta, \hat{\eta})$ is a second-order remainder that depends on $n$, $s^x_{\mu_a}, s^x_{\pi}$, and $d_x$. For example, if the nuisance functions are estimated with non-parametric models, the second-order term has the following form:
    \begin{align}
        \text{\emph{DR}}^a &: R_2(\eta, \hat{\eta}) \lesssim n^{-2 ({s^x_{\mu_a}}/({s^x_{\mu_a} + d_x}) + {s^x_{\pi}}/({s^x_{\pi} + d_x}))}, \\
        \text{\emph{DR}} &: R_2(\eta, \hat{\eta}) \lesssim n^{-2 (\min_{a \in \{0, 1\}} {s^x_{\mu_a}}/({s^x_{\mu_a} + d_x}) + {s^x_{\pi}}/({s^x_{\pi} + d_x}))}, \\
        \text{\emph{R}/\emph{IVW}}&:  R_2(\eta, \hat{\eta}) \lesssim n^{-2 (\min_{a \in \{0, 1\}} {s^x_{\mu_a}}/({s^x_{\mu_a} + d_x}) + {s^x_{\pi}}/({s^x_{\pi} + d_x}))} + n^{-4 {s^x_{\pi}}/({s^x_{\pi} + d_x})) },
    \end{align}
    where H\"older norms are omitted for clarity.
\end{numlemma}
\begin{proof}
    In the following, to simplify the notation, we drop an upper index $v$ for both $g^{*v}=g^{*}$ and $\hat{g}^{v}=\hat{g}$. We adopt a classical convergence rate of a non-parametric regression \citep{stone1982optimal}. Specifically, we define a second-stage estimator $\hat{g}_p$ as a local polynomial/linear smoother estimator of order $p \ge \lfloor s \rfloor$ (we denote the class of such estimators as $\mathcal{G}$). Under the usual regularity assumptions \citep{kennedy2023towards} and given the Taylor expansion property of H\"older smooth functions (see Appendix~\ref{app:background-holder}), there exists a constant $C > 0$ and a bandwidth $h > 0$ such that, for an oracle second-stage estimator $g^*_p$, the following holds:
    \begin{equation}
        \inf_{g_p \in \mathcal{G} } \norm{g_p - g^*}_{L_2} = \norm{{g}^* _p- g^*}_{L_2} \lesssim \sqrt{C} L^v h^{s^v}.
    \end{equation}
    Let us denote $\hat{g}^*_p =  \argmin_{g_p \in \mathcal{G}}\hat{\mathcal{L}}_{\mathcal{G}}(g_p, {\eta})$. Then, using the standard local Rademacher/VC arguments, the estimation error is as follows:
    \begin{equation}
        \norm{\hat{g}_p^* - g^*_p}_{L_2}^2 \lesssim  \frac{C}{nh^{d_v}}, 
    \end{equation}
    and, thus, the desired estimation error becomes
    \begin{equation}
        \norm{\hat{g}_p^* - g^*}_{L_2}^2 \le 2 \norm{\hat{g}_p^* - {g}_p^*}_{L_2}^2 + 2\norm{{g}_p^* - g^*}_{L_2}^2 \lesssim \frac{C}{nh^{d_v}} + C (L^v)^2 h^{2s^v}.
    \end{equation}
    Now, by choosing $h \asymp \Big(\frac{1}{(L^v)^2 n} \Big)^{1/(2s^v + d_v)}$, we recover the following bound (with the known nuisance functions):
    \begin{equation}
        \norm{\hat{g}_p^* - g^*}_{L_2}^2 \lesssim C \underbrace{(L^v)^{2d_v/(2s^v+d_v)} \cdot n^{-{2s^v}/{(2s^v + d_v)}}}_{\operatorname{Rate}_{\mathcal{D}}(\mathcal{G}; \hat{g}_p, {\eta})}.
    \end{equation}

    Finally, by using the error bound from the quasi-oracle efficiency of the Neyman-orthogonal learners (see Definition~\ref{def:quasi-oracle}) and assuming that ${\mathcal{L}}_{\mathcal{G}}(\hat{g}_p, \hat{\eta}) - {\mathcal{L}}_{\mathcal{G}}(\hat{g}_p^*, \hat{\eta})= \operatorname{Rate}_{\mathcal{D}}(\mathcal{G}; \hat{g}_p, \hat{\eta})  \lesssim  \operatorname{Rate}_{\mathcal{D}}(\mathcal{G}; \hat{g}_p, {\eta})$ (convergence rate for the target model does not depend on whether estimated or the ground-truth nuisance functions are used), we recover the desired inequality
    \begin{align}
        \norm{\hat{g}_p - g^*}_{L_2}^2 & \le 2 \norm{\hat{g}_p - \hat{g}_p^*}_{L_2}^2  + 2 \norm{\hat{g}_p^* - g^*}_{L_2}^2  \\
        &\lesssim {\mathcal{L}}_{\mathcal{G}}(\hat{g}_p, \hat{\eta}) - {\mathcal{L}}_{\mathcal{G}}(\hat{g}_p^*, \hat{\eta}) + R_2(\eta, \hat{\eta}) + \norm{\hat{g}_p^* - g^*}_{L_2}^2\\
        &\lesssim (L^v)^{2d_v/(2s^v+d_v)} \cdot n^{-{2s^v}/{(2s^v + d_v)}} + R_2(\eta, \hat{\eta}),
    \end{align}
    where $\hat{g}_p = \argmin_{g_p \in \mathcal{G}}\hat{\mathcal{L}}_{\mathcal{G}}(g_p, \hat{\eta})$.
    
    For the upper-bound on the second-order remainder, $R_2(\eta, \hat{\eta})$, we refer to existing results of \citep{curth2021nonparametric,kennedy2023towards,schulte2025adjustment}.
\end{proof}

\begin{numprop}{1} \label{prop:manifold-quasi-oracle-eff-app}
    Under Assumption~\ref{ass:manifold}, the following holds:
    (1)~$d_{\phi^*} \ll d_x$, (2)~$g^{*\phi^*}$ is an $s^{\phi^*}$-H\"older smooth function with H\"older norm $L^{\phi^*}$ such that $s^{\phi^*}\ge s^x$ and $L^{\phi^*} \le c(L^{J^*}) \cdot L^x$ with non-decreasing $c(\cdot)$. Also, when $L^{J^*}$ is sufficiently small,
    \vspace{-0.1cm}
    \begin{align}
        \norm{g^{*\phi^*} - \hat{g}^{\phi^*}}_{L_2}^2 &\lesssim \norm{g^{*x} - \hat{g}^x}_{L_2}^2.
    \end{align}
\end{numprop}

\begin{proof}
    First, (1)~$d_{\phi^*} \ll d_x$ is directly given by Assumption~\ref{ass:manifold}. Furthermore, (2) can be derived from the properties of the H\"older smooth functions. Specifically, under Assumption~\ref{ass:manifold}, the causal quantities in the representation space can be written as the composition of the functions:
    \begin{equation}
        \xi_a^{\phi^*}(\phi) = \xi_a^{\phi^*}(\Phi^* (J^*(\phi))) = \xi_a^x(J^*(\phi)) \quad \text{ and } \quad  \tau^{\phi^*}(\phi) = \tau^{\phi^*}(\Phi^* (J^*(\phi))) = \tau^x(J^*(\phi)).
    \end{equation}
    Therefore, we can use the standard boundedness of the composition operator on H\"older spaces (see Theorem 4.3 in \citep{de1999regularity}):
    \begin{equation}
        J^*\in C^{s^x+1}(\mathit{\Phi}^*), \,\, g^{x*} =\xi_a^x/ \tau^x \in C^{s^x}(\mathcal{X}) \quad \Rightarrow \quad g^{*\phi^*} =  g^{x*} \circ J^*\in C^{s^x}(\mathcal{X}),
    \end{equation}
    and 
    \begin{equation}
        \norm{g^{*\phi^*}}_{C^{s^x}(\mathit{\Phi}^*)} = \norm{ g^{x*} \circ J^* }_{C^{s^x}(\mathit{\Phi}^*)} \le K_{J^*} \norm{ g^{x*}}_{C^{s^x}(\mathcal{X}^*)},
    \end{equation}
    where $K_{J^*} = \sup_{g \neq 0}\frac{{\norm{g \circ J^*}}_{C^{s^x}(\mathit{\Phi}^*)}}{\norm{g}_{C^{s^x}(\mathcal{X})}}$ is an operator norm of the pullback $J^*$. Therefore, $s^{\phi^*} \ge s^x$ and $L^{\phi^*} \le K_{J^*} \cdot L^x$.

    Furthermore, it was demonstrated in Theorem 4.3 of \citet{de1999regularity} that (1)~if $0 < s^x \le 1$, it can be shown that $K_{J^*} \le \max \{1, \operatorname{Lip}(J^*)^{s^x}\} \le \max \{1, (L^{J^*})^{s^x}\}$; and (2)~if $s^x > 1$, then $K_{J^*} \le P(s^x, L^{J^*})$, where $P$ is a polynomial of degree $\lfloor s \rfloor$ with non-negative coefficients (follows from a Faa di Bruno expansion). Thus, in both cases (1)-(2), $K_{J^*}$ can be upper-bounded by some non-decreasing function $c$ depending on the H\"older norm $L^{J^*}$.

    Finally, to compare the error bounds between $V = \Phi^*(X)$ and $V = X$, we refer to Lemma~\ref{lemma:quasi-oracle-eff-app}. That is, the error bounds are 
    \begin{align}
        {V = \Phi^*(X)}&: \quad \norm{g^{*\phi^*} - {\hat{g}^{\phi^*}}}_{L_2}^2 \lesssim (L^{\phi^*})^{2d_{\phi^*}/(2s^{\phi^*}+d_{\phi^*})} \cdot n^{-{2s^{\phi^*}}/{(2s^{\phi^*} + d_{\phi^*})}} + R_2(\eta, \hat{\eta}), \\
        V = X &: \quad\norm{g^{*x} - \hat{g}^x}_{L_2}^2 \lesssim (L^x)^{2d_x/(2s^x+d_x)} \cdot n^{-{2s^x}/{(2s^x + d_x)}} + R_2(\eta, \hat{\eta}).
    \end{align}
    Here, the second-order error terms are the same, yet the target model error terms differ:
    \begin{align}
        (L^{\phi^*})^{2d_{\phi^*}/(2s^{\phi^*}+d_{\phi^*})} \le [c(L^{J^*}) \cdot L^x]^{2d_x/(2s^x+d_x)} \quad \text{and} \quad n^{-{2s^{\phi^*}}/{(2s^{\phi^*} + d_{\phi^*})}} \le n^{-{2s^x}/{(2s^x + d_x)}}.
    \end{align}
    Therefore, when $L^{J^*}$ is sufficiently small such that $c(L^{J^*}) \le 1$, the representation-level learners asymptotically achieve a lower error than the covariate-level learners.
\end{proof}

\begin{numprop}{2}[Smoothness of the hidden layers] \label{prop:smoothness-ass}
 We denote the trained representation network as $\hat{\mu}^x_a =\hat{h}_a \circ \hat{\Phi} = \argmin \hat{\mathcal{L}}_{\mathit{\Phi}}(h_a\circ{\Phi})$. Let $s \ge s_{\mu_a}^x$ and the trained representation network be factorized as
    \begin{equation}
        \hat{\mu}^x_a = T_{L+1}\circ T_L \circ \cdots \circ T_1,
    \end{equation}
    where $T_\ell : \mathcal{V}_{\ell-1} \to \mathcal{V}_\ell$, $\mathcal{V}_0=\mathcal X$, and $T_{L+1}:\mathcal{V}_L\to  \mathcal{Y}$. For each hidden layer $\ell=1,\dots,L$, define
    \begin{equation}
        \hat{f}^{(\ell)} := T_\ell \circ \cdots \circ T_1, \qquad \mathcal{V}_\ell := \hat{f}^{(\ell)}(\mathcal X),
    \end{equation}
    and the corresponding tail network
    \begin{equation}
        \hat{h}_a^{(\ell)} := T_{L+1}\circ T_L \circ \cdots \circ T_{\ell+1}: \mathcal{V}_\ell \to \mathcal{Y},
    \end{equation}
    so that $\hat{\mu}^x_a(x)=\hat{h}_a^{(\ell)}(\hat{f}^{(\ell)}(x))$.

    Assume:
    
    \begin{enumerate}
        \item[(i)] $\mathcal X$ is compact; $\mu_a^x \in C^{s_{\mu_a}^x}(\mathcal X); \|\mu_a^x\|_{C^{s_{\mu_a}^x}(\mathcal X)} = L_{\mu_a}^x$.
    
        \item[(ii)] There exists $B_a>0$ such that $\|T_{L+1}\|_{C^s(\mathcal{V}_L)} \le B_a$.
    
        \item[(iii)] For each hidden layer $j=2,\dots,L$, there exists $K_{T_j}\in(0,1]$ such that, for every scalar-valued $g\in C^s(\mathcal{V}_j)$,
        $\|g\circ T_j\|_{C^s(\mathcal{V}_{j-1})} \le K_{T_j} \|g\|_{C^s(\mathcal{V}_j)}$.
    
        \item[(iv)] For each $\ell=1,\dots,L$, define the conditional residual
        $\hat{r}_{\ell,n}(v) := \mathbb E[\mu_a^x(X)-\hat{h}_a^{(\ell)}(V_\ell)\mid V_\ell=v]$.
        Assume that $\hat{r}_{\ell,n}\in C^s(V_\ell)$ and $\|\hat{r}_{\ell,n}\|_{C^s(V_\ell)} \le \varepsilon_n, \varepsilon_n \to 0$.
    
        \item[(v)] There exists $\Delta>0$ such that
        $B_a \prod_{j=2}^L K_{T_j} \le L_{\mu_a}^x - \Delta, \varepsilon_n \le \Delta \text{ for all sufficiently large } n$.
    \end{enumerate}

    Then there exists a hidden layer $V_\ell=\hat{f}^{(\ell)}(\mathcal X)$ such that the
    representation-level regression target
    \begin{equation}
        \mu_a^{v_\ell}(v):= \mathbb E[\mu_a^x(X)\mid V_\ell=v]
    \end{equation}
    satisfies
    \begin{equation}
        \mu_a^{v_\ell} \in C^s(\mathcal{V}_\ell), \qquad \|\mu_a^{v_\ell}\|_{C^s(\mathcal{V}_\ell)} \le L_{\mu_a}^x.
    \end{equation}
    Hence,
    \begin{equation}
        s_{\mu_a}^{v_\ell} \ge s_{\mu_a}^x, \qquad L_{\mu_a}^{v_\ell} \le L_{\mu_a}^x.
    \end{equation}
    In fact, under assumption (v), one may take $\ell=1$.
\end{numprop}

\begin{proof}
    We choose $\ell=1$. By repeated application of assumption (iii),
    \begin{equation}
        \|\hat{h}_a^{(1)}\|_{C^s(\mathcal{V}_1)} = \|T_{L+1}\circ T_L\circ \cdots \circ T_2\|_{C^s(\mathcal{V}_1)} \le \Bigl(\prod_{j=2}^L K_{T_j}\Bigr)\|T_{L+1}\|_{C^s(\mathcal{V}_L)} \le B_a \prod_{j=2}^L K_{T_j}.
        \end{equation}
    Now define
    \begin{equation}
        \mu_a^{v_1}(v)  := \mathbb E[\mu_a^x(X)\mid V_1=v].
    \end{equation}
    By linearity of conditional expectation,
    \begin{equation}
        \mu_a^{v_1}(v) = \mathbb E\!\left[\hat{h}_a^{(1)}(V_1) + \bigl(\mu_a^x(X)-\hat{h}_a^{(1)}(V_1)\bigr) \,\middle|\, V_1=v\right] = \hat{h}_a^{(1)}(v)+\hat{r}_{1,n}(v).
    \end{equation}
    Since $\hat{h}_a^{(1)}\in C^s(\mathcal{V}_1)$ and $\hat{r}_{1,n}\in C^s(\mathcal{V}_1)$ by assumptions (ii)--(iv), it follows that $\mu_a^{v_1}\in C^s(\mathcal{V}_1)$. Using the triangle inequality for the full H\"older norm,
    \begin{equation}
        \|\mu_a^{v_1}\|_{C^s(\mathcal{V}_1)} \le \|\hat{h}_a^{(1)}\|_{C^s(\mathcal{V}_1)} + \|\hat{r}_{1,n}\|_{C^s(\mathcal{V}_1)} \le B_a \prod_{j=2}^L K_{T_j} + \varepsilon_n.
    \end{equation}
    By assumption (v), for all sufficiently large $n$,
    \begin{equation}
        \|\mu_a^{v_1}\|_{C^s(\mathcal{V}_1)} \le L_{\mu_a}^x.
    \end{equation}
    Because $s\ge s_{\mu_a}^x$, we conclude that
    \begin{equation}
        s_{\mu_a}^{v_1} \ge s_{\mu_a}^x, \qquad L_{\mu_a}^{v_1} \le L_{\mu_a}^x.
    \end{equation}
    Thus, there exists a hidden layer with the desired property.
\end{proof}

\begin{numprop}{3}[Smoothing via expanding mapping] \label{prop:expand-app}
    Assume that the trained representation network $\hat{\Phi}$ minimizes $\hat{\mathcal{L}}_{\mathit{\Phi}}(h_a \circ \Phi)$ and is $s^{\hat{\Phi}}$-H\"older smooth ($s^{\hat{\Phi}} \ge 1$). Then, under assumptions (i)-(v) of Proposition~\ref{prop:smoothness-ass} and 
    additional assumptions that:

    \begin{enumerate}
    \item[(vi)] The ground-truth target is non-constant and satisfies $\operatorname{Lip}(\mu_a^x)>0$.

    \item[(vii)] The fitted plug-in predictor is accurate in Lipschitz semi-norm: $\operatorname{Lip}(\mu_a^x-\hat{\mu}^x_a)\le \varepsilon_n, \varepsilon_n\to 0$.

    \item[(viii)] There exists $\Delta>0$ such that
    \begin{equation}
        \sqrt{d_{V_1}}\,B_a\prod_{j=2}^L K_{T_j} \le \operatorname{Lip}(\mu_a^x)-\Delta, \qquad \varepsilon_n\le \Delta \quad\text{for all sufficiently large } n,
    \end{equation}
\end{enumerate}
    
    then $\hat{\Phi}$ is an expanding mapping, namely, $\operatorname{Lip}({\hat{\Phi}}) \ge 1$. 
\end{numprop}

\begin{proof}
    By Proposition~\ref{prop:smoothness-ass}, for $\ell=1$ and $\hat{h}_a:=\hat{h}_a^{(1)}$,
    \begin{equation}
        \|\hat{h}_a\|_{C^s(\mathcal{V}_1)} \le B_a\prod_{j=2}^L K_{T_j}.
    \end{equation}
    Since $s^{\hat{\Phi}}\ge1$, the standard relation between the H\"lder norm and the Lipschitz constant yields
    \begin{equation}
        \operatorname{Lip}(\hat{h}_a) \le \sqrt{d_{V_1}}\,\|\hat{h}_a\|_{C^s(\mathcal{V}_1)} \le \sqrt{d_{V_1}}\,B_a\prod_{j=2}^L K_{T_j}.
    \end{equation}
    Hence, by assumption (viii),
    \begin{equation} \label{eq:lip-ha}
        \operatorname{Lip}(\hat{h}_a)\le \operatorname{Lip}(\mu_a^x)-\Delta.
    \end{equation}
    
    Now, by the triangle inequality for the Lipschitz constant and assumption (vii),
    \begin{equation}
        \operatorname{Lip}(\mu_a^x) \le \operatorname{Lip}(\hat{\mu}^x_a)+\operatorname{Lip}(\mu_a^x-\hat{\mu}^x_a) \le \operatorname{Lip}(\hat{\mu}^x_a)+\varepsilon_n.
    \end{equation}
    Since $\hat{\mu}^x_a=\hat{h}_a\circ \hat\Phi$, the composition rule for Lipschitz maps gives
    \begin{equation}
        \operatorname{Lip}(\hat{\mu}^x_a)\le \operatorname{Lip}(\hat{h}_a)\,\operatorname{Lip}(\hat\Phi).
    \end{equation}
    Therefore,
    \begin{equation}
        \operatorname{Lip}(\mu_a^x) \le \operatorname{Lip}(\hat{h}_a)\,\operatorname{Lip}(\hat\Phi)+\varepsilon_n.
    \end{equation}
    Rearranging,
    \begin{equation}
        \operatorname{Lip}(\hat\Phi) \ge \frac{\operatorname{Lip}(\mu_a^x)-\varepsilon_n}{\operatorname{Lip}(\hat{h}_a)}.
    \end{equation}
    Using Eq.~\eqref{eq:lip-ha},
    \begin{equation}
        \operatorname{Lip}(\hat\Phi) \ge \frac{\operatorname{Lip}(\mu_a^x)-\varepsilon_n}{\operatorname{Lip}(\mu_a^x)-\Delta}.
    \end{equation}
    For all sufficiently large $n$, assumption (viii) implies $\varepsilon_n\le \Delta$, hence
    \begin{equation}
        \operatorname{Lip}(\hat\Phi)\ge 1.
    \end{equation}
        This proves the claim.
\end{proof}

\begin{numprop}{4}[Balancing via contracting mapping] \label{prop:contract-app}
    Assume that the trained representation network with $s^{\hat{\Phi}}$-H\"older smooth $\hat{\Phi}$ ($s^{\hat{\Phi}} \ge 1$) that minimizes $\hat{\mathcal{L}}_\text{\emph{Bal}}(\Phi)$ with WM / MMD. Then, assuming that $\hat{\Phi}$ is non-constant and the class of representation networks is closed under rescaling, (1)~$\hat{\Phi}$ is a contracting mapping, namely
        $\operatorname{Lip}(\hat{\Phi}) \le 1$. Furthermore, if Assumption~\ref{ass:manifold} holds wrt. $\hat{\Phi}$ (\eg, $\hat{\Phi}$ is smoothly invertible), (2)~then the pullback map is expanding, namely, $\operatorname{Lip}(\hat{J}) \ge 1$. 
\end{numprop}
\begin{proof}
    (1) The proof proceeds separately for (a)~WM-based balancing and (b)~MMD-based balancing. Consider a scaling transformation applied to some $s^{\Phi}$-H\"older smooth representation ${\Phi}$ (with $s^{\Phi} \ge 1$): $\beta \cdot {\Phi}$ with $ \beta \in (0, 1)$. Without the loss of generality, we assume that the class of representation networks is closed under rescaling, so both $\Phi$ and $\beta \cdot \Phi$ belong to this class.

    (a)~{WM}. For any ${\Phi}$ and $ \beta \in (0, 1)$, we can use the pushforward property of the Wasserstein metric $W$:
    \begin{align}
        & W\Big((\beta \cdot \Phi)_{\#} \mathbb{P}(X \mid A=0), (\beta \cdot \Phi)_{\#} \mathbb{P}(X \mid A=1)  \Big) = \beta \, W\big(\Phi_{\#} \mathbb{P}(X \mid A=0), \Phi_{\#} \mathbb{P}(X \mid A=1)  \big) \\
        & \qquad \le  W\big(\Phi_{\#} \mathbb{P}(X \mid A=0), \Phi_{\#} \mathbb{P}(X \mid A=1) \big).
    \end{align}
    Hence, $\hat{\mathcal{L}}_\text{\emph{Bal}}(\beta \, \Phi) \le \beta \hat{\mathcal{L}}_\text{\emph{Bal}} (\Phi)$ with strict inequality unless the class-conditional distributions exactly coincide after $\Phi(\cdot)$ (given $\hat{\mathcal{L}}_\text{\emph{Bal}}$ is empirical, strict inequality holds almost always). 
    
    Therefore, the proof follows from contradiction: If for some empirical non-constant minimizer $\hat{\Phi}$, $\operatorname{Lip}(\hat{\Phi}) \ge 1$ ($\operatorname{Lip}(\cdot)$ is well-defined as $s^{\hat{\Phi}} \ge 1$), the rescaled $\tilde{{\Phi}} = \beta \, \hat{\Phi}$ with $\beta \le 1/\operatorname{Lip}({\hat{\Phi}})$ achieves almost always (strictly) better balancing loss and has
    \begin{equation}
        \operatorname{Lip}({\tilde{{\Phi}}}) = \beta \, \operatorname{Lip}({\hat{\Phi}}) \le \frac{\operatorname{Lip}({\hat{\Phi}})}{\operatorname{Lip}({\hat{\Phi}})}= 1.
    \end{equation}
    Thus, $\hat{{\Phi}}$ is not a proper minimizer of the empirical balancing loss, as we can select a strictly better minimizer with $\operatorname{Lip}({\tilde{{\Phi}}}) \le 1$. 
    
    (b)~{MMD}. Consider RKHS induced by shift-invariant, Lipschitz kernels $k(z, z') = \kappa(\norm{z-z'}_2)$ with $\kappa$ monotonically decreasing. Then, scaling with $\beta \in (0, 1)$ shrinks all pairwise distances, which reduces MMD:
    \begin{equation}
        \operatorname{MMD}\Big((\beta \cdot \Phi)_{\#} \mathbb{P}(X \mid A=0), (\beta \cdot \Phi)_{\#} \mathbb{P}(X \mid A=1) \Big) \le \operatorname{MMD}\Big(\Phi_{\#} \mathbb{P}(X \mid A=0), \Phi_{\#} \mathbb{P}(X \mid A=1) \Big),
    \end{equation}
    with strict decrease unless the two pushforwards already match. Therefore, by same rescaling argument as for (a)~WM, we can always find such $\beta$ so that the rescaled empirical minimizer $\tilde{{\Phi}} = \beta \hat{\Phi}$ has $\operatorname{Lip}({\tilde{{\Phi}}}) \le 1$.

    (2)~Finally, to show that $\operatorname{Lip}(\hat{J}) \ge 1$ under Assumption~\ref{ass:manifold}, we used the composition property of Lipschitz smooth functions:
    \begin{equation}
        \hat{\Phi} \circ \hat{J} = \operatorname{id}_{\hat{\mathit{\Phi}}} \quad \Rightarrow \quad  1 = \operatorname{Lip}(\operatorname{id}_{\hat{\mathit{\Phi}}}) \le \operatorname{Lip}(\hat{\Phi}) \cdot \operatorname{Lip}(\hat{J}).
    \end{equation}
    Hence, if $\operatorname{Lip}(\hat{\Phi}) \le 1$, it is necessary that $\operatorname{Lip}(\hat{J}) \ge 1$.
\end{proof}

\newpage
\section{VISUAL SUMMARY OF THEORETICAL RESULTS} \label{app:visual-summary}

\begin{figure*}[h]
    \centering
    \vspace{-0.2cm}
    \includegraphics[width=0.8\textwidth]{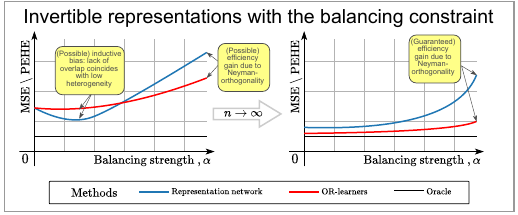} \\
    \vspace{0.2cm}
    \includegraphics[width=0.795\textwidth]{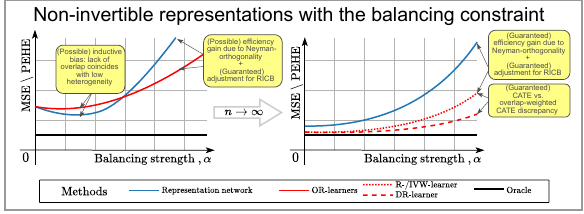}
    \setlength{\fboxsep}{0.5pt}
    \caption{\textbf{Insights for \textbf{RQ} \protect\circledred{2}}. For both figures, we highlight in  \protect\ovalbox{\colorbox{yellow}{yellow boxes}} how the \ORlearners (in \textcolor{tabred}{red}) can be beneficial in comparison with the end-to-end representation network (in \textcolor{tabblue}{blue}). Specifically, we compare the generalization performance in terms of MSE / precision in estimating heterogeneous effect (PEHE) (lower is better), depending on the strength of balancing, $\alpha$. In both cases, we show the behavior in a finite-sample vs. asymptotic regime ($n \to \infty$). The plots highlight the effectiveness of the \ORlearners in the asymptotic regime, especially when too much balancing is applied.}
    \vspace{-0.4cm}
    \label{fig:or-learner-summary}
\end{figure*}

\newpage
\section{DATASET DETAILS} \label{app:datasets}

\subsection{Synthetic Dataset}

We use a synthetic benchmark dataset with hidden confounding as proposed by \citet{kallus2019interval}, but modify it by incorporating the confounder as the second observed covariate. Specifically, synthetic covariates $X_1$ and $X_2$ along with treatment $A$ and outcome $Y$ are generated by the following data-generating process: 
\begin{equation}
    \begin{cases}
        X_1 \sim \text{Unif}(-2, 2), \\
        X_2 \sim N(0, 1),  \\
        A \sim \text{Bern}\left(\frac{1}{1 + \exp(-(0.75 \, X_1 - X_2 + 0.5))}\right) \\
        Y \sim N\big( (2\, A - 1) \, X_1 + A - 2 \, \sin(2 \, (2\,A - 1) \, X_1 + X_2) - 2\,X_2\,(1 + 0.5\,X_1) , 1\big),
    \end{cases}
\end{equation}
where $X_1, X_2$ are mutually independent.

\subsection{IHDP Dataset}
The Infant Health and Development Program (IHDP) dataset \citep{hill2011bayesian, shalit2017estimating} is a widely-used semi-synthetic benchmark for evaluating treatment effect estimation methods. It consists of 100 train/test splits, with $n_\text{train} = 672$, $n_\text{test} = 75$, and $d_x = 25$. However, this dataset suffers from significant overlap violations, leading to instability in methods that rely on inverse propensity weights \citep{curth2021nonparametric, curth2021really}.

\subsection{ACIC 2016 Dataset Collection}
The covariates for ACIC 2016 \citep{dorie2019automated} are derived from a large-scale study on developmental disorders \citep{niswander1972collaborative}. The datasets in ACIC 2016 vary in the number of true confounders, the degree of overlap, and the structure of conditional outcome distributions. ACIC 2016 features 77 distinct data-generating mechanisms, each with 100 equal-sized samples ($n = 4802, d_x = 82$) after one-hot encoding the categorical covariates.

\subsection{HC-MNIST Dataset}

The HC-MNIST benchmark was introduced as a high-dimensional, semi-synthetic dataset \citep{jesson2021quantifying}, derived from the original MNIST digit images \citep{lecun1998mnist}. It consists of $n_{\text{train}} = 60,000$ training images and $n_{\text{test}} = 10,000$ test images. Each image in HC-MNIST is compressed into a single latent coordinate, $\phi$, such that the potential outcomes are non-linear functions of both the image’s mean pixel intensity and its digit label. Treatment assignment then depends on this one-dimensional summary $\phi$ along with an additional latent (synthetic) confounder $U$, which is treated as an observed covariate. Specifically, the HC-MNIST dataset can be described by the following data-generating process:

\begingroup\makeatletter\def\f@size{9}\check@mathfonts
\begin{equation}
    \begin{cases}
        U \sim \text{Bern}(0.5), \\
        X \sim \text{MNIST-image}(\cdot), \\
        \phi := \left( \operatorname{clip} \left( \frac{\mu_{N_x} - \mu_c}{\sigma_c}; - 1.4, 1.4\right) - \text{Min}_c \right) \frac{\text{Max}_c - \text{Min}_c} {1.4 - (-1.4)}, \\
        \alpha(\phi; \Gamma^*) := \frac{1}{\Gamma^* \operatorname{sigmoid}(0.75 \phi + 0.5)} + 1 - \frac{1}{\Gamma^*}, \quad \beta(\phi; \Gamma^*) := \frac{\Gamma^*}{\operatorname{sigmoid}(0.75 \phi + 0.5)} + 1 - \Gamma^*, \\
        A \sim \text{Bern}\left( \frac{u}{\alpha(\phi; \Gamma^*)} + \frac{1 - u}{\beta(\phi; \Gamma^*)}\right), \\
        Y \sim N\big((2A - 1) \phi + (2A - 1) - 2 \sin( 2 (2A - 1) \phi ) - 2 (2U - 1)(1 + 0.5\phi), 1\big),
    \end{cases}
\end{equation}
\endgroup
where $c$ is a label of the digit from the sampled image $X$; $\mu_{N_x}$ is the average intensity of the sampled image; $\mu_c$ and $\sigma_c$ are the mean and standard deviation of the average intensities of the images with the label $c$; and $\text{Min}_c= -2 + \frac{4}{10}c, \text{Max}_c = -2 + \frac{4}{10}(c + 1)$. The parameter $\Gamma^*$ defines what factor influences the treatment assignment to a larger extent, i.e., the additional confounder or the one-dimensional summary. We set $\Gamma^* = \exp(1)$. For further details, we refer to \citet{jesson2021quantifying}.

\newpage
\section{IMPLEMENTATION DETAILS AND HYPERPARAMETERS} \label{app:implementation}

\begin{figure}[h]
    \centering
    \vspace{-0.2cm}
    \includegraphics[width=\textwidth]{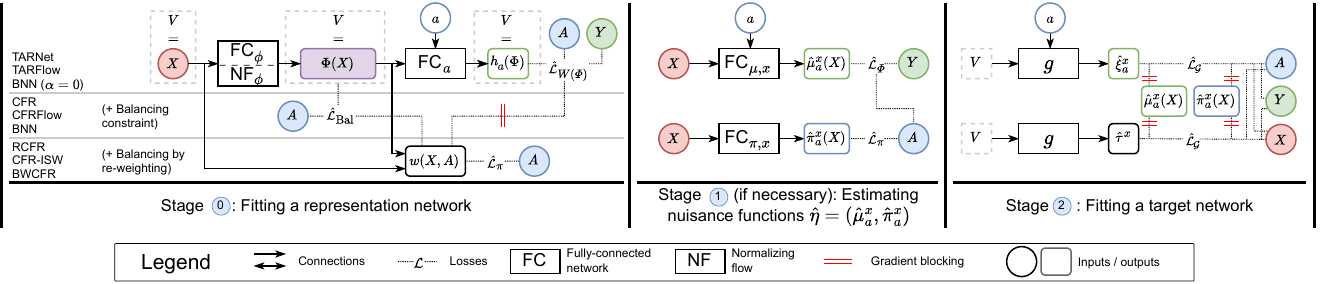}
    \vspace{-0.5cm}
    \caption{{\textbf{An overview of the \ORlearners.} The \ORlearners proceed in three stages: \protect\circled{0}~fitting a representation network, \protect\circled{1}~estimation of the nuisance functions, and \protect\circled{2}~fitting a target network. For stage \protect\circled{0}, we also show different options for the target network input $V$. Depending on the choice of the input $V$, the second-stage model $g(V)$ obtains different interpretations: it either learns a new model from scratch or performs a calibration of the representation network outputs.}}
    \vspace{-0.2cm}
    \label{fig:or-learner-overview}
\end{figure}

\textbf{Overview}. The \ORlearners use neural networks to fit a target model $g$ based on the learned representations $\hat{\Phi}(X)$. They proceed in three stages (see Fig.~\ref{fig:or-learner-overview}): \circled{0}~fitting a representation network; \circled{1}~estimating nuisance functions (if necessary); and \circled{2}~fitting a target network. The pseudocode is in Algorithm~\ref{alg:or-learners-app}. Therein: In stage \circled{0}, the \emph{representation network} consists of either (a)~a fully-connected (FC$_\phi$) or a normalizing flow (NF$_\phi$) representation subnetwork, and (b)~a fully-connected (FC$_a$) outcome subnetwork. Here, \emph{any} representation learning method can be used, and, depending on the method, additional components might be added (\eg, a propensity subnetwork for CFR-ISW). Then, in stage \circled{1}, we might need to additionally fit nuisance functions (\eg, when the constrained representations were used in stage \circled{0}, so that $\hat{\mu}^\phi_a$ is inconsistent wrt. $\hat{\mu}^x_a$). Therein, we might optionally employ two additional networks, namely, a \emph{propensity network} FC$_{\pi, x}$ and an \emph{outcome network} FC$_{\mu, x}$.  Finally, in stage \circled{2}, we utilize different DR- and R-losses, as presented in Sec.~\ref{sec:prelim}, to fit a fully-connected \emph{target network} $g$ and thus yield a final estimator of CAPOs/CATE. 

\textbf{Implementation.} We implemented the \ORlearners in PyTorch and Pyro. For better compatibility, the fully-connected subnetworks have one hidden layer with a tunable number of units and a RELU activation function. For the representation subnetworks involving normalizing flows, we employed residual normalizing flows \citep{chen2019residual} that have three hidden layers with a tunable synchronous number of units. All the networks for the \ORlearners (see stages \circled{0}--\circled{2} in Fig.~\ref{fig:or-learner-overview}) are trained with AdamW \citep{loshchilov2019decoupled}. Each network was trained with $n_\text{epoch} = 200$ epochs for the synthetic dataset and $n_\text{epoch} = 50$ for the ACIC 2016 dataset collection. To further stabilize training of the target networks in stage \circled{2}, we (i)~used exponential moving average (EMA) of model weights \citep{polyak1992acceleration} with a smoothing hyperparameter ($\lambda = 0.995$); and (ii)~clipped too low propensity scores ($\hat{\pi}_a^x(X) < 0.05$). 

\textbf{Hyperparameters.} We performed hyperparameter tuning of the \ORlearners (at stages~\circled{0} and \circled{1}) and other non-neural Neyman-orthogonal learners (at stage~\circled{1}) based on five-fold cross-validation using the training subset. At both stages, we did a random grid search with respect to different tuning criteria. For the final stage~\circled{2}, on the other hand, we used fixed hyperparameters for all the experiments, as an exact hyperparameter search is not possible for target models solely with the observational data \citep{curth2023search}. Table~\ref{tab:hyperparams} provides all the details on hyperparameter tuning. For reproducibility, we made the tuned hyperparameters available in our GitHub.\footnote{\url{https://github.com/Valentyn1997/OR-learners}.}

\newpage
\begin{figure}[th]
    \centering
    \vspace{-0.2cm}
    \begin{algorithm}[H]
        \caption{Pseudocode of the \ORlearners}\label{alg:or-learners-app}
        \begin{algorithmic}[1]
        \scriptsize
            \State {\bfseries Input:} Training dataset $\mathcal{D}$; strength of the balancing constraint $\alpha \ge 0$; $\operatorname{dist} \in \{\operatorname{WM}, \operatorname{MMD}\}$
            \State {\bfseries Stage }\circled{0}: Fit a representation network $\in \{$TARNet/TARFlow, CFR/CFRFlow, RCFR/RCFRFlow, BNN/BNNFlow, CFR-ISW/CFRFlow-ISW, BWCFR/BWCFRFlow$\}$ 
            \Indent
                \If{Representation network $\in \{$BWCFR/BWCFRFlow$\}$}
                    \State Fit a propensity network (FC$_{\pi, x}$) by minimizing a BCE loss $\hat{\mathcal{L}}_\pi$ and set $\hat{\pi}_a^x(X) \gets$ FC$_{\pi, x}(X)$
                \EndIf
                \For{$i$ = 0 {\bfseries to} $\lceil n_{\text{epochs}} \cdot n / b_{\text{R}} \rceil$}
                    \State Draw a minibatch $\mathcal{B} = \{X, A, Y\}$ of size $b_{\text{R}}$ from $\mathcal{D}$
                    \State {\bfseries Initialize:} $W \gets \mathbbm{1}_{b_R}; \quad \hat{\mathcal{L}}_\pi \gets 0; \quad \hat{\mathcal{L}}_\text{Bal} \gets 0$
                    \State $\Phi \gets $ NF$_\phi$ / FC$_\phi$$(X)$
                    \State $h_a(\Phi) \gets $ FC$_a$$(\Phi, a)$
                    \If{Representation network $\in \{$CFR-ISW/CFRFlow-ISW$\}$}
                        \State $\hat{\pi}_a^\phi(\Phi) \gets$ FC$_{\pi, \phi}(\operatorname{detach}(\Phi))$
                        \State $\hat{\mathcal{L}}_\pi \gets \operatorname{BCE}(\hat{\pi}_A^\phi(\Phi), A)$
                        \State $W \gets \operatorname{detach}\big({\mathbbm{1}\{\hat{\pi}_A^\phi(\Phi) \ge 0.05\}} / {\hat{\pi}_A^\phi(\Phi)} \big)$
                    \ElsIf{Representation network $\in \{$BWCFR/BWCFRFlow$\}$}
                        \State $W \gets {\mathbbm{1}\{\hat{\pi}_A^x(X) \ge 0.05\}} / {\hat{\pi}_A^x(X)}$
                    \ElsIf{Representation network $\in \{$RCFR/RCFRFlow$\}$}
                        \State $W \gets$ FC$_w(\operatorname{detach}(\Phi))$
                    \EndIf
                    \State $\hat{\mathcal{L}}_{W(\mathit{\Phi})} \gets \mathbb{P}_{b_R} \{ W (Y - h_A(\Phi(X)))^2\} \big/ \mathbb{P}_{b_R} \{ W\}$
                    \If{Representation network $\notin \{$TARNet/TARFlow$\}$ and $\alpha > 0 $}
                        \State $\hat{\mathcal{L}}_\text{Bal} \gets W$-weighted $ \widehat{\operatorname{dist}}(\mathbb{P}(\Phi(X) \mid A = 0), \mathbb{P}(\Phi(X) \mid A = 1))$
                    \EndIf
                    \State Gradient update of the representation network wrt. $\hat{\mathcal{L}}_{W(\mathit{\Phi})} + \alpha \hat{\mathcal{L}}_\text{Bal} + \hat{\mathcal{L}}_\pi$ 
                \EndFor
                \State $V \gets X \, \big/ \, \hat{\Phi}(X) \, \big/ \, (\hat{h}_0(\hat{\Phi}(X)), \hat{h}_1(\hat{\Phi}(X)))$
            \EndIndent
            \State {\bfseries Stage }\circled{1}: Estimate nuisance functions $\hat{\eta} = (\hat{\mu}_a^x, \hat{\pi}_a^x)$
            \Indent
                \If{Representation network $\notin \{$BWCFR/BWCFRFlow$\}$}
                    \State Fit a propensity network (FC$_{\pi, x}$) by minimizing a BCE loss $\hat{\mathcal{L}}_\pi$ and set $\hat{\pi}_a^x(X) \gets$ FC$_{\pi, x}(X)$
                \EndIf
                \If{$\alpha > 0$ and FC$_\phi$ is used at stage \circled{0}}
                    \State Fit an outcome network (FC$_{\mu, x}$) by minimizing an unweighted MSE loss $\hat{\mathcal{L}}_{\mathit{\Phi}}$ and set $\hat{\mu}_a^x(X) \gets$ FC$_{\mu, x}(X, a)$ 
                \Else 
                    \State Set $\hat{\mu}_a^x(X) \gets \hat{\mu}_a^\phi(\Phi(X))$
                \EndIf
            \EndIndent
            \State {\bfseries Stage }\circled{2}: Fit a target network $\hat{g} = \argmin \hat{\mathcal{L}}_\diamond(g, \hat{\eta})$ 
            \Indent
                \For{$i$ = 0 {\bfseries to} $\lceil n_{\text{epochs}} \cdot n / b_{\text{T}} \rceil$}
                    \State Draw a minibatch $\mathcal{B} = \{X, A, Y\}$ of size $b_{\text{T}}$ from $\mathcal{D}$
                    \State $\alpha_a(A, X) \gets {\mathbbm{1}\{A = a\} \cdot \mathbbm{1}\{\hat{\pi}_a^x(X) \ge 0.05\}}/{\hat{\pi}^x_a(X)}$
                    \If{Causal quantity $==$ CAPO}
                        \State $\hat{\mathcal{L}}_{\mathcal{G}}^{\text{DR}^\text{K}_a}(g, \hat{\eta}) \gets \mathbb{P}_{b_{\text{T}}} \big\{ \big( \alpha_a(A, X) \big( Y - \hat{\mu}_a^x(X) \big) + \hat{\mu}_a^x(X) - g(V)\big)^2 \big\}$
                        \State $\hat{\mathcal{L}}_{\mathcal{G}}^{\text{DR}^\text{FS}_a}(g, \hat{\eta}) \gets \mathbb{P}_{b_{\text{T}}} \big\{ \alpha_a(A, X) \big( Y - g(V)\big)^2  +  \big(1 - \alpha_a(A, X)\big) \big(\hat{\mu}_a^x(X) - g(V)\big)^2  \big\}$
                    \EndIf
                    \If{Causal quantity $==$ CATE}
                        \State $\hat{\mathcal{L}}_{\mathcal{G}}^{\text{DR}^\text{K}}(g, \hat{\eta}) \gets \mathbb{P}_{b_{\text{T}}} \big\{ \big(\alpha_0(A, X) \big( Y - \hat{\mu}_0^x(X) \big) + \alpha_1(A, X) \big( Y - \hat{\mu}_1^x(X) \big) + \hat{\mu}_1^x(X) - \hat{\mu}_0^x(X) - g(V)\big)^2 \big\}$
                        \State $\hat{\mathcal{L}}_{\mathcal{G}}^{\text{R}}(g, \hat{\eta}) \gets \mathbb{P}_{b_{\text{T}}} \big\{\big(\big(Y - \hat{\mu}^x(X)\big) - \big(A - \hat{\pi}_1^x(X)\big) g(V)\big)^2 \big\}$
                        \State $\hat{\mathcal{L}}_{\mathcal{G}}^{\text{IVW}}(g, \hat{\eta}) \gets \mathbb{P}_{b_{\text{T}}} \big\{\big(A - \hat{\pi}_1^x(X)\big)^2 \big(\big(\alpha_0(A, X) \big( Y - \hat{\mu}_0^x(X) \big) + \alpha_1(A, X) \big( Y - \hat{\mu}_1^x(X) \big) + \hat{\mu}_1^x(X) - \hat{\mu}_0^x(X) - g(V)\big)^2 \big\}$
                    \EndIf
                    \State Gradient \& EMA update of the target network $g$ wrt. $\hat{\mathcal{L}}_{\mathcal{G}}(g, \hat{\eta})$ 
                \EndFor
            \EndIndent
            \State {\bfseries Output:} $V$-level estimator $\hat{g}$ for CAPOs/CATE 
        \end{algorithmic}
    \end{algorithm}
    \vspace{-0.2cm}
\end{figure}

\newpage
\begin{table}[H]
    \vspace{-0.2cm}
    \caption{Hyperparameter tuning for the \ORlearners and other baselines.}
    \label{tab:hyperparams}
    \vspace{-0.3cm}
    \begin{center}
    \scalebox{.8}{
        \begin{tabu}{l|l|l|r}
            \toprule
            Stage & Model & Hyperparameter & Range / Value \\
            \midrule
            \multirow{28}{*}{\textbf{Stage \protect\circled{0}}} & \multirow{8}{*}{\begin{tabular}{l}
                 TARNet/TARFlow \\ BNN/BNNFlow \\ CFR/CFRFlow \\ BWCFR/BWCFRFlow
            \end{tabular}} & Learning rate & 0.001, 0.005, 0.01\\
            && Minibatch size, $b_R$ & 32, 64, 128 \\
            && Weight decay & 0.0, 0.001, 0.01, 0.1 \\
            && Hidden units in NF$_\phi$ / FC$_\phi$ & $R \, d_x$, 1.5 $Rd_x$, 2 $Rd_x$ \\
            && Hidden units in FC$_a$ & $R \, d_\phi$, 1.5 $Rd_\phi$, 2 $Rd_\phi$ \\
            && Tuning strategy & random grid search with 50 runs \\
            && Tuning criterion & factual MSE loss \\ 
            && Optimizer & AdamW \\ 
            \cmidrule{2-4} & \multirow{11}{*}{CFR-ISW/CFRFlow-ISW} & Representation network learning rate & 0.001, 0.005, 0.01 \\
            && Propensity network learning rate & 0.001, 0.005, 0.01 \\
            && Minibatch size, $b_R$ & 32, 64, 128 \\
            && Representation network weight decay & 0.0, 0.001, 0.01, 0.1 \\
            && Propensity network weight decay & 0.0, 0.001, 0.01, 0.1 \\
            && Hidden units in NF$_\phi$ / FC$_\phi$ & $R \, d_x$, 1.5 $Rd_x$, 2 $Rd_x$ \\
            && Hidden units in FC$_a$ & $R \, d_\phi$, 1.5 $Rd_\phi$, 2 $Rd_\phi$ \\
            && Hidden units in FC$_{\pi,\phi}$ & $R \, d_\phi$, 1.5 $Rd_\phi$, 2 $Rd_\phi$ \\
            && Tuning strategy & random grid search with 50 runs \\
            && Tuning criterion & factual MSE loss + factual BCE loss \\ 
            && Optimizer & AdamW \\
            \cmidrule{2-4} & \multirow{9}{*}{RCFR/RCFRFlow} & Learning rate & 0.001, 0.005, 0.01\\
            && Minibatch size, $b_R$ & 32, 64, 128 \\
            && Weight decay & 0.0, 0.001, 0.01, 0.1 \\
            && Hidden units in NF$_\phi$ / FC$_\phi$ & $R \, d_x$, 1.5 $Rd_x$, 2 $Rd_x$ \\
            && Hidden units in FC$_a$ & $R \, d_\phi$, 1.5 $Rd_\phi$, 2 $Rd_\phi$ \\
            && Hidden units in FC$_w$ & $R \, d_\phi$, 1.5 $Rd_\phi$, 2 $Rd_\phi$ \\
            && Tuning strategy & random grid search with 50 runs \\
            && Tuning criterion & factual MSE loss \\ 
            && Optimizer & AdamW \\
            \midrule \multirow{22}{*}{\textbf{Stage \protect\circled{1}}} & \multirow{7}{*}{Propensity network} & Learning rate & 0.001, 0.005, 0.01\\
            && Minibatch size, $b_N$ & 32, 64, 128 \\
            && Weight decay & 0.0, 0.001, 0.01, 0.1 \\
            && Hidden units in FC$_{\pi, x}$ & $R \, d_{x}$, 1.5 $Rd_{x}$, 2 $Rd_{x}$ \\
            && Tuning strategy & random grid search with 50 runs \\
            && Tuning criterion & factual BCE loss \\ 
            && Optimizer & AdamW \\
            \cmidrule{2-4} & \multirow{7}{*}{Outcome network} & Learning rate & 0.001, 0.005, 0.01\\
            && Minibatch size, $b_N$ & 32, 64, 128 \\
            && Hidden units in FC$_{\mu, x}$ & $R \, d_x$, 1.5 $Rd_x$, 2 $Rd_x$ \\
            && Weight decay & 0.0, 0.001, 0.01, 0.1 \\
            && Tuning strategy & random grid search with 50 runs \\
            && Tuning criterion & factual MSE loss \\ 
            && Optimizer & AdamW \\
            \cmidrule{2-4} & \multirow{7}{*}{XGBoost} & Number of estimators, $n_{e,N}$ & 50, 100, 150\\
            && Maximum depth & 3, 6, 9, 12 \\
            && $L_1$ regularization, $\alpha$ & 40, 80, 120, 160 \\
            && Minimum sum of instance weight in a child & 0, 3, 6, 9 \\
            && Minimum loss reduction needed for a split & 1, 3, 5, 7, 9 \\
            && Tuning strategy & random grid search with 50 runs \\
            && Tuning criterion & factual MSE/BCE loss \\ 
            \midrule \multirow{12}{*}{\textbf{Stage \protect\circled{2}}} & \multirow{6}{*}{Target network} & Learning rate &0.005\\
            && Minibatch size, $b_T$ & 64 \\
            && EMA of model weights, $\lambda$ & 0.995 \\
            && Hidden units in $g$ & Hidden units in FC$_a$ \\
            && Tuning strategy & no tuning \\
            && Optimizer & AdamW \\
            \cmidrule{2-4} & \multirow{5}{*}{XGBoost} & Number of estimators, $n_{e,T}$ & $n_{e,N}$\\
            && Maximum depth & 6 \\
            && $L_1$ regularization, $\alpha$ & 0 \\
            && Minimum sum of instance weight in a child & 1 \\
            && Minimum loss reduction needed for a split & 0 \\
            \bottomrule
            \multicolumn{4}{l}{$R = 2$ (synthetic data), $R = 1$ (IHDP dataset), $R = 0.25$ (ACIC 2016 datasets collection)}
        \end{tabu}}
    \end{center}
    \vspace{-2.5cm}
\end{table}

\newpage
\section{ADDITIONAL EXPERIMENTS} \label{app:experiments}

\subsection[Setting 1]{Setting \circledred{1}}
\textbf{(i)~Synthetic data}. Table~\ref{tab:synthetic-setting-a} shows additional results for the synthetic dataset in Setting \circledred{1}. Therein, we observe that the \ORlearners with $V = \hat{\Phi}(X)$ are as effective as other variants (\eg, $V = X/X^*$). This was expected as, in the synthetic dataset \citep{melnychuk2024bounds}, the ground-truth CAPOs/CATE densely depend on the covariates $X$ ($d_x = 2$), and, thus, the low-manifold hypothesis (\eg, with $d_{{\phi}} = 1$) can not be assumed. Hence, all four variants of the \ORlearners perform similarly well, and all the improvements can be attributed to the Neyman-orthogonality of the \ORlearners. 

\begin{table}[ht]
    \vspace{-0.1cm}
    \begin{minipage}{\linewidth}
      \caption{\textbf{Results for synthetic experiments in Setting \protect\circledred{1}.} Reported: improvements of the \ORlearners over plug-in representation networks wrt. out-of-sample rMSE / rPEHE; mean $\pm$ std over 15 runs. Here, $n_{\text{train}} = 500, d_{\hat{\phi}} = 2$.} \label{tab:synthetic-setting-a}
      \vspace{-0.25cm}
      \begin{center}
            \scriptsize
            \scalebox{0.81}{\begin{tabu}{lr|cc|cc|ccc}
\toprule
 &  & $\text{DR}_0^{\text{K}}$ & $\text{DR}_0^{\text{FS}}$ & $\text{DR}_1^{\text{K}}$ & $\text{DR}_1^{\text{FS}}$ & $\text{DR}^{\text{K}}$ & $\text{R}$ & $\text{IVW}$ \\
\midrule
\multirow{4}{*}{TARNet} & $V = (\hat{\mu}^x_0,\hat{\mu}^x_1)$ & $-$0.002 $\pm$ 0.011 & $-$0.002 $\pm$ 0.016 & \textcolor{ForestGreen}{$-$0.004 $\pm$ 0.006} & \textcolor{ForestGreen}{$-$0.004 $\pm$ 0.006} & \textcolor{ForestGreen}{$-$0.006 $\pm$ 0.012} & \textcolor{ForestGreen}{$-$0.009 $\pm$ 0.013} & \textcolor{ForestGreen}{$-$0.009 $\pm$ 0.017} \\
 & $V = X$ & \textcolor{BrickRed}{$+$0.064 $\pm$ 0.034} & \textcolor{BrickRed}{$+$0.083 $\pm$ 0.051} & \textcolor{BrickRed}{$+$0.078 $\pm$ 0.053} & \textcolor{BrickRed}{$+$0.059 $\pm$ 0.037} & \textcolor{ForestGreen}{$-$0.018 $\pm$ 0.012} & \textcolor{ForestGreen}{$-$0.021 $\pm$ 0.018} & \textcolor{ForestGreen}{$-$0.021 $\pm$ 0.015} \\
 & $V = X^*$ & \textcolor{BrickRed}{$+$0.015 $\pm$ 0.022} & \textcolor{BrickRed}{$+$0.023 $\pm$ 0.020} & \textcolor{BrickRed}{$+$0.015 $\pm$ 0.031} & $+$0.004 $\pm$ 0.016 & \textcolor{ForestGreen}{$-$0.013 $\pm$ 0.017} & \textcolor{ForestGreen}{$-$0.017 $\pm$ 0.018} & \textcolor{ForestGreen}{$-$0.017 $\pm$ 0.016} \\
 & $V = \hat{\Phi}(X)$ & $-$0.002 $\pm$ 0.013 & $\pm$0.000 $\pm$ 0.017 & \textcolor{ForestGreen}{$-$0.004 $\pm$ 0.007} & \textcolor{ForestGreen}{$-$0.003 $\pm$ 0.006} & \textcolor{ForestGreen}{$-$0.011 $\pm$ 0.012} & \textcolor{ForestGreen}{$-$0.012 $\pm$ 0.009} & \textcolor{ForestGreen}{$-$0.012 $\pm$ 0.013} \\
\cmidrule(lr){1-9}
\multirow{4}{*}{BNN ($\alpha$ = 0.0)} & $V = (\hat{\mu}^x_0,\hat{\mu}^x_1)$ & \textcolor{ForestGreen}{$-$0.006 $\pm$ 0.014} & $+$0.001 $\pm$ 0.022 & \textcolor{ForestGreen}{$-$0.009 $\pm$ 0.009} & \textcolor{ForestGreen}{$-$0.009 $\pm$ 0.009} & \textcolor{ForestGreen}{$-$0.007 $\pm$ 0.009} & \textcolor{ForestGreen}{$-$0.006 $\pm$ 0.010} & \textcolor{ForestGreen}{$-$0.009 $\pm$ 0.013} \\
 & $V = X$ & \textcolor{BrickRed}{$+$0.067 $\pm$ 0.033} & \textcolor{BrickRed}{$+$0.101 $\pm$ 0.039} & \textcolor{BrickRed}{$+$0.045 $\pm$ 0.037} & \textcolor{BrickRed}{$+$0.037 $\pm$ 0.034} & \textcolor{ForestGreen}{$-$0.020 $\pm$ 0.018} & \textcolor{ForestGreen}{$-$0.023 $\pm$ 0.016} & \textcolor{ForestGreen}{$-$0.022 $\pm$ 0.019} \\
 & $V = X^*$ & \textcolor{BrickRed}{$+$0.011 $\pm$ 0.017} & \textcolor{BrickRed}{$+$0.023 $\pm$ 0.033} & $-$0.005 $\pm$ 0.016 & \textcolor{ForestGreen}{$-$0.008 $\pm$ 0.015} & $-$0.010 $\pm$ 0.035 & \textcolor{ForestGreen}{$-$0.017 $\pm$ 0.018} & \textcolor{ForestGreen}{$-$0.016 $\pm$ 0.025} \\
 & $V = \hat{\Phi}(X)$ & \textcolor{ForestGreen}{$-$0.008 $\pm$ 0.012} & $-$0.002 $\pm$ 0.016 & \textcolor{ForestGreen}{$-$0.010 $\pm$ 0.013} & \textcolor{ForestGreen}{$-$0.011 $\pm$ 0.011} & \textcolor{ForestGreen}{$-$0.012 $\pm$ 0.011} & \textcolor{ForestGreen}{$-$0.012 $\pm$ 0.013} & \textcolor{ForestGreen}{$-$0.015 $\pm$ 0.014} \\
\bottomrule
\multicolumn{8}{l}{Lower $=$ better. Significant improvement over the baseline in \textcolor{ForestGreen}{green}, significant worsening of the baseline in \textcolor{BrickRed}{red}}
\end{tabu}
}
        \end{center}
    \end{minipage}%
    \vspace{-0.1cm}
\end{table}

\textbf{(ii)~HC-MNIST dataset}. Table~\ref{tab:hcmnist-setting-a-full} shows the results for the HC-MNIST dataset. Here, we again report the absolute performance of different methods: non-neural Neyman-orthogonal learners instantiated with XGBoost \citep{chen2015xgboost} (with S-/T-learners for the first-stage models); plug-in representation learning methods; and the OR-learners used with the pre-trained representations $V = \hat{\Phi}(X)$. We see that the OR-learners improve over the other, non-neural Neyman-orthogonal learners: This is again not surprising as the manifold hypothesis holds for the HC-MNIST dataset. Furthermore, the \ORlearners outperform the baseline plug-in representation learning methods given \emph{sufficient overlap} in the HC-MNIST data: This contrasts with the IHDP dataset results (see Table~\ref{tab:ihdp-setting-a-full}), as the IHDP dataset contains extreme propensity scores \citep{curth2021really}.

\begin{table}[ht]
    \vspace{-0.1cm}
    \begin{minipage}{\linewidth}
      \caption{\textbf{Results for HC-MNIST experiments in Setting \protect\circledred{1}.} Reported: out-of-sample rMSE / rPEHE for different causal quantities ($\xi^x_a / \tau^x$, respectively), mean $\pm$ std over 10 runs. Here, $d_{\hat{\phi}} = 78$ for neural baselines.} \label{tab:hcmnist-setting-a-full}
      \vspace{-0.25cm}
      \begin{center}
            \scriptsize
            \scalebox{0.95}{\begin{tabu}{lr|c|c|c}
\toprule
 &  & $\xi_0^x$ & $\xi_1^x$ & $\tau^x$ \\
\midrule
\multirow{3}{*}{XGBoost S} & $\text{DR}^{\text{K}}$ & 0.610 $\pm$ 0.000 & 0.499 $\pm$ 0.000 & 0.766 $\pm$ 0.000 \\
 & $\text{R}$ & --- & --- & 0.702 $\pm$ 0.000 \\
 & $\text{IVW}$ & --- & --- & 0.723 $\pm$ 0.000 \\
\cmidrule(lr){1-5}
\multirow{3}{*}{XGBoost T} & $\text{DR}^{\text{K}}$ & 0.591 $\pm$ 0.000 & 0.498 $\pm$ 0.000 & 0.763 $\pm$ 0.000 \\
 & $\text{R}$ & --- & --- & 0.706 $\pm$ 0.000 \\
 & $\text{IVW}$ & --- & --- & 0.726 $\pm$ 0.000 \\
\cmidrule(lr){1-5}
\multirow{5}{*}{TARNet} & Plug-in & 0.514 $\pm$ 0.006 & 0.485 $\pm$ 0.003 & 0.697 $\pm$ 0.004 \\
 & $\text{DR}^{\text{K}}$ & \underline{0.503 $\pm$ 0.003} & \underline{0.471 $\pm$ 0.003} & 0.681 $\pm$ 0.004 \\
 & $\text{DR}^{\text{FS}}$ & 0.520 $\pm$ 0.051 & \underline{0.471 $\pm$ 0.006} & --- \\
 & $\text{R}$ & --- & --- & 0.684 $\pm$ 0.021 \\
 & $\text{IVW}$ & --- & --- & 0.682 $\pm$ 0.006 \\
\cmidrule(lr){1-5}
\multirow{5}{*}{BNN ($\alpha$ = 0.0)} & Plug-in & 0.526 $\pm$ 0.021 & 0.508 $\pm$ 0.023 & 0.700 $\pm$ 0.020 \\
 & $\text{DR}^{\text{K}}$ & 0.507 $\pm$ 0.010 & 0.474 $\pm$ 0.006 & 0.680 $\pm$ 0.010 \\
 & $\text{DR}^{\text{FS}}$ & \textbf{0.497 $\pm$ 0.005} & \textbf{0.468 $\pm$ 0.004} & --- \\
 & $\text{R}$ & --- & --- & \textbf{0.673 $\pm$ 0.008} \\
 & $\text{IVW}$ & --- & --- & \underline{0.678 $\pm$ 0.009} \\
\cmidrule(lr){1-5}
Oracle &  & 0.363 & 0.365 & 0.513 \\
\bottomrule
\multicolumn{4}{l}{Lower $=$ better. Best in \textbf{bold}, second best \underline{underlined} }
\end{tabu}
}
        \end{center}
    \end{minipage}%
    \vspace{-0.1cm}
\end{table}

\newpage
\subsection[Setting 2]{Setting \circledred{2}}

\begin{figure}[ht]
    \centering
    \vspace{-0.2cm}
    \includegraphics[width=0.49\linewidth]{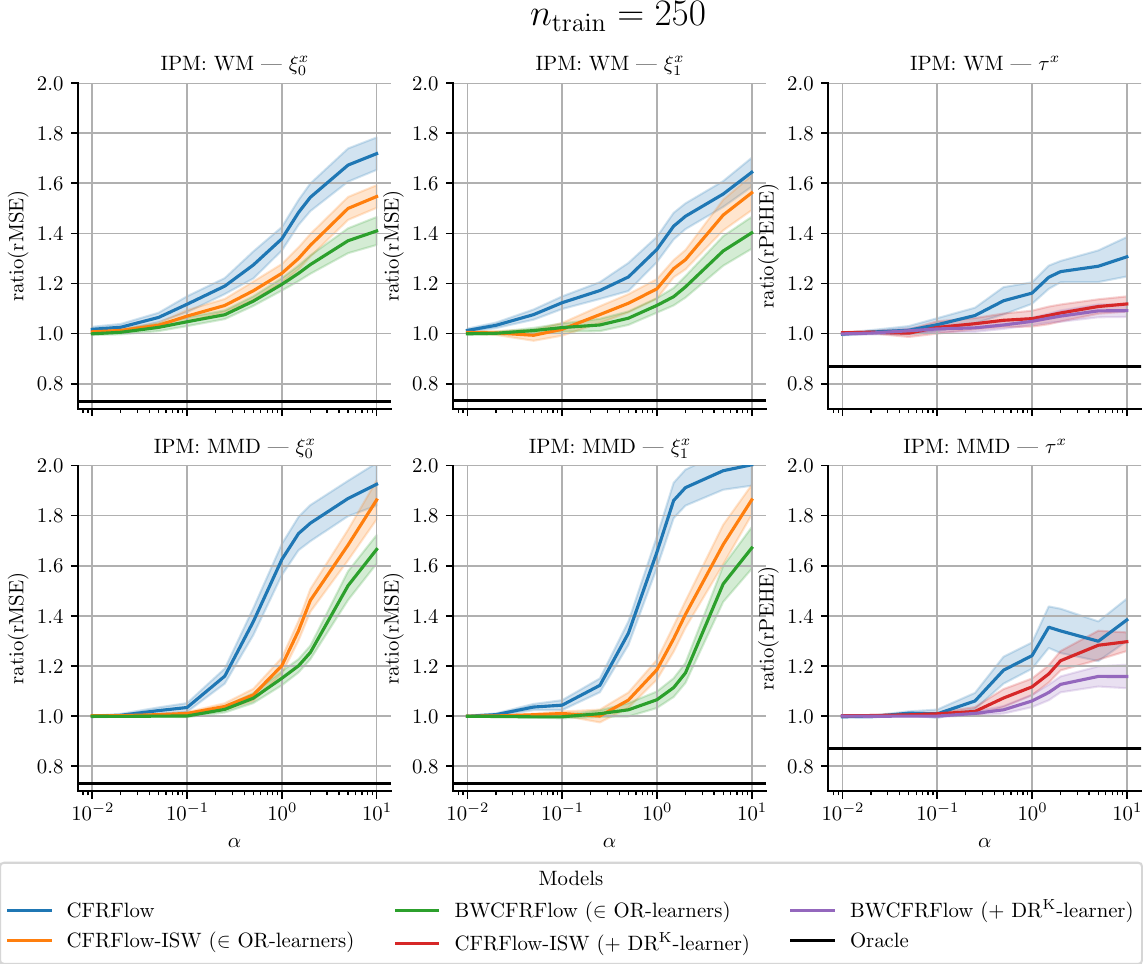} 
    \hfill
    \includegraphics[width=0.49\linewidth]{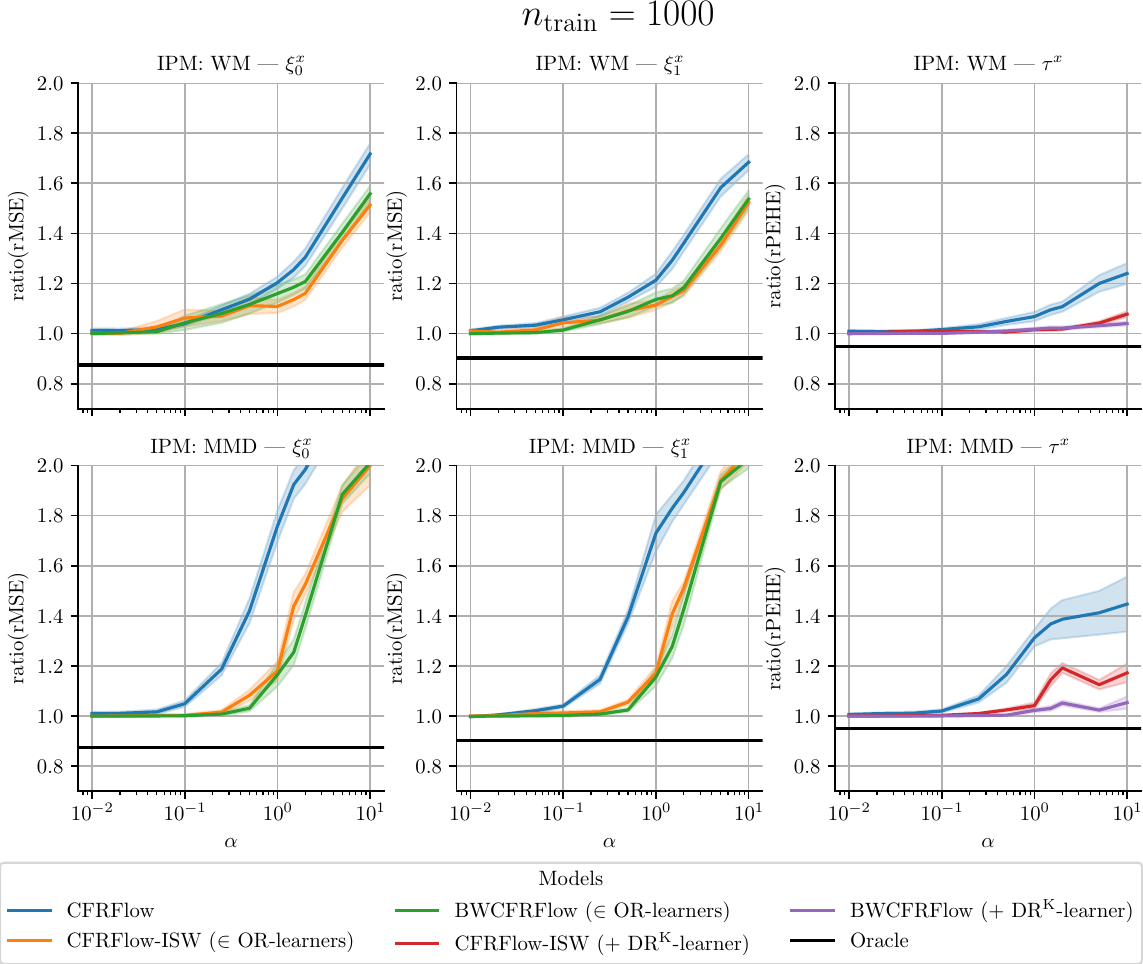} 
    \vspace{-0.25cm}
    \caption{\textbf{Results for synthetic data in Setting \protect\circledred{2}.} Reported: ratio between the performance of TARFlow (CFRFlow with $\alpha = 0$) and invertible representation networks with varying $\alpha$; mean $\pm$ SE over 15 runs. Lower is better. Here: $n_{\text{train}} \in \{250, 1000 \}$, $d_{\hat{\phi}} = 2$.}
    \label{fig:synthetic-setting-b-full}
    \vspace{-0.1cm}
\end{figure}

\textbf{(i)~Synthetic data}. We report additional results for the synthetic dataset in Fig.~\ref{fig:synthetic-setting-b-full} and Table~\ref{tab:synthetic-setting-c} for invertible and non-invertible representations, respectively. Fig.~\ref{fig:synthetic-setting-b-full} empirically demonstrates our intuition from Fig.~\ref{fig:or-learner-summary} of Appendix~\ref{app:visual-summary}. Specifically, as the data size grows, the \ORlearners manage to correct the RICB of the baseline representation learning methods more effectively (even for large values of the balancing strength $\alpha$). Notably, for this dataset, the balancing constraint acts detrimentally as the heterogeneity of the CAPOs/CATE is high for both low and high overlap regions of the covariate space (thus, the underlying inductive bias ``low overlap -- low heterogeneity'' cannot be assumed). Table~\ref{tab:synthetic-setting-c} then provides similar evidence for the non-invertible representations. Here, the \ORlearners achieve significant improvements in the majority of the cases (and no significant worsenings). 

\begin{figure}[H]
    \vspace{-0.2cm}
    \centering
    \includegraphics[width=0.48\linewidth]{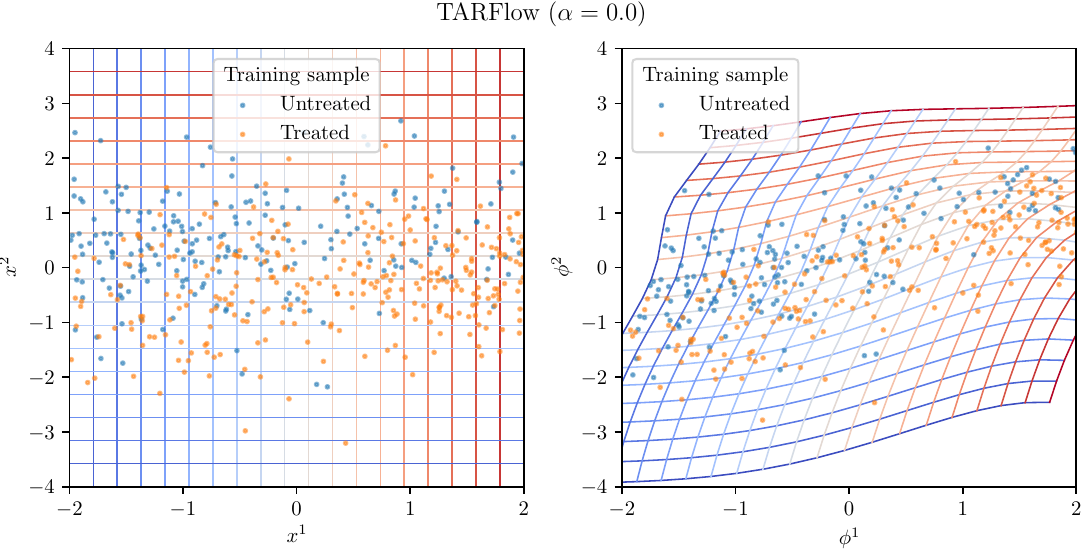} \\
    \vspace{0.1cm}
    \hrule
    \vspace{0.1cm}
    \includegraphics[width=0.48\linewidth]{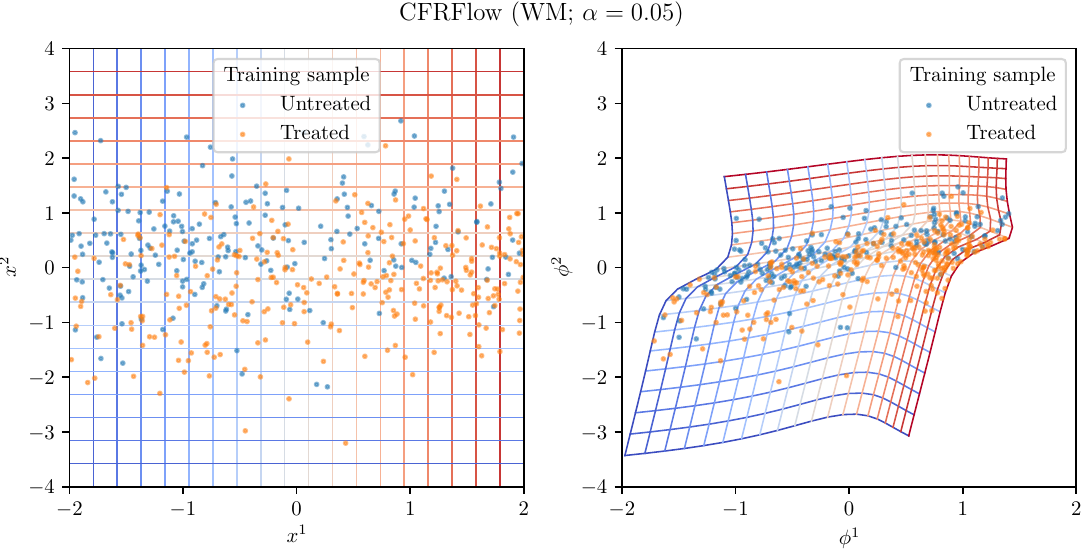} 
    \hspace{0.1cm} \vrule \hspace{0.1cm}
    \includegraphics[width=0.48\linewidth]{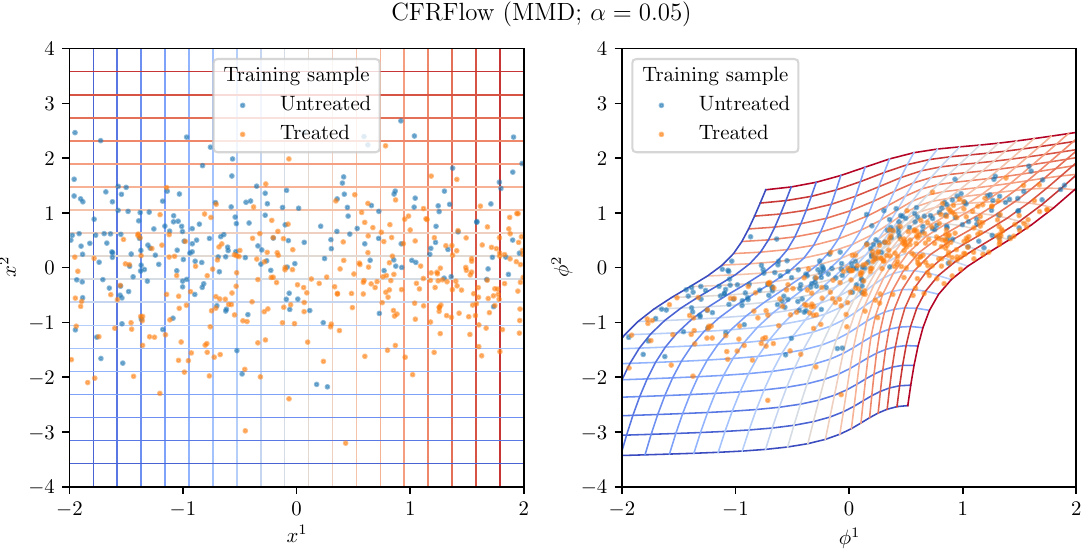} 
    \vspace{0.1cm}
    \hrule
    \vspace{0.1cm}
    \includegraphics[width=0.48\linewidth]{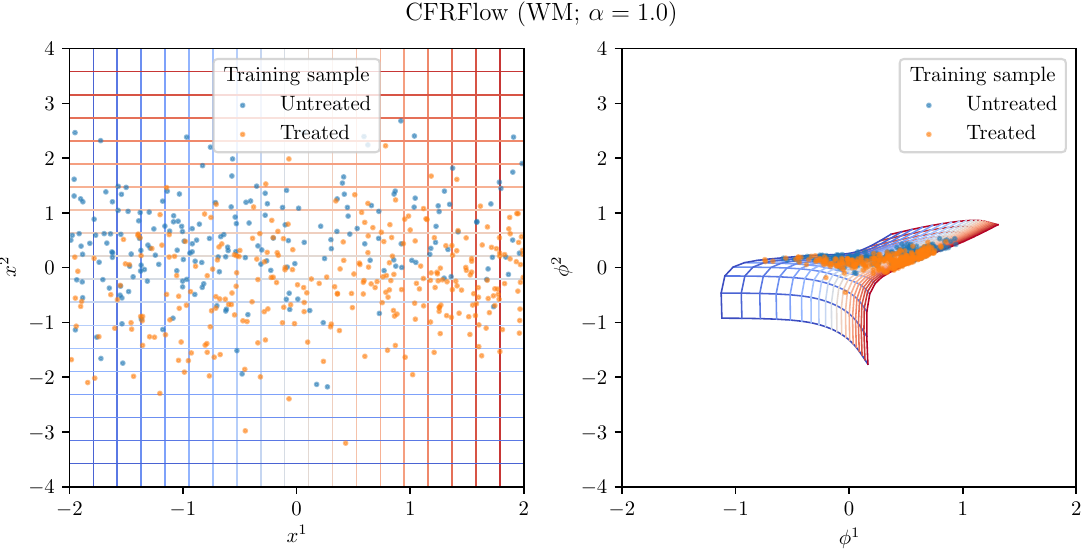} 
    \hspace{0.1cm} \vrule \hspace{0.1cm}
    \includegraphics[width=0.48\linewidth]{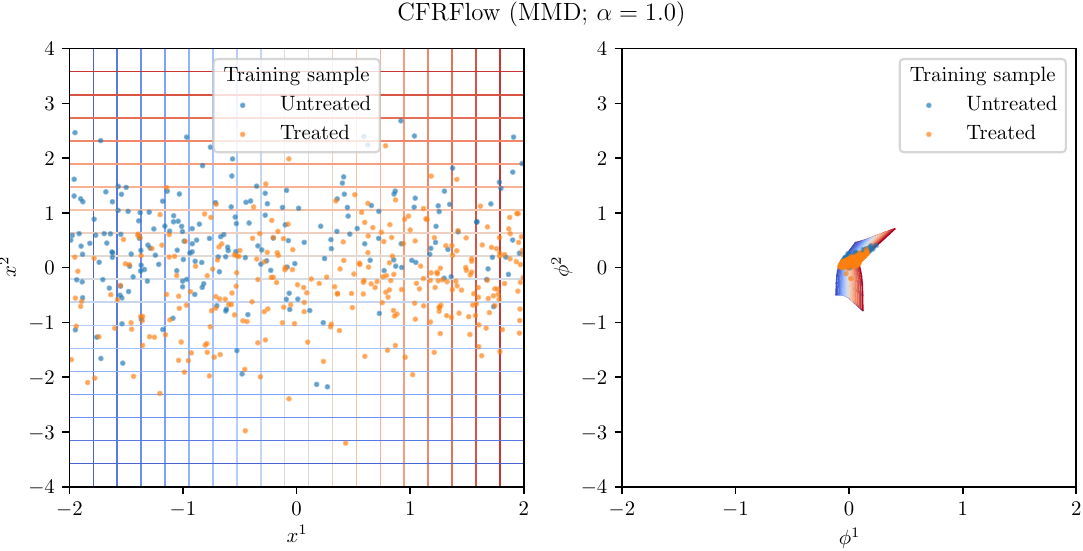} 
    \vspace{-0.2cm}
    \caption{{Visualization of the invertible transformations defined by the learned normalizing flow representation subnetworks for synthetic experiments in Setting \protect\circledred{2}. Here, $n_\text{train} = 500, d_{\hat{\phi}}=2$. Specifically, we show how a grid in the original covariate space, $\mathcal{X} \subseteq \mathbb{R}^2$, gets transformed onto the representation space, $\mathit{\hat{\Phi}}\subseteq \mathbb{R}^2$. We vary the strength of balancing $\alpha \in \{0, 0.05, 1.0\}$ and the IPM $\in \{$WM, MMD$\}$. As suggested by the theory in Sec.~\ref{sec:rq2}, the covariate space gets \emph{expanded} for $\alpha = 0$ and gets \emph{contracted} for large values of $\alpha$ (\eg, $\alpha = 1$).}}
    \label{fig:setting-b-scaling}
    \vspace{-0.2cm}
\end{figure}

In Fig.~\ref{fig:setting-b-scaling}, we additionally show how the learned normalizing flows transform the original space $\mathcal{X}$ (the models are the same as in Fig.~\ref{fig:synthetic-setting-b}). The rendered transformations match the theoretical results provided in Sec.~\ref{sec:rq2}. Specifically, TARFlow expands (scales up) the original space so that the regression task becomes easier in the representation space. At the same time, CRFFlows with different balancing hyperparameters $\alpha$ aim to contract (scale down) the space, thus achieving better balancing.

\begin{table}[ht]
    \vspace{-0.1cm}
    \begin{minipage}{\linewidth}
      \caption{\textbf{Results for synthetic experiments in Setting \protect\circledred{2}.} Reported: improvements of the \ORlearners over non-invertible plug-in / IPTW representation networks wrt. out-of-sample rMSE / rPEHE; mean $\pm$ std over 15 runs. Here, $n_{\text{train}} = 500, d_{\hat{\phi}} = 2$.} \label{tab:synthetic-setting-c}
      \vspace{-0.25cm}
      \begin{center}
            \scriptsize
            \scalebox{0.84}{\begin{tabu}{l|cc|cc|ccc}
\toprule
 & $\text{DR}_0^{\text{K}}$ & $\text{DR}_0^{\text{FS}}$ & $\text{DR}_1^{\text{K}}$ & $\text{DR}_1^{\text{FS}}$ & $\text{DR}^{\text{K}}$ & $\text{R}$ & $\text{IVW}$ \\
\midrule
CFR (MMD; $\alpha$ = 0.1) & $-$0.006 $\pm$ 0.024 & $-$0.005 $\pm$ 0.026 & $-$0.009 $\pm$ 0.026 & \textcolor{ForestGreen}{$-$0.014 $\pm$ 0.022} & $-$0.011 $\pm$ 0.039 & \textcolor{ForestGreen}{$-$0.017 $\pm$ 0.032} & $-$0.012 $\pm$ 0.042 \\
CFR (WM; $\alpha$ = 0.1) & $-$0.003 $\pm$ 0.016 & \textcolor{ForestGreen}{$-$0.006 $\pm$ 0.014} & \textcolor{ForestGreen}{$-$0.005 $\pm$ 0.010} & \textcolor{ForestGreen}{$-$0.006 $\pm$ 0.005} & $-$0.001 $\pm$ 0.023 & $-$0.005 $\pm$ 0.023 & $-$0.004 $\pm$ 0.027 \\
BNN (MMD; $\alpha$ = 0.1) & \textcolor{ForestGreen}{$-$0.058 $\pm$ 0.047} & \textcolor{ForestGreen}{$-$0.051 $\pm$ 0.046} & \textcolor{ForestGreen}{$-$0.011 $\pm$ 0.012} & $-$0.006 $\pm$ 0.018 & \textcolor{ForestGreen}{$-$0.048 $\pm$ 0.041} & \textcolor{ForestGreen}{$-$0.038 $\pm$ 0.043} & \textcolor{ForestGreen}{$-$0.039 $\pm$ 0.040} \\
BNN (WM; $\alpha$ = 0.1) & $+$0.016 $\pm$ 0.101 & $-$0.013 $\pm$ 0.035 & $-$0.005 $\pm$ 0.037 & $+$0.007 $\pm$ 0.043 & \textcolor{ForestGreen}{$-$0.026 $\pm$ 0.042} & \textcolor{ForestGreen}{$-$0.026 $\pm$ 0.041} & \textcolor{ForestGreen}{$-$0.025 $\pm$ 0.041} \\
RCFR (MMD; $\alpha$ = 0.1) & $-$0.010 $\pm$ 0.086 & \textcolor{ForestGreen}{$-$0.032 $\pm$ 0.034} & \textcolor{ForestGreen}{$-$0.012 $\pm$ 0.019} & \textcolor{ForestGreen}{$-$0.012 $\pm$ 0.020} & \textcolor{ForestGreen}{$-$0.040 $\pm$ 0.043} & \textcolor{ForestGreen}{$-$0.028 $\pm$ 0.038} & \textcolor{ForestGreen}{$-$0.034 $\pm$ 0.042} \\
RCFR (WM; $\alpha$ = 0.1) & $-$0.008 $\pm$ 0.020 & \textcolor{ForestGreen}{$-$0.009 $\pm$ 0.019} & $-$0.003 $\pm$ 0.015 & $-$0.006 $\pm$ 0.015 & \textcolor{ForestGreen}{$-$0.019 $\pm$ 0.021} & \textcolor{ForestGreen}{$-$0.015 $\pm$ 0.022} & \textcolor{ForestGreen}{$-$0.019 $\pm$ 0.022} \\
CFR-ISW (MMD; $\alpha$ = 0.1) & $+$0.002 $\pm$ 0.025 & $-$0.003 $\pm$ 0.016 & $-$0.002 $\pm$ 0.009 & \textcolor{ForestGreen}{$-$0.008 $\pm$ 0.008} & $+$0.001 $\pm$ 0.023 & $-$0.002 $\pm$ 0.014 & $-$0.001 $\pm$ 0.017 \\
CFR-ISW (WM; $\alpha$ = 0.1) & $+$0.001 $\pm$ 0.029 & $-$0.006 $\pm$ 0.017 & $-$0.004 $\pm$ 0.018 & $-$0.003 $\pm$ 0.029 & \textcolor{ForestGreen}{$-$0.009 $\pm$ 0.017} & \textcolor{ForestGreen}{$-$0.008 $\pm$ 0.014} & \textcolor{ForestGreen}{$-$0.010 $\pm$ 0.016} \\
BWCFR (MMD; $\alpha$ = 0.1) & $+$0.007 $\pm$ 0.079 & $-$0.003 $\pm$ 0.018 & $-$0.005 $\pm$ 0.014 & $-$0.003 $\pm$ 0.012 & \textcolor{ForestGreen}{$-$0.015 $\pm$ 0.016} & \textcolor{ForestGreen}{$-$0.017 $\pm$ 0.012} & \textcolor{ForestGreen}{$-$0.018 $\pm$ 0.011} \\
\bottomrule
\multicolumn{8}{l}{Lower $=$ better. Significant improvement over the baseline in \textcolor{ForestGreen}{green}, significant worsening of the baseline in \textcolor{BrickRed}{red}}
\end{tabu}
}
        \end{center}
    \end{minipage}%
    \vspace{-0.1cm}
\end{table}

\newpage

\begin{figure}[ht]
    \centering
    \vspace{-0.2cm}
    \includegraphics[width=0.55\linewidth]{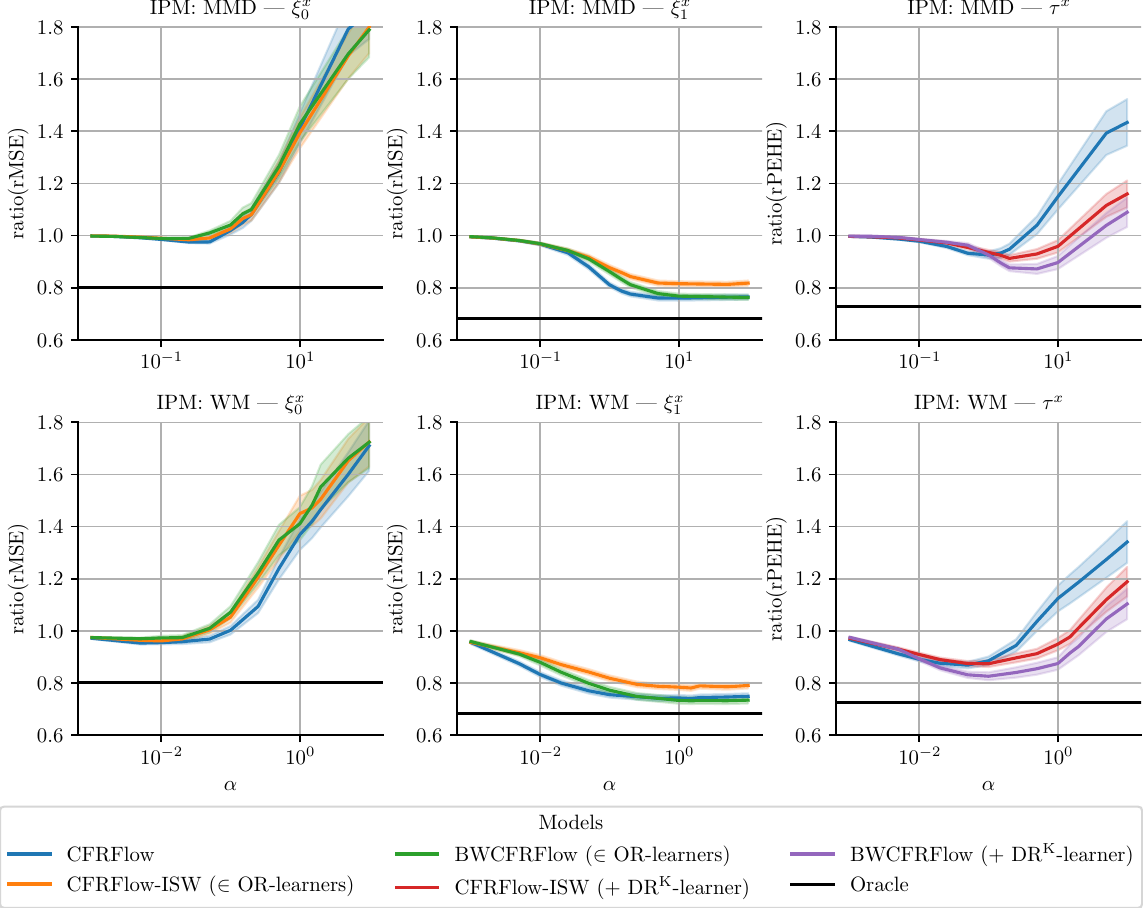} 
    \vspace{-0.25cm}
    \caption{\textbf{Results for IHDP experiments in Setting \protect\circledred{2}.} Reported: ratio between the performance of TARFlow (CFRFlow with $\alpha = 0$) and invertible representation networks with varying $\alpha$; mean $\pm$ SE over 100 train/test splits. Lower is better. Here: $d_{\hat{\phi}} = 12$.}
    \label{fig:ihdp-setting-b}
    \vspace{-0.1cm}
\end{figure}

\textbf{(ii)~IHDP dataset}. Fig.~\ref{fig:ihdp-setting-b} shows the results for the IHDP dataset in Setting~\circledred{2} for invertible representations. Here, interestingly, balancing in CFRFlow seems to outperform the \ORlearners for some values of $\alpha$. This is not surprising, as the IHDP dataset contains strong overlap violations and one of the ground-truth CAPOs is linear (namely, $\xi_1^x$). Hence, the ``low overlap -- low heterogeneity'' inductive bias partially holds for this dataset. Note, however, that the optimal $\alpha$ is different for both CAPOs and CATE, which renders balancing impractical (considering its value cannot be reliably tuned just with the observational data). 

\textbf{(iii)~HC-MNIST dataset}. Finally, in Table~\ref{tab:hcmnist-setting-c} we showcase the results of the HC-MNIST experiments for non-invertible representations with balancing. Here, the \ORlearners significantly improve over the baseline representation learning methods with the balancing constraint in the majority of the cases (and do not significantly worsen them). This was expected, as the ground-truth CAPOs for the HC-MNIST dataset are highly heterogeneous regardless of the overlap.  

\begin{table}[ht]
    \vspace{-0.1cm}
    \begin{minipage}{\linewidth}
      \caption{\textbf{Results for HC-MNIST experiments in Setting \protect\circledred{2}.} Reported: improvements of the \ORlearners over non-invertible plug-in / IPTW representation networks wrt. out-of-sample rMSE / rPEHE; mean $\pm$ std over 10 runs. Here, $d_{\hat{\phi}} = 78$.} \label{tab:hcmnist-setting-c}
      \vspace{-0.25cm}
      \begin{center}
            \scriptsize
            \scalebox{0.84}{\begin{tabu}{l|cc|cc|ccc}
\toprule
 & $\text{DR}_0^{\text{K}}$ & $\text{DR}_1^{\text{K}}$ & $\text{DR}_0^{\text{FS}}$ & $\text{DR}_1^{\text{FS}}$ & $\text{DR}^{\text{K}}$ & $\text{IVW}$ & $\text{R}$ \\
\midrule
CFR (MMD; $\alpha$ = 0.1) & \textcolor{ForestGreen}{$-$0.025 $\pm$ 0.007} & \textcolor{ForestGreen}{$-$0.011 $\pm$ 0.004} & \textcolor{ForestGreen}{$-$0.015 $\pm$ 0.008} & \textcolor{ForestGreen}{$-$0.014 $\pm$ 0.005} & \textcolor{ForestGreen}{$-$0.023 $\pm$ 0.009} & \textcolor{ForestGreen}{$-$0.022 $\pm$ 0.008} & \textcolor{ForestGreen}{$-$0.016 $\pm$ 0.009} \\
CFR (WM; $\alpha$ = 0.1) & \textcolor{ForestGreen}{$-$0.062 $\pm$ 0.010} & \textcolor{ForestGreen}{$-$0.021 $\pm$ 0.006} & \textcolor{ForestGreen}{$-$0.049 $\pm$ 0.011} & \textcolor{ForestGreen}{$-$0.018 $\pm$ 0.006} & \textcolor{ForestGreen}{$-$0.035 $\pm$ 0.008} & \textcolor{ForestGreen}{$-$0.027 $\pm$ 0.008} & \textcolor{ForestGreen}{$-$0.024 $\pm$ 0.008} \\
BNN (MMD; $\alpha$ = 0.1) & $-$0.069 $\pm$ 0.174 & $-$0.061 $\pm$ 0.132 & $-$0.080 $\pm$ 0.175 & \textcolor{ForestGreen}{$-$0.068 $\pm$ 0.130} & $-$0.025 $\pm$ 0.059 & $-$0.023 $\pm$ 0.069 & \textcolor{ForestGreen}{$-$0.035 $\pm$ 0.055} \\
BNN (WM; $\alpha$ = 0.1) & \textcolor{ForestGreen}{$-$0.064 $\pm$ 0.012} & \textcolor{ForestGreen}{$-$0.024 $\pm$ 0.004} & \textcolor{ForestGreen}{$-$0.054 $\pm$ 0.011} & \textcolor{ForestGreen}{$-$0.022 $\pm$ 0.003} & \textcolor{ForestGreen}{$-$0.050 $\pm$ 0.018} & \textcolor{ForestGreen}{$-$0.043 $\pm$ 0.016} & \textcolor{ForestGreen}{$-$0.040 $\pm$ 0.018} \\
RCFR (MMD; $\alpha$ = 0.1) & \textcolor{ForestGreen}{$-$0.106 $\pm$ 0.039} & \textcolor{ForestGreen}{$-$0.040 $\pm$ 0.018} & $+$0.044 $\pm$ 0.387 & \textcolor{ForestGreen}{$-$0.042 $\pm$ 0.020} & \textcolor{ForestGreen}{$-$0.090 $\pm$ 0.039} & \textcolor{ForestGreen}{$-$0.056 $\pm$ 0.073} & $+$0.013 $\pm$ 0.178 \\
RCFR (WM; $\alpha$ = 0.1) & \textcolor{ForestGreen}{$-$0.405 $\pm$ 0.458} & $-$0.178 $\pm$ 0.384 & \textcolor{ForestGreen}{$-$0.406 $\pm$ 0.447} & $-$0.148 $\pm$ 0.406 & \textcolor{ForestGreen}{$-$0.233 $\pm$ 0.345} & \textcolor{ForestGreen}{$-$0.225 $\pm$ 0.343} & \textcolor{ForestGreen}{$-$0.210 $\pm$ 0.359} \\
CFR-ISW (MMD; $\alpha$ = 0.1) & \textcolor{ForestGreen}{$-$0.010 $\pm$ 0.007} & \textcolor{ForestGreen}{$-$0.003 $\pm$ 0.004} & \textcolor{ForestGreen}{$-$0.006 $\pm$ 0.006} & \textcolor{ForestGreen}{$-$0.007 $\pm$ 0.002} & \textcolor{ForestGreen}{$-$0.007 $\pm$ 0.008} & $-$0.004 $\pm$ 0.010 & \textcolor{ForestGreen}{$-$0.008 $\pm$ 0.008} \\
CFR-ISW (WM; $\alpha$ = 0.1) & \textcolor{ForestGreen}{$-$0.019 $\pm$ 0.009} & \textcolor{ForestGreen}{$-$0.014 $\pm$ 0.007} & $-$0.007 $\pm$ 0.017 & \textcolor{ForestGreen}{$-$0.013 $\pm$ 0.007} & \textcolor{ForestGreen}{$-$0.024 $\pm$ 0.005} & \textcolor{ForestGreen}{$-$0.022 $\pm$ 0.006} & \textcolor{ForestGreen}{$-$0.021 $\pm$ 0.005} \\
BWCFR (MMD; $\alpha$ = 0.1) & $-$0.008 $\pm$ 0.019 & $+$0.043 $\pm$ 0.163 & \textcolor{ForestGreen}{$-$0.012 $\pm$ 0.006} & \textcolor{ForestGreen}{$-$0.011 $\pm$ 0.004} & $+$0.005 $\pm$ 0.063 & \textcolor{ForestGreen}{$-$0.013 $\pm$ 0.011} & $+$0.020 $\pm$ 0.126 \\
\bottomrule
\multicolumn{8}{l}{Lower $=$ better. Significant improvement over the baseline in \textcolor{ForestGreen}{green}, significant worsening of the baseline in \textcolor{BrickRed}{red}}
\end{tabu}
}
        \end{center}
    \end{minipage}%
    \vspace{-0.1cm}
\end{table}

\newpage
\subsection{Runtime}

Table~\ref{tab:runtimes} provides the runtime comparison of different models from different stages of the \ORlearners. Here, the \ORlearners are well scalable.

\begin{table}[ht]
    \centering
    \scalebox{0.9}{
    \begin{tabu}{l|l|l|c}
        \toprule
        Training stage &  Model  & Variant &  Average duration of a training iteration (in ms) \\
        \midrule
        \multirow{13}{*}{Stage \protect\circled{0}} & TARNet/TARFlow & --- & $\approx  9.3 / 85.4$ \\
         \cmidrule{2-4} & \multirow{2}{*}{BNN/BNNFlow} & MMD & $\approx 18.1 / 94.9 $ \\
         & & WM & $\approx 15.6 / 166.1$ \\
         \cmidrule{2-4} & \multirow{2}{*}{CFR/CFRFlow} & MMD & $\approx 20.7 / 115.2 $ \\
         & & WM & $\approx 21.7 / 174.0$ \\
          \cmidrule{2-4} & \multirow{2}{*}{BWCFR/BWCFRFlow} & MMD &  $\approx 19.4 / 102.0 $ \\
          & & WM & $\approx 19.7 / 101.1$ \\
         \cmidrule{2-4} & \multirow{2}{*}{CFR-ISW/CFRFlow-ISW} & MMD &  $\approx 21.2 / 103.0 $ \\
         & & WM & $\approx 23.1 / 116.0 $ \\
         \cmidrule{2-4} & \multirow{2}{*}{RCFR/RCFRFlow} & MMD & $\approx 27.9 / 101.7 $ \\
         & & WM & $\approx 27.1 /127.7 $ \\
         \midrule
         \multirow{2}{*}{Stage \protect\circled{1}} & Propensity network & --- & $\approx 5.2 $ \\
         \cmidrule{2-4} & Outcome network & --- & $\approx 6.7 $ \\
         \midrule
         {Stage \protect\circled{2}} & Target network & ---&  $\approx 6.5 $ \\
        \bottomrule
    \end{tabu}}
    \vspace{0.2cm}
    \caption{Total runtime (in milliseconds) for different models at stages \protect\circled{0}-\protect\circled{2} of the \ORlearners. Reported: average duration of a training iteration for the IHDP dataset (lower is better). Experiments were carried out on 2 GPUs (NVIDIA A100-PCIE-40GB) with IntelXeon Silver 4316 CPUs @ 2.30GHz.}
    \label{tab:runtimes}
\end{table}

\end{document}